\documentclass[10pt,twocolumn,letterpaper]{article}
\usepackage[pagenumbers]{cvpr}

\usepackage[T1]{fontenc}
\usepackage{graphicx}
\usepackage{amsmath}
\usepackage{amssymb}
\usepackage{mathbbol}
\usepackage{booktabs}
\usepackage{array}
\usepackage{longtable}
\usepackage{colortbl}
\usepackage{float}
\usepackage{placeins}
\usepackage{needspace}
\usepackage{tabularx}
\usepackage[most]{tcolorbox}

\definecolor{prompttitleblue}{RGB}{221,235,248}
\definecolor{promptborderblue}{RGB}{126,157,185}
\definecolor{bioloomlight}{RGB}{224,224,252}
\definecolor{ablationheadingblue}{HTML}{E8F1FB}
\definecolor{repaepink}{RGB}{236,0,139}

\definecolor{ablationinit}{RGB}{255,253,233}
\definecolor{ablationobjective}{RGB}{255,242,229}
\definecolor{ablationcaption}{RGB}{229,242,252}
\definecolor{ablationviews}{RGB}{229,255,229}
\definecolor{ablationcharts}{RGB}{255,240,240}
\newcommand{\ablationtitle}[2]{\begingroup\setlength{\fboxsep}{0pt}\colorbox{#1}{\strut\textbf{#2}}\endgroup}

\newtcolorbox{judgeprompt}[1]{
  enhanced,
  breakable,
  colback=gray!4,
  colframe=promptborderblue,
  boxrule=0.6pt,
  arc=1.5mm,
  left=7pt,
  right=7pt,
  top=6pt,
  bottom=6pt,
  title={#1},
  fonttitle=\bfseries,
  coltitle=black,
  colbacktitle=prompttitleblue,
  before skip=7pt,
  after skip=7pt
}

\newtcolorbox{judgepromptwide}[1]{
  enhanced,
  colback=gray!4,
  colframe=promptborderblue,
  boxrule=0.6pt,
  arc=1.5mm,
  left=7pt,
  right=7pt,
  top=5pt,
  bottom=5pt,
  title={#1},
  fonttitle=\bfseries,
  coltitle=black,
  colbacktitle=prompttitleblue,
  before skip=3pt,
  after skip=3pt
}

\newcommand{\resulttablefont}{\fontsize{8.5}{10}\selectfont}

\newcommand{\methodname}{BioLoom}
\newcolumntype{Y}{>{\centering\arraybackslash}X}

\usepackage{calc}
\newcommand{\tabnum}[1]{\makebox[\widthof{\normalfont 100.00}][r]{#1}}
\newcommand{\tabci}[2]{[\tabnum{#1},\,\tabnum{#2}]}

\definecolor{cvprblue}{rgb}{0.21,0.49,0.74}
\usepackage[breaklinks,colorlinks,allcolors=cvprblue]{hyperref}
\usepackage{enumitem}

\def\confName{CVPR}
\def\confYear{2027}

\title{\methodname: Scaling Biomedical Vision-Language Models by\\ Learning from the Structure of Scientific Articles}

\author{%
\begin{minipage}{0.96\textwidth}
\centering
\large
Yuwei Yang\textsuperscript{1}%
\quad
Xiaoxiao Sun\textsuperscript{2}\quad
Alejandro Lozano\textsuperscript{2}\quad
Yuan-Sen Ting\textsuperscript{4}\\[4pt]
Yuhui Yuan\textsuperscript{5}\quad
Liang Zheng\textsuperscript{1,3}\quad
Serena Yeung-Levy\textsuperscript{2}\\[8pt]
\textsuperscript{1}Australian National University\qquad
\textsuperscript{2}Stanford University\qquad
\textsuperscript{3}The University of Sydney\\[3pt]
\textsuperscript{4}The Ohio State University\qquad
\textsuperscript{5}Canva\\[5pt]
{\small
\texttt{\{yuwei.yang, liang.zheng\}@anu.edu.au}\quad
\texttt{\{xxsun, lozanoe, syyeung\}@stanford.edu}\\[2pt]
\texttt{ting.74@osu.edu}\quad
\texttt{ryanyuan@canva.com}}
\end{minipage}}

\makeatletter
\patchcmd{\@maketitle}{\vskip .375in}{\vskip .15in}{}{}
\patchcmd{\@maketitle}{\vspace*{12pt}}{\vspace*{7pt}}{}{}
\patchcmd{\@maketitle}{\vspace*{24pt}}{\vspace*{12pt}}{}{}
\patchcmd{\abstract}{\vspace*{12pt}}{\vspace*{4.25pt}}{}{}
\makeatother

\begin{document}

\maketitle

\begin{abstract}
The development of biomedical vision-language models (VLMs) with contrastive image-text pretraining has been driven by scaling to increasingly large and diverse image-text datasets. Beyond scaling, further progress can come from learning richer representations from existing data. We hypothesize that the structure of scientific literature itself provides an additional source of supervision: figures, panels, captions, and accompanying article text encode rich relationships across modalities. Yet, treating these as independent image-text pairs overlooks their relationships. To address this limitation, we incorporate these structural relationships into contrastive image-text supervision. We preserve exact image-text matches while introducing weaker positive supervision for related pairs, such as a panel and the caption of a sibling panel. We upweight cross-article negatives of the same content type to encourage finer discrimination. We also enrich panel descriptions using visual content, original figure captions, and mentioning paragraphs. Pairing panels with these descriptions and whole figures with their original captions yields \textbf{Loom52M}, a dataset of 52.4 million image-text pairs spanning scientific charts and biomedical images. Using Loom52M, \textbf{BioLoom} (DFN-2B) achieves 56.82\% domain-balanced mean accuracy across 67 zero-shot classification tasks from 43 datasets in 10 domains, compared with 48.77\% for MedPMC-CLIP. On BioLoomBench with concise captions and BioMedFlickr, it achieves bidirectional mean Recall@1 of 63.20\% and 10.59\%, versus 46.37\% and 7.55\% for the respective strongest evaluated baselines, BiomedCLIP and MedPMC-CLIP. Code, datasets, benchmarks, and models are available at \href{https://bioloom-clip.github.io/}{\textcolor{repaepink}{https://bioloom-clip.github.io/}}.
\end{abstract}

\section{Introduction}
\label{sec:introduction}

\begin{figure}[!t]
  \centering
  \includegraphics[width=0.95\columnwidth]{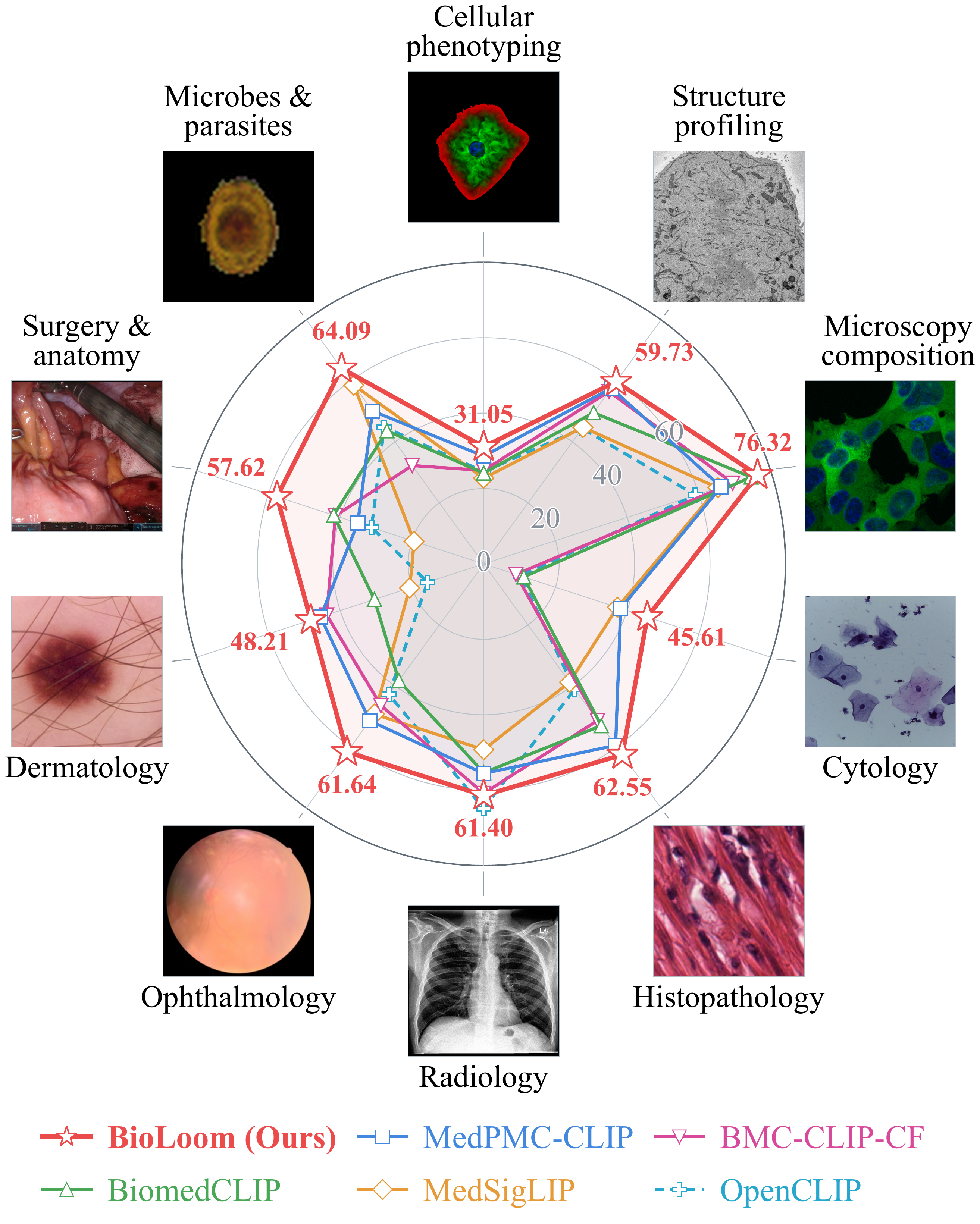}
  \captionsetup{skip=3pt}
  \caption{\textbf{Zero-shot classification accuracy (\%) across 10 domains (43 datasets, 67 tasks).} BioLoom has the highest domain-balanced mean accuracy (56.82\%) among the \mbox{evaluated biomedical models}.}
  \label{fig:zero-shot-domain-radar}
\end{figure}

The success of vision-language models (VLMs) in general domains has sparked growing interest in developing biomedical embedding models that learn shared representations of images and text across diverse biomedical disciplines. By mapping images and their textual descriptions into a common representation space, these models enable cross-modal retrieval, connecting visual findings with large-scale scientific literature and clinical databases. This is particularly valuable in biomedicine, where images across modalities such as radiology, histopathology, and microscopy capture \mbox{biological phenomena at multiple scales}.

To this end, scientific literature has emerged as a source of large-scale biomedical image-text supervision for both pretraining and post-training~\cite{zhang2023biomedclipmultimodalbiomedicalfoundation,lozano2025biomedica}. Recent work improves panel-caption alignment, enriches captions with article context, and introduces hierarchical supervision~\cite{lin2023pmcclipcontrastivelanguageimagepretraining,kim2026medpmc,sun2025tokenswastedleveraginglong,yuan2026panel2patch}. However, standard pairwise contrastive supervision does not distinguish exact matches from related but non-matching pairs.
Scientific articles contain multi-panel figures, figure-level captions, panel-specific descriptions, and textual discussions that collectively convey experimental context and scientific findings. These components are not independent: neighboring panels may present complementary measurements or controls, while panels across figures may describe related stages of an experiment.

These relationships offer guidance on how each image-text pair should contribute to training. Building on these cues, we introduce BioLoom, a biomedical vision-language model that uses figure and article relationships for contrastive learning. BioLoom achieves strong zero-shot classification performance across ten biomedical domains (Fig.~\ref{fig:zero-shot-domain-radar}).

\begin{figure}[t]
  \centering
  \includegraphics[width=\columnwidth]{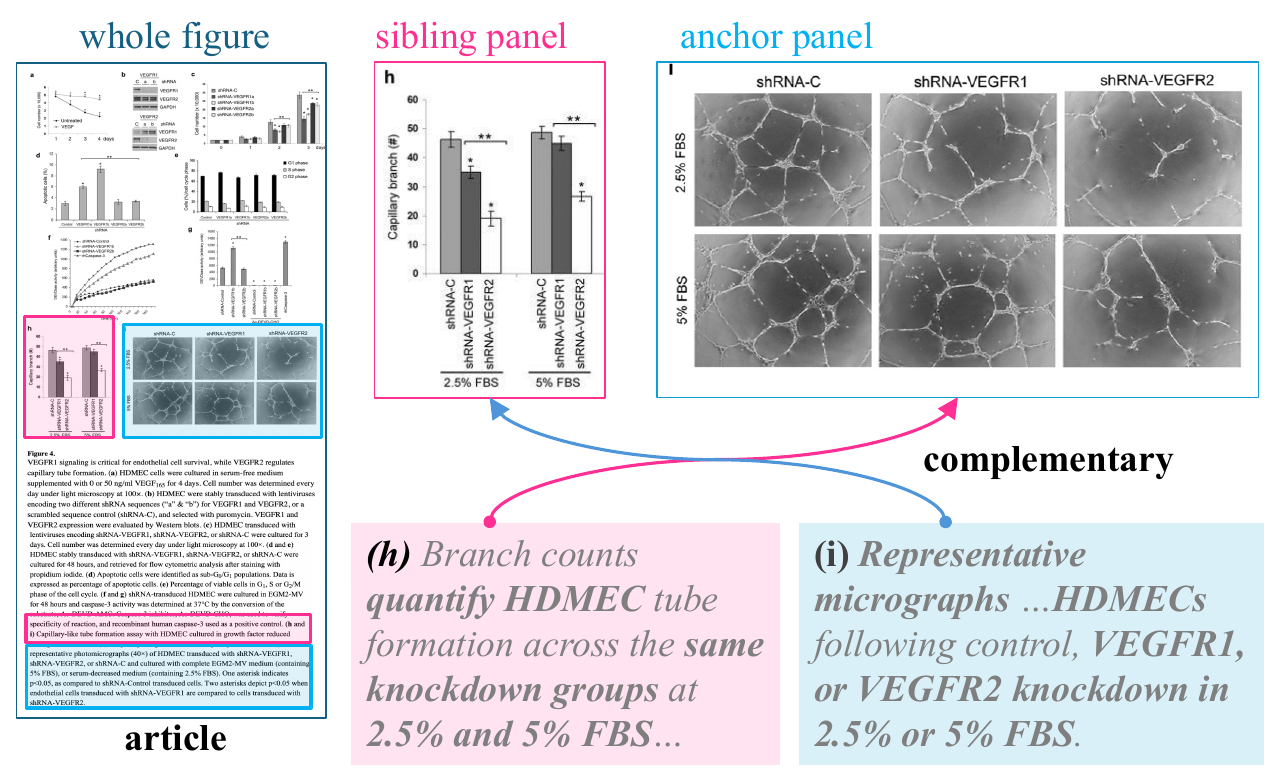}
  \captionsetup{skip=3pt}
  \caption{\textbf{Semi-positive image-text pairs within a whole figure.} The anchor microscopy panel (i) shows tube formation, while the sibling chart (h) reports capillary branch counts under the same experimental conditions. The chart caption provides complementary context for the anchor image. We treat this image-text pair as \textit{semi-positive} rather than an exact match or a negative.}
  \label{fig:article-structure}
\end{figure}

A key challenge in designing BioLoom is determining how different image-text relationships should contribute to training: related components should not all be treated as equivalent training examples.
The dedicated caption of a panel provides a direct semantic match, whereas the caption of a sibling panel or the whole figure may describe complementary aspects of the same experiment. For example, the anchor microscopy panel in Fig.~\ref{fig:article-structure} shows cellular morphology, while its sibling chart quantifies the observed changes.

Panels from different figures within the same article may share experimental context, although their semantic relatedness may vary.
Panels from different articles may depict similar content types while describing unrelated biological findings. Treating all these relationships as either positive or negative pairs risks introducing noisy supervision: overly broad positive assignments can blur distinctions between visual concepts, while treating related components as negatives can penalize meaningful semantic relationships.

BioLoom addresses this challenge through two complementary objectives. The primary contrastive objective aligns exact image-caption pairs while excluding other pairs from the same article from its negative set. The relational objective provides weaker positive supervision between a panel image and captions of sibling panels or the whole figure, and upweights cross-article negatives of the same content type.

To support this structure-aware training, we introduce a data restructuring and enrichment pipeline on top of the figure records in BIOMEDICA~\cite{lozano2025biomedica}. When captioning a panel in a figure, we use its visual content, the original figure caption, and article paragraphs that mention the figure. Each panel receives a visual description of visible content and a contextual caption that incorporates relevant article context. We pair the panel crops with these enriched descriptions and retain whole figures with their original captions. To prioritize biomedical visual content, we exclude the whole figures composed entirely of charts. We retain figures that combine biomedical images and charts, as the chart panels can provide quantitative context for the accompanying biomedical images. The resulting dataset, Loom52M, contains 52.4M image-text pairs spanning biomedical images and charts.

We train \methodname{} on Loom52M with a pretrained BioClinical ModernBERT text encoder and a ViT-L/14 vision encoder initialized either randomly or from DFN-2B. The DFN-2B-initialized variant achieves 56.82\% domain-balanced mean accuracy across 67 zero-shot classification tasks from 43 datasets in 10 domains, compared with 48.77\% for MedPMC-CLIP. We also introduce BioLoomBench, a retrieval benchmark of 15,000 biomedical and chart figures. \methodname{} achieves bidirectional mean Recall@1 of 63.20\% with concise captions on BioLoomBench and 10.59\% on BioMedFlickr. We contribute:

\begin{itemize}[leftmargin=*,itemsep=1pt,topsep=2pt]

\item \textbf{Structure-aware image-text labeling.} We label image-text pairs using article, figure, and panel relationships.

\item \textbf{Revised contrastive objective.} We combine exact-pair and semi-positive alignment with related-pair masking and same-content-type hard-negative weighting.

\item \textbf{Training data.} We construct Loom52M with 52.4M pairs, matching whole figures with original captions and panels with enriched descriptions.

\item \textbf{Models and evaluation.} \methodname{} achieves strong performance in biomedical classification and image-text retrieval. We introduce BioLoomBench for retrieval evaluation across biomedical images, charts, and diagrams, and will release our dataset, models, and benchmarks.
\end{itemize}

\section{Related Work}
\label{sec:related_work}

\textbf{Biomedical image-text datasets and curation.} Clinical images, reports, and scientific articles support biomedical AI training. Clinical collections provide chest radiographs with reports or diagnostic labels~\cite{demnerfushman2016preparing,johnson2019mimiccxr,wang2017chestxray8,irvin2019chexpert,bustos2020padchest,nguyen2022vindrcxr}, while Quilt-1M focuses on pathology, mainly using educational videos and other public sources~\cite{ikezogwo2023quilt}. Scientific articles support large-scale, multi-domain collections such as PMC-15M and BIOMEDICA~\cite{zhang2023biomedclipmultimodalbiomedicalfoundation,lozano2025biomedica}. ROCO pairs radiology images with captions and concepts~\cite{pelka2018roco}; MedICaT adds in-text references and subfigure-subcaption annotations for a subset of figures~\cite{subramanian2020medicat}. Open-PMC and MedPMC improve pair quality through image filtering and figure separation~\cite{baghbanzadeh2025openpmc,kim2026medpmc}. Original captions may still omit panel-specific details. We construct Loom52M to link whole figures with panels and enrich descriptions using visible content, original captions, and mentioning paragraphs.

\textbf{Biomedical vision-language learning.} CLIP, ALIGN, and OpenCLIP learn shared image-text representations through large-scale contrastive training~\cite{radford2021clip,jia2021align,cherti2023reproduciblescalinglawscontrastive}. SigLIP uses a pairwise sigmoid loss~\cite{zhai2023siglip}, while CoCa adds caption generation~\cite{yu2022cocacontrastivecaptionersimagetext}. ALBEF, BLIP, and BLIP-2 explore multimodal fusion and generation~\cite{li2021albef,li2022blip,li2023blip2}; Long-CLIP and LoTLIP support longer text~\cite{zhang2024longclip,wu2024lotlip}. Biomedical models learn global or local alignment from radiology images and reports~\cite{pmlr-v182-zhang22a,huang2021gloria,boecking2022making,mueller2022lovt,wang2022mgca}, while MedCLIP uses medical semantics beyond original pairings~\cite{wang2022medclip}. Other radiology studies incorporate temporal information~\cite{bannur2023biovilt}, scale chest X-ray pretraining~\cite{you2023cxrclip}, or introduce medical knowledge~\cite{wu2023medklip,zhang2023kad}. PMC-CLIP, BiomedCLIP, and BIOMEDICA models extend coverage across biomedical domains using scientific image-caption pairs~\cite{lin2023pmcclipcontrastivelanguageimagepretraining,zhang2023biomedclipmultimodalbiomedicalfoundation,lozano2025biomedica}. Alongside broader domain coverage, studies have improved text supervision and pair quality. BMC-LongCLIP enriches captions with article context~\cite{sun2025tokenswastedleveraginglong}, while MedPMC-CLIP uses image filtering, panel separation, and text alignment~\cite{kim2026medpmc}. Another line of work studies finer image-text correspondences by aligning image patches or regions with text or learning from related image-text groups~\cite{yao2022filip,zeng2022xvlm,zhong2022regionclip,javed2024cplip}. For scientific figures, Panel2Patch constructs supervision at the figure, panel, and region levels~\cite{yuan2026panel2patch}, while FigEx2 detects panels and generates captions~\cite{song2026figex2}. Our approach uses figure and article relationships to guide contrastive learning beyond individual image-text pairs.

\section{Method}
\label{sec:method}
\begin{figure*}[t]
  \centering

  \includegraphics[width=\textwidth]{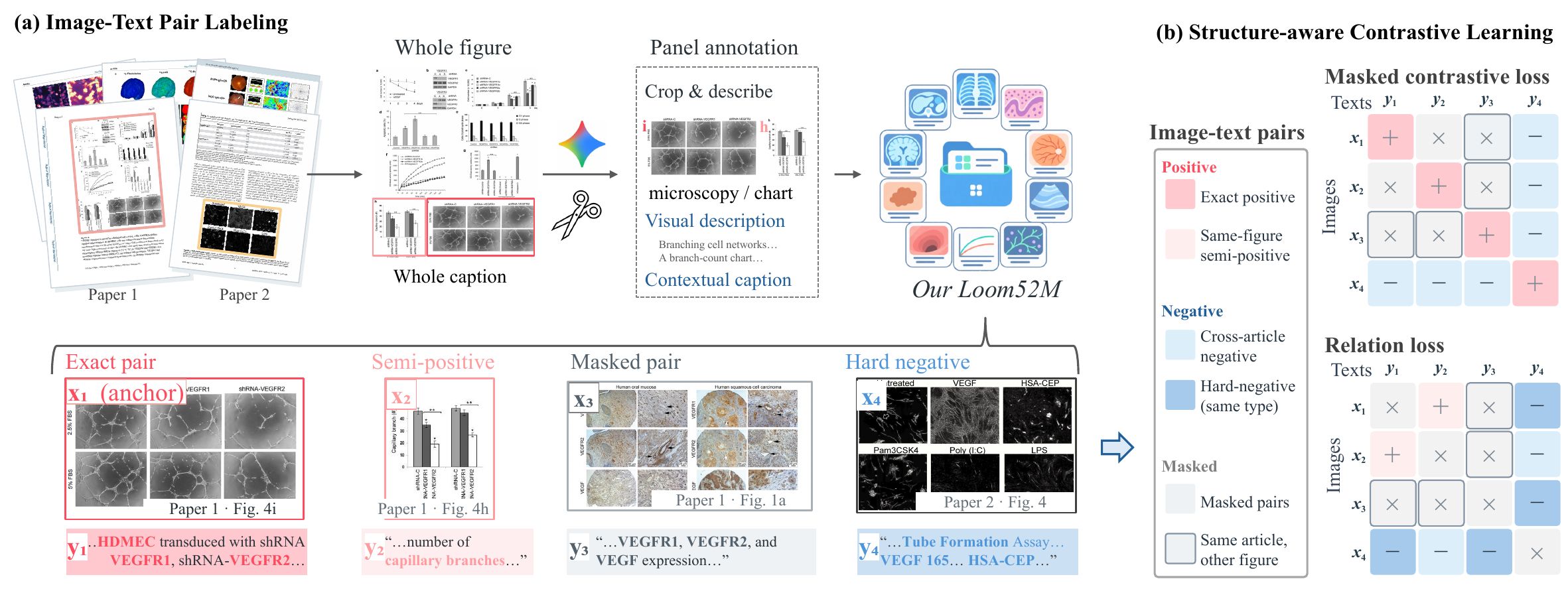}
  \caption{\textbf{Method overview.} (a) Whole figures, original captions, and mentioning paragraphs support panel localization and the generation of visual descriptions and contextual captions. For image anchor $x_1$, texts $y_1$-$y_4$ illustrate an exact pair, a same-figure semi-positive pair, a masked same-article cross-figure pair, and a same-type cross-article hard negative, respectively. The sibling chart provides quantitative context for the microscopy image without being an exact match. Displayed text excerpts illustrate these relationships. (b) $\mathcal{L}_{\mathrm{mask}}$ aligns exact pairs while masking related pairs; $\mathcal{L}_{\mathrm{rel}}$ aligns semi-positive pairs and upweights same-type cross-article negatives.}
  \label{fig:method}
\end{figure*}

\textbf{Overview.} BioLoom uses relationships among figures, panels, and text in scientific articles to guide image-text learning (Fig.~\ref{fig:method}). We first define pair labels (Sec.~\ref{sec:labels}) and the training objective (Sec.~\ref{sec:loss}), followed by Loom52M construction (Sec.~\ref{sec:data}) and training settings (Sec.~\ref{sec:pretraining}).

\subsection{Image-Text Pair Labeling}
\label{sec:labels}

\textbf{Pair definitions.} Each sample pairs a whole figure with its original caption or a panel with its visual description and contextual caption (Sec.~\ref{sec:data}). For image anchor $x_i$ and candidate text $y_j$, we define four disjoint candidate sets:
\begin{itemize}[leftmargin=*,itemsep=0pt,parsep=0pt,topsep=2pt]
    \item \textit{Exact pairs ($\mathcal{E}_i$):} $y_j$ is paired with the same image as~$x_i$.
    \item \textit{Semi-positive pairs ($\mathcal{P}_i$):} $y_j$ is paired with a different sample from the same figure, including sibling panels and whole-figure/panel combinations.
    \item \textit{Within-article cross-figure pairs ($\mathcal{A}_i$):} $y_j$ is paired with an image from a different figure in the same article.
    \item \textit{Cross-article pairs ($\mathcal{N}_i$):} $y_j$ is paired with an \mbox{image from another article}.
\end{itemize}

\subsection{Structure-aware Contrastive Learning}
\label{sec:loss}

\textbf{InfoNCE loss.} These labels distinguish exact alignment from related-pair supervision. Standard CLIP uses the following InfoNCE objective:

\begingroup
\setlength{\abovedisplayskip}{6pt plus 2pt minus 2pt}
\setlength{\belowdisplayskip}{6pt plus 2pt minus 2pt}
\begin{equation}
\label{eq:infonce-i2t}
\mathcal{L}_{\mathrm{I2T}}
=-\frac{1}{N}\sum_{i=1}^{N}
\log\frac{\exp(v_i^\top t_i/\tau)}
{\sum_{j=1}^{N}\exp(v_i^\top t_j/\tau)},
\end{equation}
\begin{equation}
\label{eq:infonce-t2i}
\mathcal{L}_{\mathrm{T2I}}
=-\frac{1}{N}\sum_{i=1}^{N}
\log\frac{\exp(t_i^\top v_i/\tau)}
{\sum_{j=1}^{N}\exp(t_i^\top v_j/\tau)},
\end{equation}
\begin{equation}
\label{eq:infonce-symmetric}
\mathcal{L}_{\mathrm{CLIP}}
=\tfrac12(\mathcal{L}_{\mathrm{I2T}}+\mathcal{L}_{\mathrm{T2I}}).
\end{equation}
\endgroup

Here, $N$ is the batch size, $v_i$ and $t_i$ are the $\ell_2$-normalized image and text embeddings of pair $i$, and $\tau>0$ is the temperature. The anchor pair is $i$, and $j$ denotes a candidate in the batch. Standard CLIP treats $(v_i,t_i)$ as the sole positive for image $i$ and pairs with $j\neq i$ as negatives; InfoNCE favors positive over negative pairs. $\mathcal{L}_{\mathrm{I2T}}$ and $\mathcal{L}_{\mathrm{T2I}}$ denote image-to-text and text-to-image losses, \mbox{with average $\mathcal{L}_{\mathrm{CLIP}}$}.

\textbf{Exact-pair alignment with related-pair masking.} To avoid treating semi-positive pairs as negatives, we exclude $\mathcal{P}_i$ from the main loss denominator. We also exclude $\mathcal{A}_i$ because different figures in an article may share research context. These are semi-positive and cross-figure masking, respectively; the original paired text remains the positive target. With $s_{ij}=v_i^\top t_j/\tau$, the I2T loss is
\begin{equation}
\ell_{i,\mathrm{mask}}^{\mathrm{I2T}}
=
-\log
\frac{\exp(s_{ii})}
{\sum_{j\in\mathcal{E}_i\cup\mathcal{N}_i}\exp(s_{ij})}.
\label{eq:masked-exact}
\end{equation}
Aggregating over anchors and averaging the image-to-text and text-to-image directions gives $\mathcal{L}_{\mathrm{mask}}$.

\textbf{Semi-positive alignment.} Our relation loss, $\mathcal{L}_{\mathrm{rel}}$, uses within-figure semi-positive pairs $\mathcal{P}_i$ as positives and cross-article pairs $\mathcal{N}_i$ as negatives. It masks $\mathcal{E}_i$ to focus on semi-positive associations and $\mathcal{A}_i$ to avoid negatives with shared article context. The numerator therefore contains $\mathcal{P}_i$, and the denominator contains $\mathcal{P}_i\cup\mathcal{N}_i$. To distinguish images of the same content type, we upweight same-type cross-article pairs as hard negatives. Let $c_i$ be the content-type label assigned to sample $i$ by our data pipeline. The denominator weight is $w_{ij}=\alpha\geq1$ when $j\in\mathcal{N}_i$ and $c_j=c_i$, and $w_{ij}=1$ for all other included pairs. For anchors with semi-positive candidates, we use:
\begin{equation}
\ell_{i,\mathrm{rel}}^{\mathrm{I2T}}
=-\log\frac{\sum_{j\in\mathcal{P}_i}\exp(s_{ij})}
{\sum_{j\in\mathcal{P}_i\cup\mathcal{N}_i}w_{ij}\exp(s_{ij})}.
\label{eq:relation-alignment}
\end{equation}
We obtain $\mathcal{L}_{\mathrm{rel}}$ by averaging these losses over anchors and both directions.

\textbf{Final loss function.} We combine both objectives:
\begin{equation}
\mathcal{L}
=\mathcal{L}_{\mathrm{mask}}+\lambda\mathcal{L}_{\mathrm{rel}},
\label{eq:final-objective}
\end{equation}
where $\lambda$ weights the relation loss. The defaults are $\lambda=0.05$ and $\alpha=1.1$ for same-content-type hard negatives.

\subsection{Data Pipeline}
\label{sec:data}
To support this training objective, we construct Loom52M with linked article, figure, and panel records from BIOMEDICA~\cite{lozano2025biomedica}. We provide Gemini 3.5 Flash-Lite with the whole figure, its original caption, and relevant article paragraphs. The model uses these inputs to locate panels and generate their descriptions through cross-component caption enrichment. The inputs and outputs follow.

\textbf{Input: the whole figure.} Of the retained whole figures in Loom52M, 57.68\% contain at least two extracted panels. We provide the whole figure for panel localization and description. This gives the annotation model access to each panel and its surrounding figure context.

\textbf{Input: the original figure caption.} We provide the full original caption as context for panel annotation. It supplies information about the figure and its constituent panels, such as experimental conditions and panel references.

\textbf{Input: figure mentions in the article.} We additionally provide up to two paragraphs from the article body that mention the figure. These paragraphs supply context that may not be available from the image or caption alone.

\textbf{Output: panel bounding boxes.} The model returns panel boxes as $[y_{\min},x_{\min},y_{\max},x_{\max}]$, normalized to $[0,1000]$. We convert to pixels and crop each panel.

\textbf{Output: panel descriptions.} For each panel, the model generates two complementary text fields. The \emph{visual description} summarizes visible content only. The \emph{contextual caption} combines this content with information from the original caption and mentioning paragraphs. Each field contains one or two sentences, and the prompt discourages repetition between them. We concatenate the two fields and pair the resulting text with the panel crop. We also retain whole-figure samples, pairing each uncropped image with its original caption.
\hyperref[app:figure-annotation]{Supp.~A} provides the full prompt. \hyperref[app:computational-cost]{Supp.~K} details annotation and pretraining costs.

\begin{table}[!t]
  \caption{\textbf{Comparison of biomedical image-text datasets.} Whole/Panel: whole-figure/panel-text pairs; Ctx.: article mentioning paragraphs; Visual: independent image-content descriptions; Bio/Charts: biomedical images and charts/diagrams. $\checkmark$: provided; $\triangle$: subset; $\times$: not specified.}
  \label{tab:dataset-comparison}
  \centering
  \resulttablefont
  \setlength{\tabcolsep}{1pt}
  \renewcommand{\arraystretch}{1.12}
  \begin{tabularx}{\linewidth}{lc>{\hsize=1.2\hsize\linewidth=\hsize\centering\arraybackslash}X>{\hsize=1.0\hsize\linewidth=\hsize\centering\arraybackslash}X>{\hsize=0.75\hsize\linewidth=\hsize\centering\arraybackslash}X>{\hsize=1.15\hsize\linewidth=\hsize\centering\arraybackslash}X>{\hsize=0.65\hsize\linewidth=\hsize\centering\arraybackslash}X>{\hsize=1.25\hsize\linewidth=\hsize\centering\arraybackslash}X}
    \toprule
    & & \multicolumn{2}{c}{\textbf{Views}} & \multicolumn{2}{c}{\textbf{Text}} & \multicolumn{2}{c}{\textbf{Coverage}} \\
    \cmidrule(lr){3-4}\cmidrule(lr){5-6}\cmidrule(lr){7-8}
    \textbf{Dataset} & \textbf{Pairs} & \textbf{Whole} & \textbf{Panel} & \textbf{Ctx.} & \textbf{Visual} & \textbf{Bio} & \textbf{Charts} \\
    \midrule
    ROCO~\cite{pelka2018roco} & 81K & $\checkmark$ & $\times$ & $\times$ & $\times$ & $\checkmark$ & $\times$ \\
    MedICaT~\cite{subramanian2020medicat} & 217K & $\checkmark$ & $\triangle$ & $\triangle$ & $\times$ & $\checkmark$ & $\times$ \\
    PMC-OA~\cite{lin2023pmcclipcontrastivelanguageimagepretraining} & 1.65M & $\times$ & $\checkmark$ & $\times$ & $\times$ & $\checkmark$ & $\times$ \\
    PMC-15M~\cite{zhang2023biomedclipmultimodalbiomedicalfoundation} & 15M & $\checkmark$ & $\times$ & $\times$ & $\times$ & $\checkmark$ & $\checkmark$ \\
    Quilt-1M~\cite{ikezogwo2023quilt} & 1M & $\times$ & $\times$ & $\times$ & $\times$ & $\checkmark$ & $\times$ \\
    BIOMEDICA~\cite{lozano2025biomedica} & 24M & $\checkmark$ & $\times$ & $\checkmark$ & $\times$ & $\checkmark$ & $\checkmark$ \\
    Open-PMC~\cite{baghbanzadeh2025openpmc} & 2.2M & $\times$ & $\checkmark$ & $\checkmark$ & $\times$ & $\checkmark$ & $\times$ \\
    MedPMC~\cite{kim2026medpmc} & 11M & $\checkmark$ & $\checkmark$ & $\checkmark$ & $\times$ & $\checkmark$ & $\times$ \\
    \midrule
    \rowcolor[RGB]{237,244,232}
    Loom52M (Ours) & 52.4M & $\checkmark$ & $\checkmark$ & $\checkmark$ & $\checkmark$ & $\checkmark$ & $\checkmark$ \\
    \bottomrule
  \end{tabularx}
\end{table}

\textbf{Filtering and dataset construction.} We remove over 10.8M whole figures whose annotated panels consist entirely of charts, together with their panel crops, to focus training on biomedical visual content. We retain figures that mix biomedical images and charts, including their chart panels, to preserve potentially related visual observations within each figure. To ensure valid training pairs, we also discard records with unreadable images or empty paired text. We retain article and figure IDs to preserve the links between articles, whole figures, and panels for structure-aware training. The constructed dataset contains 13.1M whole-figure pairs and 39.3M panel pairs. We use the full 52.4M-pair training dataset for pretraining.

\textbf{Comparison with existing datasets.} Tab.~\ref{tab:dataset-comparison} compares dataset scale, image views, text supervision, and visual coverage. Beyond whole figures paired with original captions, Loom52M provides finer-grained, richer supervision through panel-level visual descriptions and contextual captions. Its 52.4M pairs preserve article--figure--panel links for structure-aware learning. Retaining charts in mixed figures supports joint learning and text-based retrieval of biomedical images and scientific charts.

\subsection{Pretraining}
\label{sec:pretraining}
BioLoom uses the CLIP dual-encoder architecture~\cite{radford2021clip}, with a ViT-L/14 vision encoder~\cite{dosovitskiy2021image} and a BioClinical ModernBERT text encoder~\cite{sounack2025bioclinicalmodernbert}. By default, the vision encoder is randomly initialized, while the text encoder uses weights pretrained on biomedical literature and clinical text. Both encoders are jointly trained on all 52.4M Loom52M image-text pairs for 10 epochs using Eq.~\ref{eq:final-objective}. A second variant initializes the vision encoder from DFN-2B-pretrained ViT-L/14~\cite{fang2024data}, retaining the same text initialization. Sec.~\ref{sec:experimental-setup} provides implementation details.

\begin{table*}[!t]
\def\jointmaintables{1}

\ifdefined\jointmaintables\begingroup\else\begin{table*}[t]\fi
  \caption{\textbf{Model comparison on zero-shot classification across ten biomedical domains.} We report accuracy (\%) on 67 tasks from 43 datasets. We average tasks within each dataset, then datasets within each domain. `Overall' is the domain-balanced mean accuracy, with equal weight for each of the 10 domains.
  Cell: cellular phenotyping. Struct.: structure profiling. Micro.: microscopy composition. Histo.: histopathology. Radiol.: radiology. Ophth.: ophthalmology. Derm.: dermatology. Microb.: microbial and parasitic recognition. Surgery: surgery and anatomy. For \methodname{}, Scratch and DFN-2B denote random and DFN-2B initialization of the vision encoder, respectively; in both cases we use the same pretrained text encoder. $^{\ast}$ denotes random vision-encoder initialization for image-text pretraining. Among biomedical models, \textbf{bold} and \underline{underlined values} indicate the best and second-best scores in each column, respectively.}
  \vspace{-1mm}
  \label{tab:zero-shot}
  \centering
  \resulttablefont
  \renewcommand{\arraystretch}{1.05}
  \setlength{\tabcolsep}{0.8pt}
  \begin{tabularx}{\textwidth}{>{\raggedright\arraybackslash}p{0.17\textwidth}>{\centering\arraybackslash}p{0.065\textwidth}*{11}{Y}}
    \toprule
    \textbf{Model} & \textbf{Params.} & \multicolumn{1}{c}{\textbf{Cell}} & \multicolumn{1}{c}{\textbf{Struct.}} & \multicolumn{1}{c}{\textbf{Micro.}} & \multicolumn{1}{c}{\textbf{Cytology}} & \multicolumn{1}{c}{\textbf{Histo.}} & \multicolumn{1}{c}{\textbf{Radiol.}} & \multicolumn{1}{c}{\textbf{Ophth.}} & \multicolumn{1}{c}{\textbf{Derm.}} & \multicolumn{1}{c}{\textbf{Surgery}} & \multicolumn{1}{c}{\textbf{Microb.}} & \multicolumn{1}{c}{\textbf{Overall}} \\
    \midrule
    Random & - & \tabnum{18.65} & \tabnum{17.78} & \tabnum{34.42} & \tabnum{13.77} & \tabnum{28.96} & \tabnum{38.18} & \tabnum{29.67} & \tabnum{13.23} & \tabnum{19.63} & \tabnum{29.23} & \tabnum{24.35} \\
    CoCa$^{\ast}$~\cite{yu2022cocacontrastivecaptionersimagetext}  &  430M & \tabnum{22.18} & \tabnum{33.45} & \tabnum{56.86} & \tabnum{41.56} & \tabnum{39.80} & \tabnum{36.31} & \tabnum{50.53} & \tabnum{17.37} & \tabnum{23.05} & \tabnum{31.48} & \tabnum{35.26} \\
    OpenCLIP$^{\ast}$~\cite{cherti2023reproduciblescalinglawscontrastive}  &  428M & \tabnum{24.49} & \tabnum{44.44} & \tabnum{59.04} & \tabnum{10.25} & \tabnum{41.03} & \tabnum{64.86} & \tabnum{42.80} & \tabnum{15.67} & \tabnum{31.26} & \tabnum{44.87} & \tabnum{37.87} \\
    \midrule
    \mbox{PMC-CLIP}$^{\ast}$~\cite{lin2023pmcclipcontrastivelanguageimagepretraining}  &  137M & \tabnum{28.16} & \tabnum{48.95} & \tabnum{69.20} & \tabnum{15.16} & \tabnum{57.30} & \tabnum{53.91} & \tabnum{38.74} & \tabnum{16.34} & \tabnum{29.09} & \tabnum{38.16} & \tabnum{39.50} \\
    MedSigLIP~\cite{sellergren2025medgemmatechnicalreport}   &   800M & \tabnum{22.74} & \tabnum{44.77} & \tabnum{65.31} & \tabnum{37.28} & \tabnum{38.77} & \tabnum{49.26} & \tabnum{49.24} & \tabnum{20.62} & \tabnum{19.38} & \tabnum{\underline{58.58}} & \tabnum{40.59} \\
    BiomedCLIP~\cite{zhang2023biomedclipmultimodalbiomedicalfoundation}  &  196M & \tabnum{24.11} & \tabnum{49.59} & \tabnum{74.55} & \tabnum{11.16} & \tabnum{53.02} & \tabnum{55.36} & \tabnum{38.27} & \tabnum{30.48} & \tabnum{41.83} & \tabnum{43.63} & \tabnum{42.20} \\
    Panel2Patch~\cite{yuan2026panel2patch} & 428M & \tabnum{29.40} & \tabnum{47.12} & \tabnum{75.66} & \tabnum{9.50} & \tabnum{52.30} & \tabnum{\underline{63.31}} & \tabnum{47.56} & \tabnum{43.76} & \tabnum{37.34} & \tabnum{27.77} & \tabnum{43.37} \\
    \mbox{BMC-CLIP-CF}~\cite{lozano2025biomedica}  &  428M & \tabnum{24.79} & \tabnum{56.53} & \tabnum{69.39} & \tabnum{8.97} & \tabnum{51.34} & \tabnum{60.85} & \tabnum{46.38} & \tabnum{43.76} & \tabnum{41.35} & \tabnum{32.25} & \tabnum{43.56} \\
    \mbox{BMC-LongCLIP}~\cite{sun2025tokenswastedleveraginglong}  &  454M & \tabnum{25.62} & \tabnum{51.49} & \tabnum{76.23} & \tabnum{35.61} & \tabnum{53.94} & \tabnum{60.73} & \tabnum{35.53} & \tabnum{40.87} & \tabnum{37.38} & \tabnum{28.62} & \tabnum{44.60} \\
    \mbox{MedPMC-CLIP}~\cite{kim2026medpmc}  &  428M & \tabnum{28.60} & \tabnum{\underline{57.60}} & \tabnum{66.17} & \tabnum{\underline{38.18}} & \tabnum{59.49} & \tabnum{55.57} & \tabnum{51.40} & \tabnum{\underline{45.49}} & \tabnum{35.11} & \tabnum{50.11} & \tabnum{48.77} \\
    \midrule
    \rowcolor{bioloomlight}
    \methodname{} (Scratch)$^{\ast}$ & 454M & \tabnum{\textbf{34.22}} & \tabnum{56.69} & \tabnum{\textbf{79.70}} & \tabnum{37.96} & \tabnum{\underline{61.81}} & \tabnum{\textbf{64.57}} & \tabnum{\underline{56.65}} & \tabnum{38.75} & \tabnum{\underline{52.65}} & \tabnum{45.90} & \tabnum{\underline{52.89}} \\
    \rowcolor{bioloomlight}
    \methodname{} (DFN-2B) & 454M & \tabnum{\underline{31.05}} & \tabnum{\textbf{59.73}} & \tabnum{\underline{76.32}} & \tabnum{\textbf{45.61}} & \tabnum{\textbf{62.55}} & \tabnum{61.40} & \tabnum{\textbf{61.64}} & \tabnum{\textbf{48.21}} & \tabnum{\textbf{57.62}} & \tabnum{\textbf{64.09}} & \tabnum{\textbf{56.82}} \\
    \bottomrule
  \end{tabularx}

\ifdefined\jointmaintables\endgroup\else\end{table*}\fi

\vspace{8pt}

\ifdefined\jointmaintables\begingroup\else\begin{table*}[t]\fi
  \caption{\textbf{Model comparison on BioLoomBench retrieval.} We report Recall@$K$ (R@$K$, \%) for concise and original captions. I2T and T2I denote image-to-text and text-to-image retrieval, respectively. The asterisk and score highlighting follow Tab.~\ref{tab:zero-shot}.}
  \vspace{1mm}
  \label{tab:bioloombench-retrieval}
  \centering
  \resulttablefont
  \renewcommand{\arraystretch}{1.05}
  \setlength{\tabcolsep}{2pt}
  \begin{tabularx}{\textwidth}{>{\raggedright\arraybackslash}p{0.15\textwidth}*{12}{>{\centering\arraybackslash}X}}
    \toprule
    & \multicolumn{6}{c}{\textbf{Concise caption}} & \multicolumn{6}{c}{\textbf{Original caption}} \\
    \cmidrule(lr){2-7}\cmidrule(lr){8-13}
    \textbf{Model} & \multicolumn{3}{c}{\textbf{I2T}} & \multicolumn{3}{c}{\textbf{T2I}} & \multicolumn{3}{c}{\textbf{I2T}} & \multicolumn{3}{c}{\textbf{T2I}} \\
    & R@1 & R@10 & R@100 & R@1 & R@10 & R@100 & R@1 & R@10 & R@100 & R@1 & R@10 & R@100 \\
    \midrule
    CoCa$^{\ast}$~\cite{yu2022cocacontrastivecaptionersimagetext} & \tabnum{24.25} & \tabnum{46.37} & \tabnum{74.44} & \tabnum{22.35} & \tabnum{42.85} & \tabnum{71.15} & \tabnum{18.37} & \tabnum{36.88} & \tabnum{64.68} & \tabnum{18.53} & \tabnum{36.72} & \tabnum{62.48} \\
    OpenCLIP$^{\ast}$~\cite{cherti2023reproduciblescalinglawscontrastive} & \tabnum{27.65} & \tabnum{50.47} & \tabnum{76.93} & \tabnum{25.69} & \tabnum{47.82} & \tabnum{75.24} & \tabnum{25.34} & \tabnum{47.37} & \tabnum{73.49} & \tabnum{24.68} & \tabnum{45.66} & \tabnum{71.75} \\
    \midrule
    \mbox{PMC-CLIP}$^{\ast}$~\cite{lin2023pmcclipcontrastivelanguageimagepretraining} & \tabnum{4.75} & \tabnum{14.76} & \tabnum{33.11} & \tabnum{4.84} & \tabnum{15.87} & \tabnum{35.96} & \tabnum{4.47} & \tabnum{13.69} & \tabnum{30.34} & \tabnum{6.05} & \tabnum{18.04} & \tabnum{37.51} \\
    MedSigLIP~\cite{sellergren2025medgemmatechnicalreport} & \tabnum{27.39} & \tabnum{56.07} & \tabnum{82.69} & \tabnum{16.38} & \tabnum{38.63} & \tabnum{67.55} & \tabnum{19.88} & \tabnum{44.66} & \tabnum{72.75} & \tabnum{9.43} & \tabnum{26.52} & \tabnum{53.27} \\
    \mbox{MedPMC-CLIP}~\cite{kim2026medpmc} & \tabnum{41.27} & \tabnum{70.51} & \tabnum{89.33} & \tabnum{36.84} & \tabnum{63.63} & \tabnum{84.71} & \tabnum{55.42} & \tabnum{80.70} & \tabnum{93.03} & \tabnum{56.14} & \tabnum{81.63} & \tabnum{93.35} \\
    Panel2Patch~\cite{yuan2026panel2patch} & \tabnum{42.43} & \tabnum{71.55} & \tabnum{92.02} & \tabnum{36.96} & \tabnum{64.62} & \tabnum{87.45} & \tabnum{47.75} & \tabnum{74.83} & \tabnum{91.61} & \tabnum{49.57} & \tabnum{77.01} & \tabnum{92.84} \\
    \mbox{BMC-LongCLIP}~\cite{sun2025tokenswastedleveraginglong} & \tabnum{44.39} & \tabnum{78.15} & \tabnum{95.83} & \tabnum{35.47} & \tabnum{68.14} & \tabnum{90.77} & \tabnum{67.35} & \tabnum{87.45} & \tabnum{95.83} & \tabnum{67.17} & \tabnum{87.65} & \tabnum{95.90} \\
    \mbox{BMC-CLIP-CF}~\cite{lozano2025biomedica} & \tabnum{47.11} & \tabnum{76.90} & \tabnum{94.85} & \tabnum{38.74} & \tabnum{66.15} & \tabnum{88.41} & \tabnum{52.09} & \tabnum{78.96} & \tabnum{92.94} & \tabnum{51.89} & \tabnum{78.97} & \tabnum{93.25} \\
    BiomedCLIP~\cite{zhang2023biomedclipmultimodalbiomedicalfoundation} & \tabnum{51.65} & \tabnum{84.38} & \tabnum{97.67} & \tabnum{41.09} & \tabnum{75.63} & \tabnum{94.71} & \tabnum{69.80} & \tabnum{89.67} & \tabnum{97.11} & \tabnum{69.11} & \tabnum{89.48} & \tabnum{96.97} \\
    \midrule
    \rowcolor{bioloomlight}
    \methodname{} (Scratch)$^{\ast}$& \tabnum{\underline{61.93}} & \tabnum{\underline{90.61}} & \tabnum{\underline{98.79}} & \tabnum{\underline{52.33}} & \tabnum{\underline{83.86}} & \tabnum{\underline{97.44}} & \tabnum{\underline{75.91}} & \tabnum{\underline{92.94}} & \tabnum{\underline{98.09}} & \tabnum{\underline{74.22}} & \tabnum{\underline{91.52}} & \tabnum{\underline{97.43}} \\
    \rowcolor{bioloomlight}
    \methodname{} (DFN-2B)& \tabnum{\textbf{66.81}} & \tabnum{\textbf{92.81}} & \tabnum{\textbf{99.28}} & \tabnum{\textbf{59.59}} & \tabnum{\textbf{88.81}} & \tabnum{\textbf{98.77}} & \tabnum{\textbf{79.29}} & \tabnum{\textbf{94.57}} & \tabnum{\textbf{98.39}} & \tabnum{\textbf{78.34}} & \tabnum{\textbf{94.10}} & \tabnum{\textbf{98.41}} \\
    \bottomrule
  \end{tabularx}

\ifdefined\jointmaintables\endgroup\else\end{table*}\fi

\end{table*}

\begin{table*}[!t]
  \def\jointretrievalplot{1}
  \noindent\begin{minipage}[t]{\columnwidth}
    \vspace{0pt}
    
\ifdefined\jointretrievalplot\else\begin{table}[t]\fi
  \caption{\textbf{Model comparison on BioMedFlickr retrieval.} Recall@$K$ (\%) is reported for 7,185 image-caption pairs under our revised evaluation protocol. R@$K$ denotes Recall@$K$. The asterisk and score highlighting follow Tab.~\ref{tab:zero-shot}.}
  \vspace{-1.5mm}
  \label{tab:retrieval}
  \centering
  \begingroup
  \resulttablefont

  \renewcommand{\tabnum}[1]{\makebox[\widthof{\normalfont 00.00}][r]{#1}}
  \renewcommand{\arraystretch}{1.05}
  \setlength{\tabcolsep}{1.5pt}
  \begin{tabularx}{\linewidth}{@{}>{\raggedright\arraybackslash}Xrrrrrr@{}}
    \toprule
    & \multicolumn{3}{c}{\textbf{Image $\rightarrow$ Text}} & \multicolumn{3}{c}{\textbf{Text $\rightarrow$ Image}} \\
    \cmidrule(lr){2-4}\cmidrule(lr){5-7}
    \textbf{Model} & \multicolumn{1}{c}{{R@1}} & \multicolumn{1}{c}{{R@10}} & \multicolumn{1}{c}{{R@100}} & \multicolumn{1}{c}{{R@1}} & \multicolumn{1}{c}{{R@10}} & \multicolumn{1}{c}{{R@100}} \\
    \midrule
    CoCa$^{\ast}$~\cite{yu2022cocacontrastivecaptionersimagetext} & \tabnum{1.64} & \tabnum{5.58} & \tabnum{15.30} & \tabnum{2.06} & \tabnum{7.95} & \tabnum{23.98} \\
    OpenCLIP$^{\ast}$~\cite{cherti2023reproduciblescalinglawscontrastive} & \tabnum{2.41} & \tabnum{8.48} & \tabnum{23.42} & \tabnum{3.13} & \tabnum{11.36} & \tabnum{34.25} \\
    \midrule
    \mbox{PMC-CLIP}$^{\ast}$~\cite{lin2023pmcclipcontrastivelanguageimagepretraining} & \tabnum{2.03} & \tabnum{8.38} & \tabnum{25.61} & \tabnum{2.67} & \tabnum{11.75} & \tabnum{34.38} \\
    MedSigLIP~\cite{sellergren2025medgemmatechnicalreport} & \tabnum{3.56} & \tabnum{10.83} & \tabnum{31.45} & \tabnum{4.62} & \tabnum{15.18} & \tabnum{38.78} \\
    BiomedCLIP~\cite{zhang2023biomedclipmultimodalbiomedicalfoundation} & \tabnum{4.76} & \tabnum{14.04} & \tabnum{35.46} & \tabnum{4.68} & \tabnum{19.05} & \tabnum{50.45} \\
    Panel2Patch~\cite{yuan2026panel2patch} & \tabnum{3.60} & \tabnum{12.40} & \tabnum{33.00} & \tabnum{7.00} & \tabnum{19.82} & \tabnum{46.93} \\
    \mbox{BMC-LongCLIP}~\cite{sun2025tokenswastedleveraginglong} & \tabnum{5.00} & \tabnum{15.57} & \tabnum{37.52} & \tabnum{6.67} & \tabnum{20.97} & \tabnum{52.32} \\
    \mbox{BMC-CLIP-CF}~\cite{lozano2025biomedica} & \tabnum{4.86} & \tabnum{14.18} & \tabnum{35.81} & \tabnum{6.99} & \tabnum{22.95} & \tabnum{50.47} \\
    \mbox{MedPMC-CLIP}~\cite{kim2026medpmc} & \tabnum{5.94} & \tabnum{19.94} & \tabnum{45.26} & \tabnum{\underline{9.16}} & \tabnum{\underline{31.61}} & \tabnum{\underline{61.38}} \\
    \midrule
    \rowcolor{bioloomlight}
    \methodname{} (Scratch)$^{\ast}$& \tabnum{\underline{8.00}} & \tabnum{\underline{21.85}} & \tabnum{\underline{45.66}} & \tabnum{8.84} & \tabnum{27.63} & \tabnum{58.75} \\
    \rowcolor{bioloomlight}
    \methodname{} (DFN-2B)& \tabnum{\textbf{9.73}} & \tabnum{\textbf{25.58}} & \tabnum{\textbf{49.46}} & \tabnum{\textbf{11.45}} & \tabnum{\textbf{31.90}} & \tabnum{\textbf{63.51}} \\
    \bottomrule
  \end{tabularx}
  \endgroup
\ifdefined\jointretrievalplot\else\end{table}\fi

  \end{minipage}\hfill
  \begin{minipage}[t]{\columnwidth}
    \vspace{0pt}
    \centering
    \includegraphics[width=\linewidth]{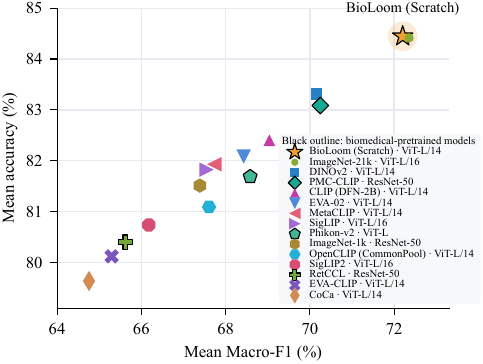}
    \captionsetup{skip=4.5pt,justification=raggedright,singlelinecheck=false}
    \captionof{figure}{\textbf{Linear probing on MedMNIST.} Accuracy and Macro-F1 are averaged equally over 12 datasets. Star: BioLoom; black outlines: biomedical-pretrained models.}
    \label{fig:linear-probe-medmnist}
  \end{minipage}
  \vspace{-2mm}
\end{table*}

\section{Experiments}
\label{sec:experiments}

\subsection{Datasets, evaluation, and implementation}
\label{sec:experimental-setup}
\textbf{Datasets.}
For biomedical transfer evaluation, we assemble 43 public datasets used in BIOMEDICA~\cite{lozano2025biomedica}, MedPMC~\cite{kim2026medpmc}, and other public benchmarks, including evaluation sets released in 2025 and 2026. This zero-shot benchmark covers 67 tasks across 10 biomedical domains. \hyperref[app:zero-shot-classification-benchmark]{Supp.~C} details datasets, targets, and evaluation splits. For broader retrieval evaluation, we construct BioLoomBench from whole figures in papers released in 2026 through bioRxiv~\cite{sever2019biorxiv} and the Europe PMC Open Access subset~\cite{europepmc2015}. It contains 15,000 figures across 15 biomedical domains, with 1,000 per domain, covering biological images, mixed biological images and charts, diagrams, and chart-only figures. It includes single-panel and multi-panel figures, each paired with a concise visual caption and its original caption. \hyperref[app:external-retrieval-benchmark]{Supp.~B} provides details. We do not train on this benchmark. We also evaluate on the publicly available BioMedFlickr test set of 7,185 image-caption pairs~\cite{lozano2025biomedica}.

\textbf{Evaluation protocols.}
Following BIOMEDICA~\cite{lozano2025biomedica}, we average normalized embeddings of two equivalent prompts per class to form a prototype and classify images by cosine similarity. We report accuracy averaged over tasks within datasets, datasets within domains, and the 10 domains equally. \hyperref[app:zero-shot-classification-benchmark]{Supp.~C} details prompts and aggregation; \hyperref[app:secondary-metrics]{Supp.~E} discusses Macro-F1 and one-vs-rest AUC. We evaluate bidirectional retrieval on BioLoomBench and BioMedFlickr using Recall@$K$ for $K\in\{1,10,100\}$. BioLoomBench results are reported separately for concise and original captions. For BioMedFlickr, rows with identical released captions form one relevant set. \hyperref[app:retrieval-protocol]{Supp.~F} defines this revised protocol and reports one-to-one retrieval results. \hyperref[app:bootstrap]{Supp.~D} reports bootstrap confidence intervals.

\textbf{Implementation details.}
We use a ViT-L/14 vision encoder~\cite{dosovitskiy2021image} at $224\times224$ resolution and a pretrained BioClinical ModernBERT text encoder~\cite{sounack2025bioclinicalmodernbert}, with a maximum text length of 512 tokens. BioLoom (Scratch) randomly initializes the vision encoder; BioLoom (DFN-2B) uses DFN-2B ViT-L/14 weights~\cite{fang2024data}. Both use the same pretrained text encoder. Learning rates are $5\times10^{-4}$ for random initialization and $1\times10^{-4}$ for DFN-2B initialization. We train for 10 epochs with the masked contrastive and relation losses on eight H200 GPUs with a global batch size of 1,536. Same-figure pairs are split into groups of at most three and shuffled through a buffer to form batches; whole-figure and panel pairs both count toward this limit. Each GPU computes losses for local anchors against candidates gathered across all GPUs. All anchors contribute to the main loss; only those with semi-positive candidates contribute to the relation loss. We use AdamW with $(\beta_1,\beta_2)=(0.9,0.95)$, 2,000 warm-up steps, cosine decay, and mixed precision. The defaults are 0.05 for the relation loss and 1.1 for same-content-type hard negatives.

\subsection{Main Evaluation}

\textbf{Comparison with state-of-the-art methods.} Tabs.~\ref{tab:zero-shot}, \ref{tab:bioloombench-retrieval}, and \ref{tab:retrieval} summarize zero-shot classification and bidirectional image-text retrieval results. We also evaluate the released Panel2Patch ViT-L/14 checkpoint~\cite{yuan2026panel2patch}, initialized from BMC-CLIP, under the same protocols.

BioLoom (Scratch) achieves 52.89\% domain-balanced zero-shot mean accuracy, exceeding MedPMC-CLIP by 4.12 percentage points. It outperforms the evaluated baselines on BioLoomBench with both caption views: image-to-text (I2T) and text-to-image (T2I) R@1 are 61.93\% and 52.33\% with concise captions, and 75.91\% and 74.22\% with original captions. On BioMedFlickr, its I2T R@1 exceeds MedPMC-CLIP (8.00\% versus 5.94\%), while T2I R@1 remains slightly lower (8.84\% versus 9.16\%). Panel2Patch performs strongly in microscopy and radiology but transfers unevenly across domains. Its focus on localized correspondences may transfer less effectively to BioLoomBench, where caption matching can require biomedical image and chart information.

DFN-2B initialization further improves classification and retrieval. BioLoom (DFN-2B) reaches 56.82\% domain-balanced zero-shot mean accuracy, 3.93 percentage points above Scratch, and exceeds MedPMC-CLIP in all ten domains. BioLoomBench I2T and T2I R@1 reach 66.81\% and 59.59\% with concise captions, with gains also observed for original captions. \hyperref[app:mmbu]{Supp.~H} reports MMBU results. \hyperref[app:panel-retrieval]{Supp.~I} further demonstrates the strong performance of BioLoom in cross-level image-text retrieval. On BioMedFlickr benchmark, I2T and T2I R@1 reach 9.73\% and 11.45\%, respectively.

\Needspace{5\baselineskip}
\textbf{Linear probing.}
To assess visual transferability, we linearly probe frozen vision encoders on the 12 MedMNIST 2D datasets~\cite{yang2023medmnist} (\hyperref[app:linear-probing]{Supp.~G}). Fig.~\ref{fig:linear-probe-medmnist} reports test accuracy and Macro-F1 averaged equally across datasets. BioLoom (Scratch) achieves 84.45\% accuracy and 72.20\% Macro-F1, comparable to ImageNet-21k-pretrained ViT-L/16 (84.42\% and 72.36\%). BioLoom is pretrained for image-text matching, whereas classification pretraining is more closely aligned with linear probing. It also exceeds DINOv2~\cite{oquab2024dinov2} (83.31\% accuracy and 70.15\% Macro-F1). These results suggest that image-text pretraining on Loom52M can learn transferable features from a randomly initialized vision encoder.

\begin{figure*}[t]
  \centering
  \includegraphics[width=\textwidth,trim=0 8bp 0 0,clip]{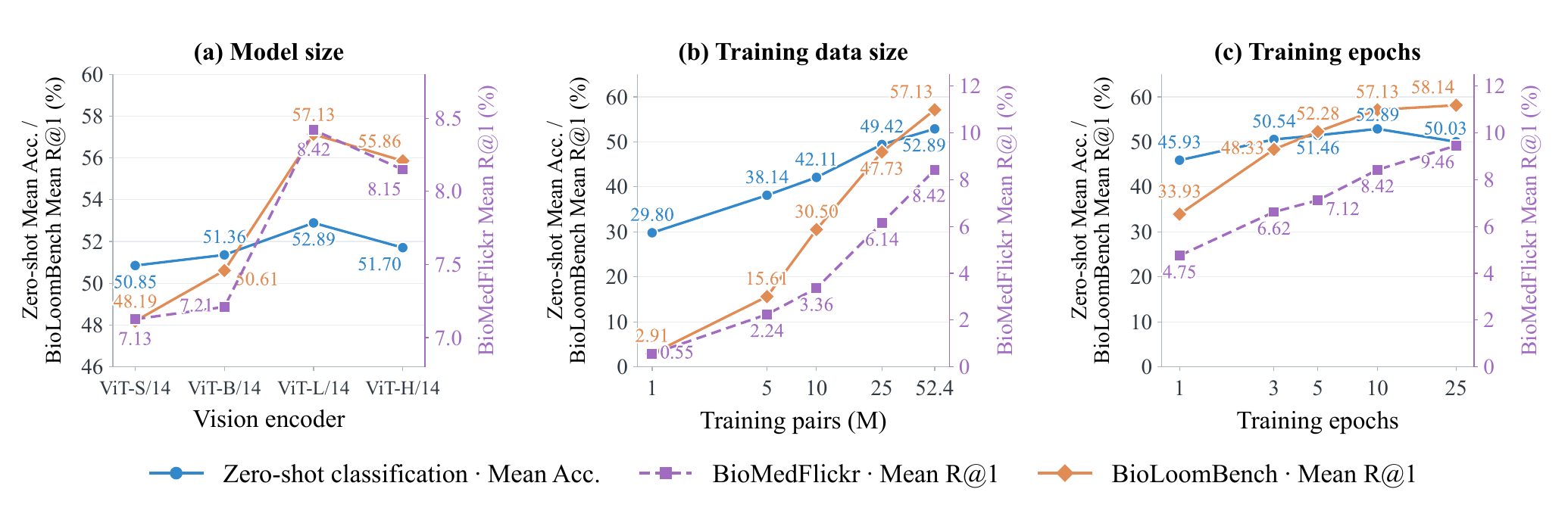}
  \captionsetup{skip=3pt}
  \caption{\textbf{Analysis of (a) model size,  (b) training data size, and (c) training epochs.}}
  \label{fig:scaling}
\end{figure*}

\ifdefined\jointablationtables\begingroup\else\begin{table}[t]\fi
  \centering
  \caption{\textbf{Vision initialization and contrastive-objective ablations.} We report zero-shot accuracy and bidirectional mean R@1 (\%); BioLoomBench uses concise captions. Pair sets follow Sec.~\ref{sec:labels}: semi-positive pairs ($\mathcal{P}_i$) are non-exact pairs within a figure; cross-figure pairs ($\mathcal{A}_i$) span different figures in one article.}
  \vspace{-1mm}
  \label{tab:objective-ablation}
  \resulttablefont
  \setlength{\tabcolsep}{1.1pt}
  \renewcommand{\arraystretch}{1.12}
  \begin{tabularx}{\columnwidth}{@{}>{\raggedright\arraybackslash}Xccc@{}}
    \toprule
    \textbf{Variant} & {\fontsize{7.8}{9.5}\selectfont \shortstack{\textbf{Zero-shot}\\Mean Acc.}} & {\fontsize{7.8}{9.5}\selectfont \shortstack{\textbf{BioLoomBench}\\Mean R@1}} & {\fontsize{7.8}{9.5}\selectfont \shortstack{\textbf{BioMedFlickr}\\Mean R@1}} \\
    \midrule
    \rowcolor{ablationinit}[0pt][0pt]
    \multicolumn{4}{@{}l}{\textit{(a) Vision initialization (ViT-L/14)}} \\
    BioLoom (Scratch) & \tabnum{52.89} & \tabnum{57.13} & \tabnum{8.42} \\
    CommonPool & \tabnum{53.66} & \tabnum{61.53} & \tabnum{10.69} \\
    DINOv2~\cite{oquab2024dinov2} & \tabnum{54.94} & \tabnum{61.20} & \tabnum{10.19} \\
    DFN-2B & \tabnum{56.82} & \tabnum{63.20} & \tabnum{10.59} \\
    \midrule
    \rowcolor{ablationobjective}[0pt][0pt]
    \multicolumn{4}{@{}l}{\textit{(b) Contrastive objective (Scratch)}} \\
    Standard InfoNCE & \tabnum{51.55} & \tabnum{49.71} & \tabnum{6.60} \\
    \addlinespace[2pt]
    \rowcolor{ablationobjective!65!white}[0pt][0pt]
    \multicolumn{4}{@{}l}{\textit{Exact-pair alignment with related-pair masking ($\mathcal{L}_{\mathrm{mask}}$)}} \\
    \mbox{w/o semi-positive mask~($\mathcal{P}_i$)} & \tabnum{52.48} & \tabnum{53.91} & \tabnum{7.52} \\
    \mbox{w/o cross-figure mask~($\mathcal{A}_i$)} & \tabnum{53.31} & \tabnum{54.21} & \tabnum{7.34} \\
    \mbox{w/o both masks~($\mathcal{P}_i\cup\mathcal{A}_i$)} & \tabnum{52.19} & \tabnum{52.44} & \tabnum{6.98} \\
    \addlinespace[2pt]
    \rowcolor{ablationobjective!65!white}[0pt][0pt]
    \multicolumn{4}{@{}l}{\textit{Semi-positive alignment ($\mathcal{L}_{\mathrm{rel}}$)}} \\
    w/o relation loss & \tabnum{52.24} & \tabnum{51.68} & \tabnum{7.24} \\
    w/ exact pairs ($\mathcal{E}_i$) & \tabnum{54.41} & \tabnum{52.79} & \tabnum{7.30} \\
    \mbox{w/o cross-figure mask~($\mathcal{A}_i$)} & \tabnum{49.85} & \tabnum{54.53} & \tabnum{7.90} \\
    \mbox{w/o hard-neg. emphasis} & \tabnum{50.71} & \tabnum{55.48} & \tabnum{7.93} \\
    \bottomrule
  \end{tabularx}
\ifdefined\jointablationtables\endgroup\else

  \vspace{-10pt}
\end{table}\fi

\textbf{Impact of model size.} With randomly initialized vision encoders, scaling from ViT-S to ViT-L (Fig.~\ref{fig:scaling}(a)) improves domain-balanced zero-shot mean accuracy from 50.85\% to 52.89\% and bidirectional mean R@1 from 48.19\% to 57.13\% on BioLoomBench and from 7.13\% to 8.42\% on BioMedFlickr. ViT-H scores slightly lower on all three metrics, possibly because larger vision transformers require more data to benefit from their capacity~\cite{dosovitskiy2021image,zhai2022scaling}. Future work will explore larger datasets for ViT-H.

\textbf{Impact of data scale and training epochs.} Figs.~\ref{fig:scaling}(b) and (c) examine data size and training epochs, respectively. Increasing data from 1M to 52.4M pairs consistently improves all three metrics. Retrieval improves up to 25 epochs, while zero-shot accuracy peaks at 10 epochs and then declines, suggesting a trade-off between image-text alignment and category recognition. We use 10 epochs to balance classification, retrieval, and training cost.

\ablationtitle{ablationinit}{Vision encoder initialization.} Tab.~\ref{tab:objective-ablation}~(a) compares four ViT-L/14 initializations. All three pretrained variants outperform Scratch across all three metrics. DFN-2B achieves the highest zero-shot accuracy (56.82\%) and BioLoomBench mean R@1 (63.20\%), while CommonPool performs best on BioMedFlickr (10.69\%).

\ablationtitle{ablationobjective}{Effectiveness of the proposed contrastive objective.} Tab.~\ref{tab:objective-ablation}~(b) compares objectives with a randomly initialized vision encoder. The full objective exceeds standard InfoNCE in zero-shot accuracy (52.89\% versus 51.55\%) and bidirectional mean R@1 on BioLoomBench (57.13\% versus 49.71\%) and BioMedFlickr (8.42\% versus 6.60\%). It leads the reported variants in retrieval, although two have higher zero-shot accuracy.

Removing the main-loss cross-figure mask raises zero-shot accuracy to 53.31\% but lowers retrieval to 54.21\% and 7.34\%, respectively. Using within-article cross-figure pairs as negatives may help distinguish visual content while separating semantically related pairs. Adding exact pairs to the relation loss raises accuracy to 54.41\% but lowers retrieval to 52.79\% and 7.30\%, suggesting a classification--retrieval trade-off. Removing either or both main-loss masks lowers retrieval on both benchmarks. Removing the relation loss, its cross-figure mask, or hard-negative upweighting lowers all three metrics relative to the full model. \hyperref[app:hyperparameter-sensitivity]{Supp.~J} examines relation-loss and hard-negative weight sensitivity.

\subsection{Further Analysis}

\ifdefined\jointablationtables\begingroup\else\begin{table}[!t]\fi
  \centering
  \caption{\textbf{Ablations of the data pipeline.} We report zero-shot classification accuracy and bidirectional mean R@1 (\%); BioLoomBench uses concise captions. \textit{w/o all charts} removes chart panels and chart-containing whole figures, retaining non-chart panels. \textit{w/ pure-chart views} replaces part of the 25M mixture with figures composed entirely of charts and their panels.}
  \vspace{-1mm}
  \label{tab:ablation}
  \resulttablefont
  \setlength{\tabcolsep}{1.9pt}
  \renewcommand{\arraystretch}{1.12}
  \begin{tabularx}{\columnwidth}{@{}>{\raggedright\arraybackslash}Xccc@{}}
    \toprule
    \textbf{Variant} & \shortstack{\textbf{Zero-shot}\\Mean Acc.} & \shortstack{\textbf{BioLoomBench}\\Mean R@1} & \shortstack{\textbf{BioMedFlickr}\\Mean R@1} \\
    \midrule
    BioLoom (25M pairs) & \tabnum{49.42} & \tabnum{47.73} & \tabnum{6.14} \\
    \midrule
    \rowcolor{ablationcaption}[0pt][0pt]
    \multicolumn{4}{@{}l}{\textit{(a) Caption content}} \\
    Visual description only & \tabnum{49.30} & \tabnum{41.14} & \tabnum{5.81} \\
    Contextual caption only & \tabnum{49.41} & \tabnum{43.76} & \tabnum{6.08} \\
    \midrule
    \rowcolor{ablationviews}[0pt][0pt]
    \multicolumn{4}{@{}l}{\textit{(b) Image views}} \\
    Panel only & \tabnum{40.22} & \tabnum{31.14} & \tabnum{2.57} \\
    Whole figure only & \tabnum{32.49} & \tabnum{15.61} & \tabnum{1.54} \\
    \midrule
    \rowcolor{ablationcharts}[0pt][0pt]
    \multicolumn{4}{@{}l}{\textit{(c) Chart inclusion}} \\
    \mbox{w/o all charts} & \tabnum{47.59} & \tabnum{37.99} & \tabnum{5.83} \\

    \mbox{w/ pure-chart views} & \tabnum{44.63} & \tabnum{45.13} & \tabnum{4.57} \\
    \bottomrule
  \end{tabularx}
  \vspace{-2mm}
\ifdefined\jointablationtables\endgroup\else\end{table}\fi

We compare data configurations using 25M pairs.

\noindent\ablationtitle{ablationcaption}{Caption content.} We compare contextual captions, visual descriptions, and their combination for panels, retaining original whole-figure captions. The full setting leads on all three metrics. Combining both text fields yields 47.73\% BioLoomBench mean R@1, versus 41.14\% for visual descriptions and 43.76\% for contextual captions alone, suggesting complementary information for image-text alignment.

\noindent\ablationtitle{ablationviews}{Image views.} We compare panel-only, multi-panel whole-only, and mixed-view training. Whole-only training uses original captions and omits the relation loss because no sibling-panel pairs are available. The full setting scores highest on all three metrics: mixed views yield 49.42\% zero-shot accuracy and 47.73\% BioLoomBench mean R@1, versus 40.22\% and 31.14\% for panel-only, and 32.49\% and 15.61\% for whole-only training. This suggests that whole-figure context and panel details complement each other in image-text learning.

\noindent\ablationtitle{ablationcharts}{Chart inclusion.} Our full setting removes pure-chart figures and their panels while retaining charts in mixed biomedical figures. We compare removing all chart content and restoring pure-chart figures and panels under the same sampling budget. The full setting scores highest on all three metrics. Removing all charts or restoring pure-chart views lowers BioLoomBench mean R@1 from 47.73\% to 37.99\% and 45.13\%, respectively, and BioMedFlickr from 6.14\% to 5.83\% and 4.57\%. Both also lower zero-shot accuracy, suggesting that charts are useful within mixed biomedical figures, but adding pure-chart views can hurt performance.

\section{Conclusion}
\label{sec:conclusion}
BioLoom is a CLIP-style foundation model for the biomedical domain. We use the structure of scientific articles in two ways. First, we curate Loom52M, a large-scale biomedical image-text dataset. It combines biomedical images and charts, links whole figures with their panels, and enriches panel descriptions using visual content, original captions, and relevant article paragraphs. Second, we use semi-positive pairs, including a panel image paired with the caption of a sibling panel. We combine related-pair masking, semi-positive supervision, and hard-negative weighting to learn both exact and semi-positive alignment. BioLoom (DFN-2B) achieves the highest domain-balanced mean accuracy and overall bidirectional mean R@1 on both retrieval benchmarks among the compared baselines. In future work, we will further scale up BioLoom and develop a generative vision-language model using its vision encoder.


{\small

}

\clearpage
\appendix
\twocolumn[{%
  \begin{center}
    {\Large\bfseries \methodname: Scaling Biomedical Vision-Language Models by\\
    Learning from the Structure of Scientific Articles\par}
    \vspace{5pt}
    {\large Supplementary Material\par}
    \vspace{10pt}
    {\large\textbf{Contents}\par}
    \vspace{6pt}
    \begin{minipage}[t]{0.485\textwidth}
      \raggedright\fontsize{10.5}{13}\selectfont
      \setlength{\parskip}{4pt}
      \hyperref[app:figure-annotation]{A. Training-Data Annotation and Metadata}\dotfill\hyperref[app:figure-annotation]{\pageref*{app:figure-annotation}}\par
      \hyperref[app:external-retrieval-benchmark]{B. BioLoomBench: Our Retrieval Benchmark}\dotfill\hyperref[app:external-retrieval-benchmark]{\pageref*{app:external-retrieval-benchmark}}\par
      \hyperref[app:zero-shot-classification-benchmark]{C. Zero-shot Classification Benchmark}\dotfill\hyperref[app:zero-shot-classification-benchmark]{\pageref*{app:zero-shot-classification-benchmark}}\par
      \hyperref[app:bootstrap]{D. Bootstrap Confidence Intervals}\dotfill\hyperref[app:bootstrap]{\pageref*{app:bootstrap}}\par
      \hyperref[app:secondary-metrics]{E. Secondary Zero-shot Metrics}\dotfill\hyperref[app:secondary-metrics]{\pageref*{app:secondary-metrics}}\par
      \hyperref[app:retrieval-protocol]{F. Retrieval Protocol}\dotfill\hyperref[app:retrieval-protocol]{\pageref*{app:retrieval-protocol}}\par
    \end{minipage}\hfill
    \begin{minipage}[t]{0.485\textwidth}
      \raggedright\fontsize{10.5}{13}\selectfont
      \setlength{\parskip}{4pt}
      \hyperref[app:linear-probing]{G. Linear Probing Across Biomedical Datasets}\dotfill\hyperref[app:linear-probing]{\pageref*{app:linear-probing}}\par
      \hyperref[app:mmbu]{H. MMBU Question-Option Evaluation}\dotfill\hyperref[app:mmbu]{\pageref*{app:mmbu}}\par
      \hyperref[app:panel-retrieval]{I. Fine-grained and Cross-level Retrieval}\dotfill\hyperref[app:panel-retrieval]{\pageref*{app:panel-retrieval}}\par
      \hyperref[app:hyperparameter-sensitivity]{J. Hyperparameter Sensitivity}\dotfill\hyperref[app:hyperparameter-sensitivity]{\pageref*{app:hyperparameter-sensitivity}}\par
      \hyperref[app:computational-cost]{K. Computational Cost}\dotfill\hyperref[app:computational-cost]{\pageref*{app:computational-cost}}\par
      \hyperref[app:discussion]{L. Discussion}\dotfill\hyperref[app:discussion]{\pageref*{app:discussion}}\par
    \end{minipage}
    \vspace{8pt}
  \end{center}
}]
\setlength{\textfloatsep}{8pt plus 2pt minus 2pt}
\setlength{\dbltextfloatsep}{8pt plus 2pt minus 2pt}
\setlength{\floatsep}{6pt plus 2pt minus 2pt}

\section{Training-Data Annotation and Metadata}
\label{app:figure-annotation}

This section describes the figure annotations used to construct the training dataset. The separate retrieval benchmark is described in Sec.~\ref{app:external-retrieval-benchmark}.

\paragraph{Source records and provenance.}
We start from the PMC-derived figure records released through
BIOMEDICA~\citep{lozano2025biomedica}. BIOMEDICA associates each figure with
its source identifiers, original figure caption, article metadata, and textual context. Our annotation stage reads four fields from these records:
\texttt{metadata.image\_id}, \texttt{image.source\_member},
\texttt{metadata.caption}, and \texttt{evidence.mention\_paragraphs}. The first
two locate the figure and its source image. The latter two provide the original caption and the article paragraphs that discuss the figure. We use at most the first two paragraphs that mention the figure. Figure and article identifiers are retained for the whole image and all panel crops.

BIOMEDICA also provides global and local concept annotations at the figure level. We do not treat those concepts as panel labels. Instead, we add the panel-level schema below because one biomedical figure can contain several visual forms, such as a microscopy image, a statistical plot, and a schematic.
Thus, BIOMEDICA supplies the source records and provenance, while our schema supplies the panel boundaries and panel-level visual labels.

The visual description and contextual caption in the main text correspond to the \texttt{data\_description} and \texttt{caption} fields in our annotation schema, respectively.

\paragraph{Complete annotation prompt.}
We send the full figure and the following prompt to
Gemini-3.5-Flash-Lite. The two bracketed fields are filled directly from the source metadata. The text below follows the production prompt.

\begin{figure*}[tbh]
\centering
\begin{minipage}{\textwidth}
\begin{judgepromptwide}{Figure-to-panel annotation prompt}

Segment this biomedical figure into its labeled sub-panels (e.g., A/B/C/D),
or a single region if uniform.

\medskip
\textbf{Figure caption:}

\texttt{<ORIGINAL-FIGURE-CAPTION>}

\medskip
\textbf{Passages from the article body that discuss this figure:}

\texttt{<UP-TO-TWO-MENTION-PARAGRAPHS>}

\medskip
For each panel, give:

\begin{itemize}
  \setlength{\itemsep}{1pt}
  \setlength{\topsep}{2pt}
  \item \texttt{region\_id};
  \item \texttt{content\_type} and \texttt{modality}: pick the closest match
        from the fixed lists in the \mbox{schema; do not invent new values};
  \item \texttt{box\_2d}: a tight box in the order
        $[y_{\min},x_{\min},y_{\max},x_{\max}]$, with every coordinate an
        integer normalized to $[0,1000]$ relative to the full figure;
  \item \texttt{layout}: if the panel contains an internal grid of plots or
        images, return \texttt{\{"rows": r, "cols": c\}}; otherwise return
        \texttt{\{"rows": 1, "cols": 1\}};
  \item \texttt{data\_description}: one or two sentences describing the data
        itself using only \mbox{visible evidence, without article context};
  \item \texttt{caption}: one or two sentences combining the panel's visual
        content with only the relevant information from the original figure
        caption and article passages.
\end{itemize}

The \texttt{data\_description} and \texttt{caption} must not overlap in
content. Do not copy source sentences verbatim; synthesize them. Return only
the schema JSON.

\end{judgepromptwide}
\end{minipage}
\end{figure*}

\paragraph{Structured response.}
The request uses a fixed response schema. All seven region fields are required,
and \texttt{box\_2d} must contain exactly four integers. Its logical structure
is:

\begin{figure*}[!t]
\centering
\begin{minipage}{\textwidth}
\begin{judgepromptwide}{Structured response for figure annotation}

\texttt{\{}\\
\hspace*{1em}\texttt{"regions": [}\\
\hspace*{2em}\texttt{\{}\\
\hspace*{3em}\texttt{"region\_id": "<PANEL-ID>",}\\
\hspace*{3em}\texttt{"content\_type": "<CONTENT-TYPE>",}\\
\hspace*{3em}\texttt{"modality": "<VISUAL-MODALITY>",}\\
\hspace*{3em}\texttt{"box\_2d": [<YMIN>, <XMIN>, <YMAX>, <XMAX>],}\\
\hspace*{3em}\texttt{"layout": \{"rows": <INTEGER>, "cols": <INTEGER>\},}\\
\hspace*{3em}\texttt{"data\_description": "<VISIBLE-FACTS>",}\\
\hspace*{3em}\texttt{"caption": "<PANEL-CAPTION>"}\\
\hspace*{2em}\texttt{\}}\\
\hspace*{1em}\texttt{]}\\
\texttt{\}}

\end{judgepromptwide}
\end{minipage}
\end{figure*}

The response contains one object for every detected panel. A visually uniform
figure has one region. Coordinates always refer to the full input figure, not
to a resized panel: $(0,0)$ is the upper-left corner and $(1000,1000)$ is the
lower-right corner. The box is converted back to pixel coordinates before the
panel crop is extracted. The original figure, every crop, and their captions
share \mbox{figure and article identifiers}.

\paragraph{Panel-level visual taxonomy.}
We use a fixed, task-oriented taxonomy for consistent annotation and data
sampling. It is designed to describe the visual forms found in BIOMEDICA
figures, but it is not a copy of the BIOMEDICA concept labels and is not meant
to be an exhaustive biomedical ontology. We use two labels because they
describe different levels of detail. \texttt{content\_type} (Tab.~\ref{tab:content-type-definitions}) answers what broad
kind of visual a panel is, while \texttt{modality} identifies its specific
visual, acquisition, or rendering subtype. For example, a survival curve is
\texttt{chart}/\texttt{survival\_curve}, a fluorescence image is
\texttt{microscopy}/\texttt{fluorescence\_microscopy}, and a CT slice is
\texttt{radiograph}/\texttt{ct\_scan}. The first label supports coarse
grouping; the second \mbox{captures fine visual distinctions}.

Each region receives exactly one value from each vocabulary. When a region is
compound, the annotator selects the closest value from the fixed lists and
records its internal arrangement in \texttt{layout}. The \texttt{other} value
prevents an unsupported visual form from being forced into an unrelated class.
The response schema rejects labels outside these lists. The modality headings
below organize the presentation; the production schema has a single
73-value list.

\begin{figure*}[!t]
\centering
\begin{minipage}{\textwidth}
\begin{judgepromptwide}{Fixed content-type vocabulary}
\small
\texttt{chart}, \texttt{microscopy}, \texttt{blot\_gel},
\texttt{schematic}, \texttt{photo}, \texttt{chemical\_structure},
\texttt{table}, \texttt{map}, \texttt{radiograph}, \texttt{simulation},
and \texttt{other}.
\end{judgepromptwide}
\end{minipage}
\end{figure*}

\begin{table*}[t]
\centering
\caption{\textbf{Definition of the 11 panel-level content types.} The content type
describes the broad visual form; modality specifies the subtype.}
\label{tab:content-type-definitions}
\resulttablefont
\setlength{\tabcolsep}{4pt}
\begin{tabularx}{\textwidth}{@{}>{\ttfamily}l X l@{}}
\toprule
\textbf{Content type} & \textbf{Operational definition} & \textbf{Examples}\\
\midrule
chart & Quantitative values encoded with axes, marks, curves, or statistical summaries. & line chart, heatmap\\
microscopy & Cellular, tissue, or material images acquired with a microscope. & fluorescence, histology\\
blot\_gel & Band-based molecular assay images. & western blot, gel\\
schematic & Explanatory diagrams of a process, mechanism, anatomy, or workflow. & pathway, flow diagram\\
photo & Camera-like photographs of patients, specimens, instruments, or procedures. & clinical photo, endoscopy\\
chemical\_structure & Symbolic or rendered molecular and crystal structures. & molecule, crystal\\
table & Information arranged primarily in rows and columns. & result table\\
map & Spatial or geographic information organized by location. & choropleth, geographic map\\
radiograph & Clinical imaging and physiological recordings. & X-ray, CT, ECG\\
simulation & Images produced by a computational model or simulation. & molecular dynamics\\
other & A visual form not represented by the ten named types. & unsupported form\\
\bottomrule
\end{tabularx}
\end{table*}

\begin{figure*}[!t]
\centering
\begin{minipage}{\textwidth}
\begin{judgepromptwide}{Fixed modality vocabulary}
\small
\textbf{Charts:}
\texttt{bar\_chart}, \texttt{grouped\_bar\_chart},
\texttt{stacked\_bar\_chart}, \texttt{line\_chart},
\texttt{scatter\_plot}, \texttt{box\_plot}, \texttt{violin\_plot},
\texttt{heatmap}, \texttt{grid\_of\_heatmaps}, \texttt{pie\_chart},
\texttt{density\_plot}, \texttt{histogram},
\texttt{histogram\_overlay}, \texttt{survival\_curve},
\texttt{volcano\_plot}, \texttt{forest\_plot}, \texttt{dot\_plot},
\texttt{radar\_chart}, \texttt{venn\_diagram},
\texttt{network\_graph}, \texttt{dendrogram},
\texttt{flow\_cytometry\_plot}, \texttt{raster\_plot},
\texttt{psth\_plot}, \texttt{spectrum\_plot}, and
\texttt{chromatogram}.

\medskip
\textbf{Microscopy and tissue:}
\texttt{fluorescence\_microscopy}, \texttt{confocal\_microscopy},
\texttt{immunofluorescence}, \texttt{brightfield\_microscopy},
\texttt{phase\_contrast\_microscopy},
\texttt{electron\_microscopy\_tem},
\texttt{electron\_microscopy\_sem},
\texttt{atomic\_force\_microscopy},
\texttt{scanning\_tunneling\_microscopy},
\texttt{live\_cell\_imaging},
\texttt{optical\_coherence\_tomography}, \texttt{histology\_stain},
\texttt{immunohistochemistry}, and \texttt{in\_situ\_hybridization}.

\medskip
\textbf{Blots and gels:}
\texttt{western\_blot}, \texttt{northern\_blot},
\texttt{southern\_blot}, and \texttt{gel\_electrophoresis}.

\medskip
\textbf{Medical imaging and physiological signals:}
\texttt{x\_ray}, \texttt{ct\_scan}, \texttt{mri}, \texttt{pet\_scan},
\texttt{spect\_scan}, \texttt{ultrasound}, \texttt{mammography},
\texttt{angiography}, \texttt{endoscopy}, \texttt{fluoroscopy},
\texttt{dermoscopy}, \texttt{ecg\_trace}, \texttt{eeg\_trace},
\mbox{\texttt{meg\_trace}, and \texttt{emg\_trace}}.

\medskip
\textbf{Structures and simulations:}
\texttt{3d\_structural\_model}, \texttt{molecular\_structure},
\texttt{crystal\_structure}, and
\texttt{molecular\_dynamics\_simulation}.

\medskip
\textbf{Diagrams, maps, and photographs:}
\texttt{pathway\_diagram}, \texttt{flow\_diagram},
\texttt{workflow\_diagram}, \texttt{anatomical\_illustration},
\texttt{phylogenetic\_tree}, \texttt{choropleth\_map},
\texttt{geographic\_map}, \texttt{clinical\_photo},
\texttt{illustration}, and \texttt{other}.
\end{judgepromptwide}
\end{minipage}
\end{figure*}

\paragraph{Use in training.}
The annotation record is richer than the input used by the current model.
For a panel sample, \texttt{box\_2d} is used once to extract the crop, and
\texttt{data\_description} and \texttt{caption} are concatenated to form its
training text. A whole-figure sample uses the original BIOMEDICA figure
caption. Standard CLIP training receives only the resulting image and text.
Our relation loss additionally uses sample, figure, and article identifiers,
\texttt{content\_type}, and whether the sample is a whole figure or a panel.
These fields define exact pairs, sibling panels, same-article masks, and
hard negatives of the same content type. The remaining annotations are retained as metadata
rather than used as direct optimization targets.

\section{BioLoomBench: Our Retrieval Benchmark}
\label{app:external-retrieval-benchmark}

\paragraph{Evaluation data and separation from training data.}
We constructed BioLoomBench to evaluate retrieval independently of our training data and across a broader range of biomedical figure types. Existing biomedical retrieval sets largely center on biological images; BioLoomBench also includes image-chart combinations, diagrams, and chart-only figures. To distinguish it from BIOMEDICA, published in 2025~\citep{lozano2025biomedica}, we collected papers published or posted in 2026 from bioRxiv and Europe PMC Open Access~\citep{sever2019biorxiv,europepmc2015}. The benchmark contains 15,000 whole figures from 6,990 articles: 4,967 figures from bioRxiv and 10,033 from Europe PMC Open Access.

\begin{figure*}[t]
\centering
\begin{minipage}[t]{0.465\textwidth}
\vspace{0pt}
\captionof{table}{\textbf{Composition of BioLoomBench.} Counts and shares are shown for all 15,000 whole figures.}
\label{tab:bioloombench-overview}
\resulttablefont
\setlength{\tabcolsep}{4pt}
\begin{tabular}{@{}lrr@{}}
\toprule
\textbf{Group} & \textbf{Count} & \textbf{Share (\%)}\\
\midrule
Figures & 15,000 & \tabnum{100.00}\\
Source articles & 6,990 & -\\
\midrule
\multicolumn{3}{@{}l}{\emph{Source}}\\
bioRxiv & 4,967 & \tabnum{33.11}\\
Europe PMC Open Access & 10,033 & \tabnum{66.89}\\
\midrule
\multicolumn{3}{@{}l}{\emph{Visual content}}\\
Biological image only & 5,500 & \tabnum{36.67}\\
Biological image and chart & 5,500 & \tabnum{36.67}\\
Diagram or schematic & 2,500 & \tabnum{16.67}\\
Chart only & 1,500 & \tabnum{10.00}\\
\midrule
\multicolumn{3}{@{}l}{\emph{Whole-figure layout}}\\
Multi-panel figures & 10,800 & \tabnum{72.00}\\
Single-panel figures & 4,200 & \tabnum{28.00}\\
\midrule
Mean / median pixel count & 1.24 / 0.82 MP & -\\
Mean image width $\times$ height & $1{,}188\times922$ px & -\\
\bottomrule
\end{tabular}
\end{minipage}\hfill
\begin{minipage}[t]{0.505\textwidth}
\vspace{0pt}
\centering
\captionsetup{skip=6pt}
\captionof{figure}{\textbf{BioLoomBench composition.} Rings show counts and shares by visual content, layout, and source; domains appear below.}
\includegraphics[width=\linewidth]{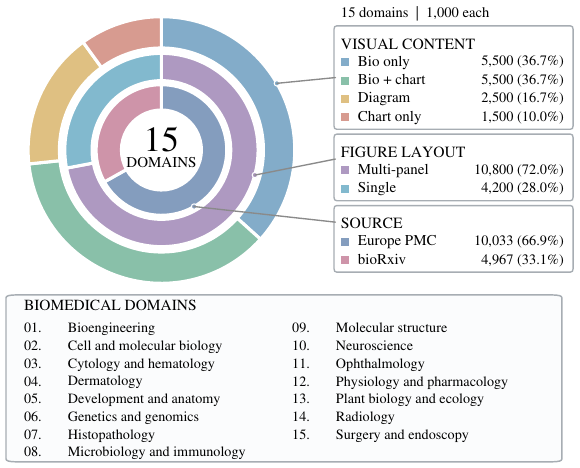}
\label{fig:bioloombench-overview}
\end{minipage}
\end{figure*}

\paragraph{Construction procedure.}
We collected eligible papers from the two sources and extracted whole figures with their original captions and article identifiers. This whole-figure benchmark uses uncropped figures. We assigned each figure a biomedical domain, a visual-content category, and a whole-figure layout label. We then selected at most four figures per article and exactly 1,000 figures per domain. This yields balanced domain coverage while retaining variation in visual content and figure layout. Tab.~\ref{tab:bioloombench-overview} and Fig.~\ref{fig:bioloombench-overview} summarize the benchmark.

Each of the 15 domains contributes exactly 1,000 figures; their names are listed below the chart. The released image files have a median long edge of 1,119 pixels (10th-90th percentile: 708-1,895 pixels). Pixel count is width multiplied by height; MP denotes one million pixels.

\paragraph{Caption views.}
Each figure has two retrieval texts. We generated concise captions with Gemini-3.5-Flash-Lite and manually checked a sample against the corresponding figures for quality. The original-caption view uses the caption from the original paper; we remove a leading figure-number label when detected. Concise captions contain 17.67 words on average (median 18; range 8-25). Original captions contain 94.50 words on average (median 73; range 6-397).

We evaluate retrieval under three text settings: concise captions encoded at the native context length of each model, original captions encoded at the native context length of the model, and original captions with a common maximum input length of 77 tokens. If the native context of a model is shorter than 77 tokens, we use its native limit. For image-to-text (I2T), a figure is the query and captions are ranked; for text-to-image (T2I), a caption is the query and figures are ranked. We report both directions for each text setting.

\section{Zero-shot classification benchmark}
\label{app:zero-shot-classification-benchmark}

\paragraph{Construction.}
We assemble 43 public datasets into a benchmark of 67 classification tasks across 10 biomedical domains. The collection includes datasets used in BIOMEDICA and MedPMC~\citep{lozano2025biomedica,kim2026medpmc}, alongside public datasets covering cellular phenotypes, cytology, tissue pathology, medical imaging, and microscopic organisms. A dataset may define several prediction targets or imaging views; these contribute separate tasks within that dataset. The three OrganMNIST 2D views form one dataset group, as do the three LIDS parasite tasks and the two bacterial classification tasks. We count each group once among the 43 datasets. Tab.~\ref{tab:task-mapping} lists datasets, targets, and counts.

\paragraph{Evaluation images and splits.}
We use the specified evaluation partitions and labels provided with each dataset. COOS-7 contributes four test conditions, averaged as one subcellular-localization task. LIDS supplies three test partitions for each parasite task; we evaluate each partition and average its metrics. Raabin-WBC combines Test A and Test B. PAPILA uses the union of the test eyes in its five Test 1 folds, and HyperKvasir uses the labeled static images in its two published folds. MedIMeta-AML uses the MedIMeta test partition. Omnipose provides a modality-classification task derived from its phase-contrast and fluorescence test images. For Bacteria2026, we follow the authors' code to obtain whole-image morphology and Gram labels and reproduce its image-level split. The retained chest-imaging tasks use CheXpert validation images and the labeled RSNA Stage 2 training release for evaluation. We freeze encoders and fit no classifier on evaluation images.

\paragraph{Prompts and prediction.}
Each candidate label is represented by two semantically equivalent prompts. We normalize their text features, average them, and normalize the resulting class feature. Each image is encoded once and assigned to the candidate with the highest cosine similarity.

\paragraph{Metrics and aggregation.}
Let $N$ and $C$ denote the numbers of images and classes, respectively. Let $s_{ic}$ denote the score of image $i$ for class $c$, and $P_c$ and $R_c$ denote class precision and recall. For AUC, $\mathcal{P}_c$ and $\mathcal{N}_c$ are the positive and negative images, and $\mathcal{C}'$ contains classes that have both. The metrics and aggregation are defined below:
\begingroup
\allowdisplaybreaks[4]
\begin{align}
\operatorname{Acc}&=\frac{1}{N}\sum_i\mathbb{1}[\arg\max_c s_{ic}=y_i],\\[-2pt]
\operatorname{Acc}_{\mathrm{multi}}&=\frac{1}{NC}\sum_{i,c}\mathbb{1}[\hat y_{ic}=y_{ic}],\\[-2pt]
\operatorname{MacroF1}&=\frac{1}{C}\sum_c\frac{2P_cR_c}{P_c+R_c},\\[-2pt]
\operatorname{AUC}_c&=\frac{1}{|\mathcal P_c||\mathcal N_c|}
\sum_{i\in\mathcal P_c}\sum_{j\in\mathcal N_c}\phi(s_{ic},s_{jc}),\\[-2pt]
\phi(a,b)&=\mathbb{1}[a>b]+\tfrac12\mathbb{1}[a=b],\\[-2pt]
\operatorname{AUC}&=\frac{1}{|\mathcal C'|}\sum_{c\in\mathcal C'}\operatorname{AUC}_c,\\[-2pt]
m_d&=\frac{1}{|\mathcal T_d|}\sum_{t\in\mathcal T_d}m_t,\quad
m_g=\frac{1}{|\mathcal D_g|}\sum_{d\in\mathcal D_g}m_d,\\[-2pt]
m_{\mathrm{overall}}&=\frac{1}{10}\sum_{g=1}^{10}m_g.
\end{align}
\endgroup
For Macro-F1, a class with zero precision and recall receives zero. For tasks with image-specific candidate sets, Macro-F1 uses the semantic labels and AUC pools candidate probabilities across images. The last line applies to accuracy, Macro-F1, and AUC: tasks are averaged within a dataset, datasets within a domain, and the ten domains equally. Tabs.~\ref{tab:zero-shot-f1} and~\ref{tab:zero-shot-auc} report Macro-F1 and AUC.

\makeatletter
\setlength{\@dblfptop}{0pt}
\setlength{\@dblfpbot}{0pt plus 1fil}
\makeatother
\begingroup
\small
\setlength{\tabcolsep}{3pt}
\renewcommand{\arraystretch}{0.96}
\begin{table*}[p]
\caption{\textbf{Datasets and tasks in the zero-shot classification benchmark.} The benchmark includes 43 datasets and 67 tasks across 10 domains. Year denotes publication or dataset release. Counts give images per task or test condition; LIDS counts cover three test partitions.}\label{tab:task-mapping}
\centering
\resulttablefont
\begin{tabularx}{\textwidth}{@{}>{\raggedright\arraybackslash}p{0.13\textwidth}>{\raggedright\arraybackslash}p{0.215\textwidth}>{\centering\arraybackslash}p{0.055\textwidth}>{\raggedright\arraybackslash}X>{\centering\arraybackslash}p{0.05\textwidth}>{\raggedright\arraybackslash}p{0.18\textwidth}@{}}
\toprule
\rowcolor{gray!15}\textbf{Domain} & \textbf{Dataset} & \textbf{Year} & \textbf{Prediction target(s)} & \textbf{Tasks} & \textbf{Images used}\\
\midrule
Cellular phenotyping & Burgess et al.~\cite{burgess2024orientation} & 2024 & Contour; texture; eccentricity & 3 & 600 per task\\
& Eulenberg et al.~\cite{eulenberg2017reconstructing} & 2017 & Cell cycle: brightfield, darkfield, epifluorescence & 3 & 743 per task\\
 & COOS-7~\cite{coos2019} & 2019 & Subcellular localization & 1 & 10,364; 17,021; 32,596; 30,772\\
\midrule
Structure & Nirschl fluorescence~\cite{lozano2024microbench} & 2024 & Subcellular structure & 1 & 934\\
& EMPIAR SBF-SEM~\cite{iudin2023empiar} & 2023 & Ultrastructure & 1 & 577\\
& POLLEN13K~\cite{battiato2020pollen13k} & 2020 & Pollen classification & 1 & 700\\
\midrule
Microscopy composition & Micro-Bench~\cite{lozano2024microbench} & 2024 & Domain; modality; stain; submodality & 4 & 17,315; 17,235; 15,196; 17,235\\
 & Omnipose~\cite{omnipose2022} & 2022 & Phase-contrast versus fluorescence & 1 & 223\\
\midrule
Cytology & Acevedo et al.~\cite{acevedo2020bloodcells} & 2020 & Blood-cell type & 1 & 1,600\\
& Jung et al.~\cite{jung2022wbc} & 2022 & Blood-cell type & 1 & 1,000\\
& Hussain et al.~\cite{hussain2020papsmear} & 2020 & Cervical-cell type & 1 & 193\\
 & TissueMNIST~\cite{yang2023medmnist} & 2023 & Kidney cell type & 1 & 47,280\\
 & KU-Optofil PBC~\cite{kuoptofil2026,kuoptofil2025data} & 2025 & Peripheral blood cell type & 1 & 5,765\\
 & MedIMeta-AML~\cite{medimeta2025} & 2025 & AML cytomorphology & 1 & 3,674\\
 & CytoData~\cite{cytodata2025} & 2025 & Blood-cell morphology and artefacts & 1 & 1,000\\
 & Raabin-WBC~\cite{raabin2022} & 2022 & White blood cell type & 1 & 6,458\\
\midrule
Histopathology & Kather et al.~\cite{kather2016colorectal} & 2016 & Colorectal tissue class & 1 & 1,000\\
& NCT-CRC-HE-100K~\cite{kather2019survival} & 2019 & Colorectal tissue class & 1 & 1,314\\
& LC25000~\cite{borkowski2019lc25000} & 2019 & Colon tissue; lung tissue & 2 & 1,492; 2,258\\
& PatchCamelyon~\cite{veeling2018pcam} & 2018 & Metastasis detection & 1 & 32,768\\
& Tang et al.~\cite{tang2019alzheimer} & 2019 & Histology class & 1 & 491\\
& Wong et al.~\cite{wong2022amyloid} & 2022 & Histology class & 1 & 800\\
& Nirschl et al.~\cite{nirschl2018heartfailure} & 2018 & Tissue classification & 1 & 400\\
& Wu et al.~\cite{wu2023cryoet} & 2023 & Mitochondrial morphology & 1 & 256\\
\midrule
Radiology & CheXpert~\cite{irvin2019chexpert} & 2019 & Chest finding classification & 1 & 234\\
& PneumoniaMNIST~\cite{yang2023medmnist} & 2023 & Pneumonia classification & 1 & 624\\
& ChestMNIST~\cite{yang2023medmnist} & 2023 & 14 chest findings & 1 & 22,433\\
& RSNA Pneumonia~\cite{shih2019rsna} & 2019 & Pneumonia classification & 1 & 26,684\\
& BreastMNIST~\cite{yang2023medmnist} & 2023 & Breast ultrasound classification & 1 & 156\\
 & OrganMNIST 2D~\cite{yang2023medmnist} & 2023 & Organ identity: axial, coronal, sagittal CT & 3 & 17,778; 8,216; 8,827\\
\midrule
Ophthalmology & RetinaMNIST~\cite{yang2023medmnist} & 2023 & Retinal disease classification & 1 & 400\\
& DeepDRiD~\cite{liu2022deepdrid} & 2022 & Diabetic-retinopathy grading & 1 & 400\\
 & OCTMNIST~\cite{yang2023medmnist} & 2023 & Retinal OCT disease & 1 & 1,000\\
 & RFMiD~\cite{rfmid2021} & 2021 & Normal versus abnormal retina & 1 & 640\\
 & PAPILA~\cite{papila2022} & 2022 & Glaucoma status & 1 & 488\\
\midrule
Dermatology & DermaMNIST~\cite{yang2023medmnist} & 2023 & Skin-lesion classification & 1 & 2,005\\
& HAM10000~\cite{tschandl2018ham10000} & 2018 & Skin-lesion classification & 1 & 1,512\\
 & DERM12345~\cite{derm123452024} & 2024 & Nine skin-lesion classes (second level) & 1 & 2,485\\
\midrule
Surgery / anatomy & Dresden Surgical Anatomy~\cite{carstens2023dresden} & 2023 & 11 binary anatomy targets; one multilabel task & 12 & 2,942 per task\\
 & GastroVision~\cite{gastrovision2023} & 2023 & Endoscopic anatomy, findings, procedures & 1 & 1,586\\
 & HyperKvasir~\cite{hyperkvasir2020} & 2020 & Endoscopic anatomy, findings, procedures & 1 & 10,662\\
\midrule

Microbial / parasitic recognition & LIDS parasites~\cite{lidsdatasets} & 2025 & Helminth eggs, larvae and protozoan cysts & 3 & 2,564; 1,759; 4,791 per split\\
 & Bacteria2026~\cite{bacteria2026} & 2026 & Bacterial morphology and Gram status & 2 & 294 per task\\
\bottomrule
\end{tabularx}
\end{table*}
\endgroup

\setcounter{dbltopnumber}{4}
\renewcommand{\dbltopfraction}{0.95}
\renewcommand{\dblfloatpagefraction}{0.8}
\setlength{\dblfloatsep}{10pt plus 2pt minus 2pt}
\makeatletter
\setlength{\@dblfptop}{0pt}
\setlength{\@dblfpsep}{10pt}
\setlength{\@dblfpbot}{0pt plus 1fil}
\makeatother
\section{Bootstrap confidence intervals}
\label{app:bootstrap}
Tabs.~\ref{tab:bootstrap-classification}-\ref{tab:bootstrap-flickr} supplement all models in the three main performance tables with 95\% percentile confidence intervals from 1,000 bootstrap replicates. For classification, we resample test images with replacement independently within each task while keeping all 67 tasks and their aggregation weights fixed. We average task accuracies within datasets, datasets within domains, and the ten domains equally. Multilabel outcomes are resampled together at the image level; shared image draws are retained across the Dresden tasks and overlapping LIDS test partitions. Retrieval resamples query outcomes with the gallery and relevance rules fixed. These intervals reflect \mbox{test-sample variability for fixed checkpoints}.

\section{Secondary zero-shot metrics}
\label{app:secondary-metrics}
Accuracy is the primary classification metric in the main paper. We also report strict Macro-F1 and one-vs-rest AUC to show class-balanced performance and per-class ranking quality. Macro-F1 averages class-wise F1 scores equally;
AUC measures how well each class score ranks positive images above negative images. We use the same task-to-dataset-to-domain aggregation as for accuracy.

\section{Retrieval protocol}
\label{app:retrieval-protocol}

\subsection{BioLoomBench retrieval metrics}
We evaluate image-to-text (I2T) and text-to-image (T2I) retrieval with three text views: concise captions, original captions encoded at the native context length of each model, and original captions with a shared 77-token input limit. For each view, we report Recall@$K$ in both directions over the full 15,000-figure gallery. Tab.~\ref{tab:bioloombench-cap77} summarizes the 77-token results. Tabs.~\ref{tab:bioloombench-domain-r1} and~\ref{tab:bioloombench-content-r1} give bidirectional mean Recall@1 by domain and visual-content category. In these breakdowns, queries come from the corresponding subset, while the full gallery remains available for ranking.

\begin{figure*}[p]
\centering
\includegraphics[width=\textwidth,height=0.90\textheight,keepaspectratio]{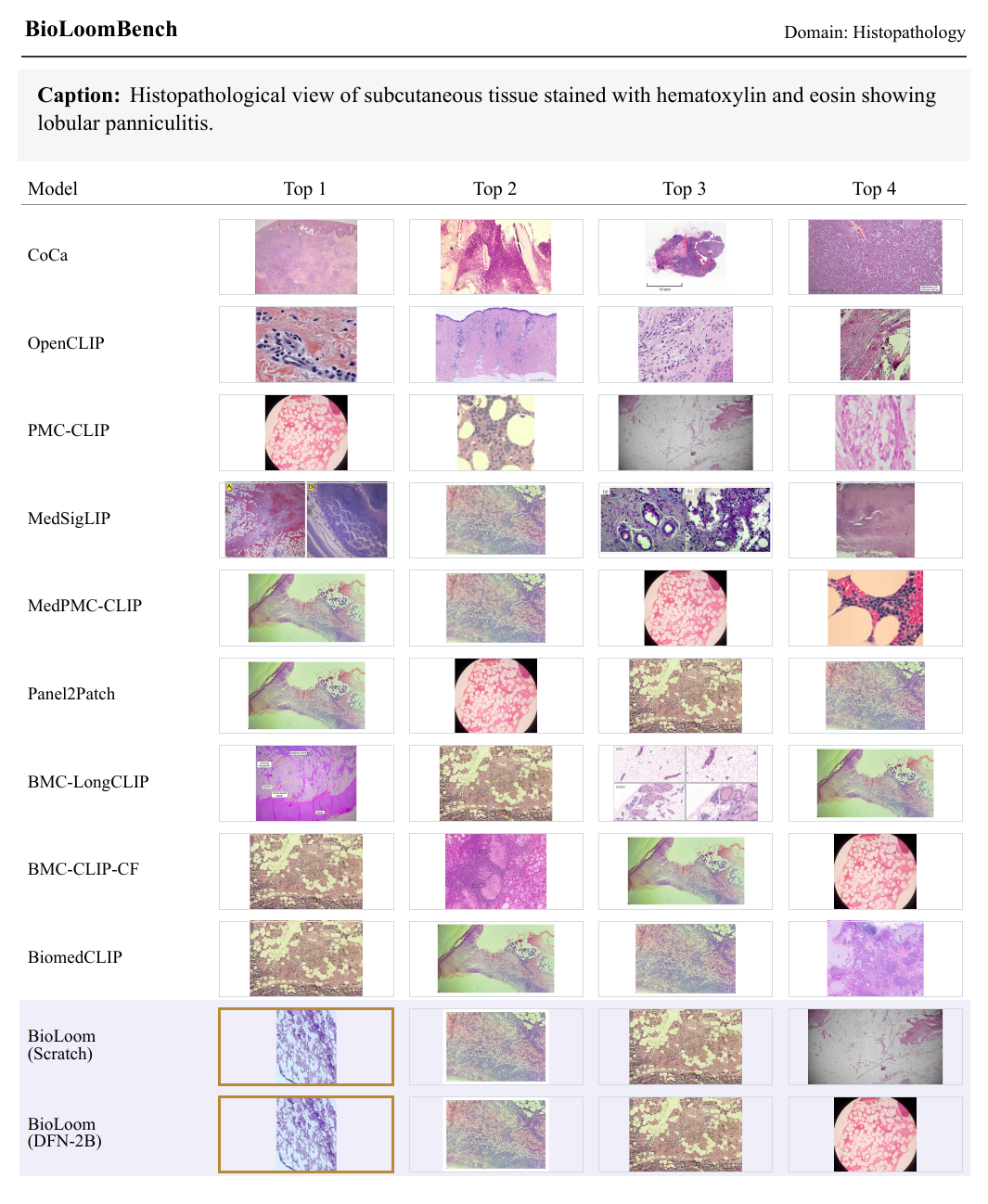}
\caption{\textbf{Concise-caption retrieval on BioLoomBench.} Top-4 results are shown for each model. Gold borders highlight retrieved images corresponding to the query caption.}
\label{fig:supp-retrieval-1}
\end{figure*}

\begin{figure*}[p]
\centering
\includegraphics[width=\textwidth,height=0.90\textheight,keepaspectratio]{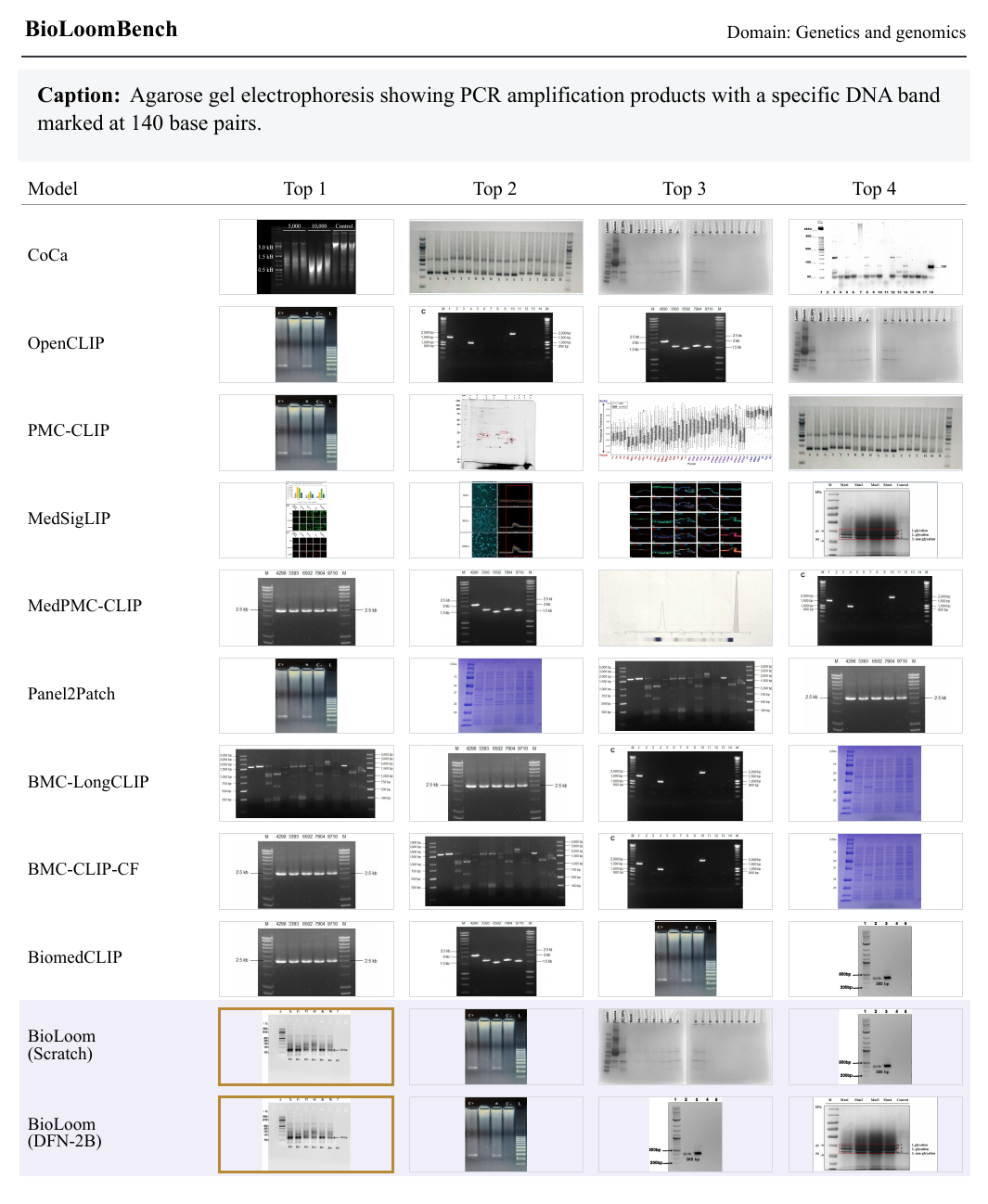}
\caption{\textbf{Concise-caption retrieval on BioLoomBench.} Top-4 results are shown for each model. Gold borders highlight retrieved images corresponding to the query caption.}
\label{fig:supp-retrieval-2}
\end{figure*}

\begin{figure*}[p]
\centering
\includegraphics[width=\textwidth,height=0.90\textheight,keepaspectratio]{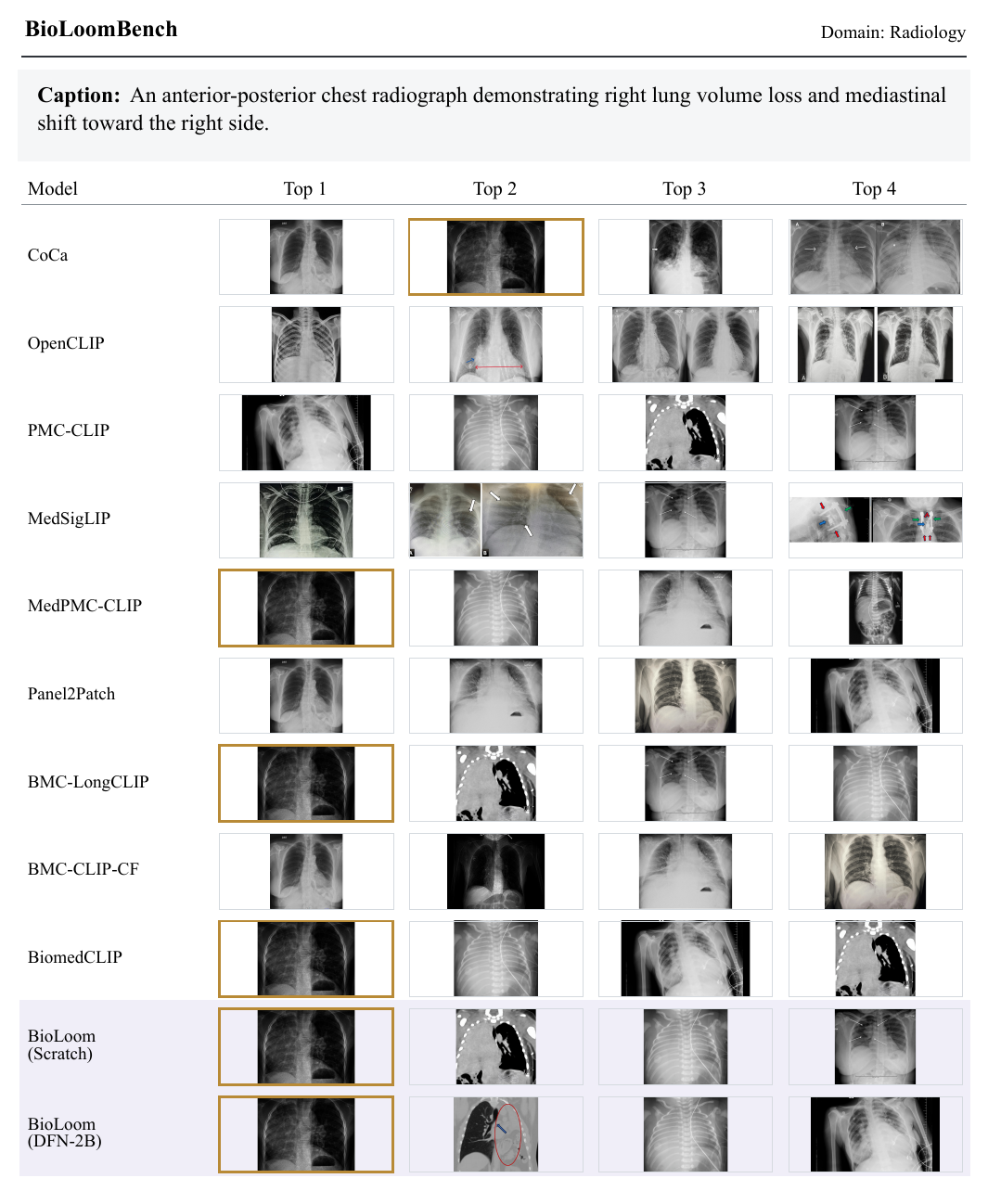}
\caption{\textbf{Concise-caption retrieval on BioLoomBench for a chest radiograph.} Top-4 results are shown for each model. Gold borders highlight retrieved images corresponding to the query caption.}
\label{fig:supp-retrieval-3}
\end{figure*}

\begin{figure*}[p]
\centering
\includegraphics[width=\textwidth,height=0.90\textheight,keepaspectratio]{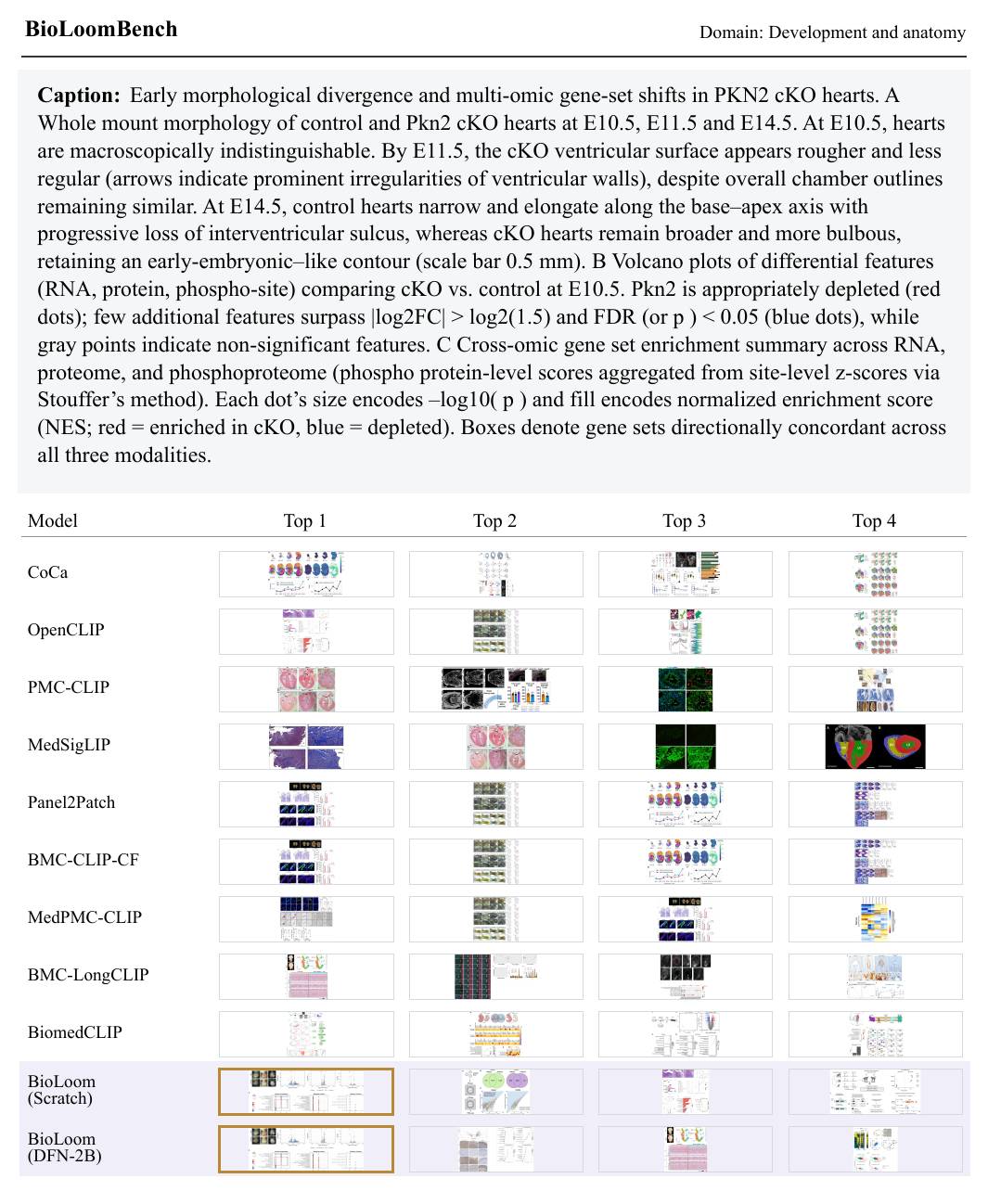}
\caption{\textbf{Original-caption retrieval on BioLoomBench for a figure combining embryonic heart images with multi-omic charts.} Top-4 results are shown for each model. Gold borders highlight retrieved images corresponding to the query caption.}
\label{fig:supp-retrieval-4}
\end{figure*}

\begin{figure*}[p]
\centering
\includegraphics[width=\textwidth,height=0.90\textheight,keepaspectratio]{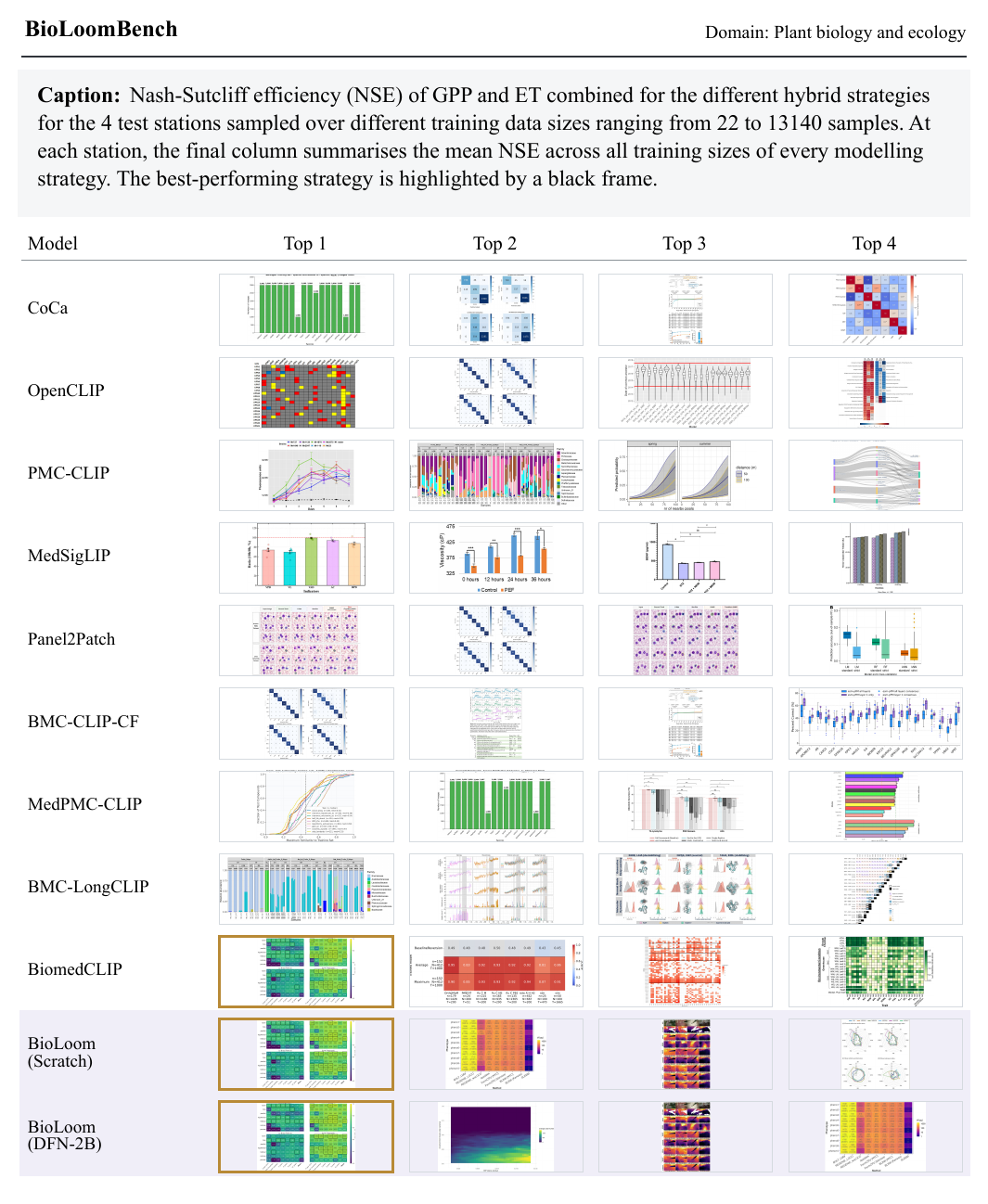}
\caption{\textbf{Original-caption retrieval on BioLoomBench for a chart query.} Top-4 results are shown for each model. Gold borders highlight retrieved images corresponding to the query caption.}
\label{fig:supp-retrieval-5}
\end{figure*}

Figs.~\ref{fig:supp-retrieval-1}-\ref{fig:supp-retrieval-3} show concise-caption retrieval, including a chest radiograph. Original-caption examples follow in Figs.~\ref{fig:supp-retrieval-4} and~\ref{fig:supp-retrieval-5}, covering a mixed biological-image/chart figure and a chart. All use the full 15,000-figure gallery and different queries from the main paper. Domains are labeled on each page. Both BioLoom variants rank the paired figure first in these selected cases.

\subsection{BioMedFlickr retrieval protocol}

The BioMedFlickr test set introduced by BIOMEDICA~\citep{lozano2025biomedica} contains 7,185 image-caption pairs. Among its captions, 749 texts occur more than once, with up to 67 images sharing a caption. We therefore treat images with identical captions as a common relevant set. For row $i$ with caption $c_i$, let $P_i=\{j:c_j=c_i\}$. We compute
\begin{equation}
  \operatorname{R@K}_{\mathrm{text}}
  =\frac{1}{N}\sum_{i=1}^{N}
  \mathbb{1}\!\left[\operatorname{TopK}(i)\cap P_i\neq\varnothing\right].
  \label{eq:identical-caption-recall}
\end{equation}
We use this rule in both directions over the full gallery. Paraphrases remain separate. Tab.~\ref{tab:retrieval-official-exact} additionally reports the original exact-pair protocol, which requires retrieval of the paired row. All rows report our evaluations using released or trained checkpoints. For CoCa, OpenCLIP, BiomedCLIP, and BMC-CLIP-CF, the exact-pair scores come from the same saved evaluation outputs as the main-table scores, with the model, queries, and gallery fixed. With rankings fixed, expanding the relevant set does not reduce Recall@$K$ in these paired results.

\begin{figure*}[p]
\centering
\includegraphics[width=\textwidth,height=0.90\textheight,keepaspectratio]{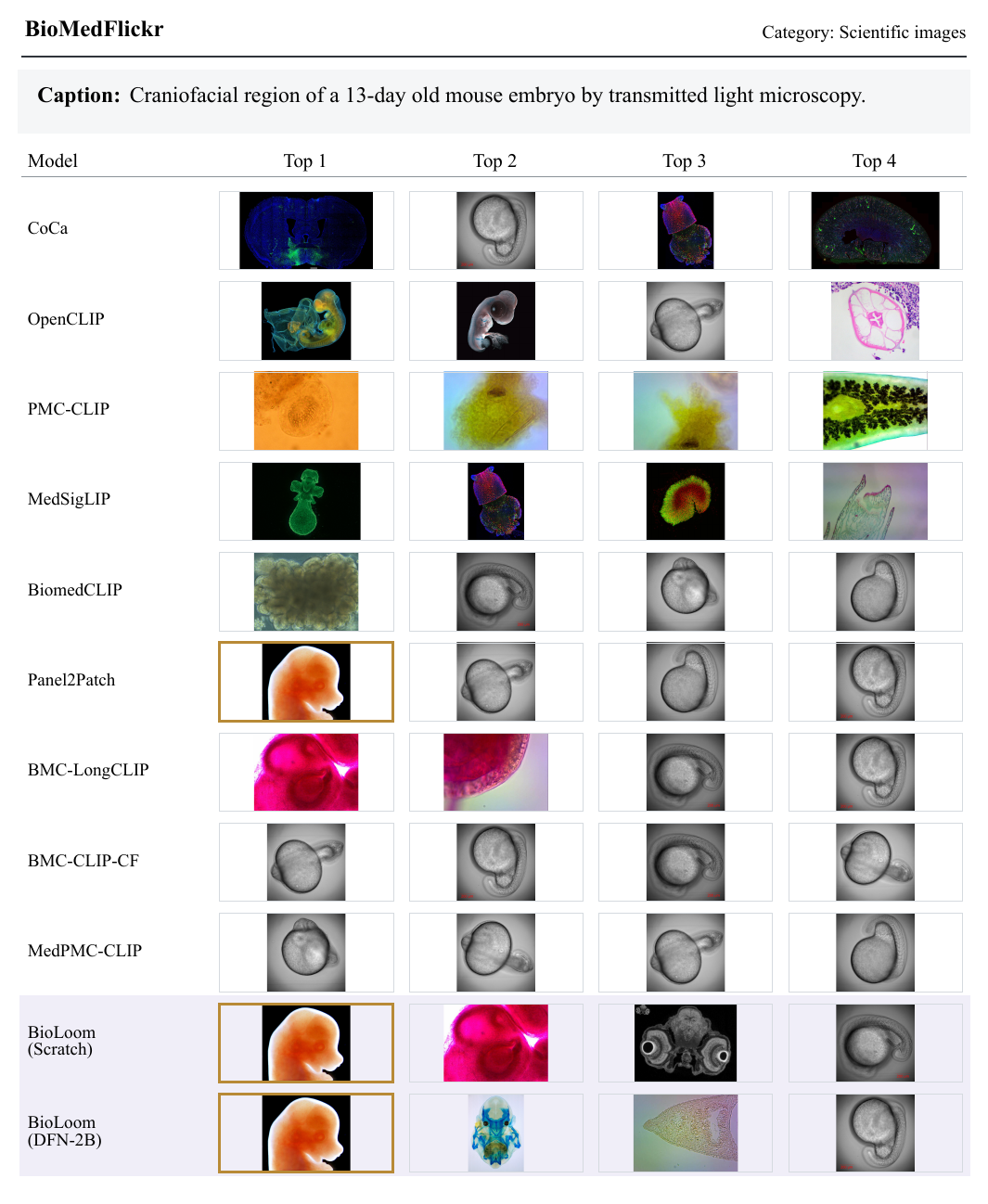}
\caption{\textbf{Original-caption retrieval on BioMedFlickr.} Top-4 results are shown for each model. Gold borders highlight retrieved images corresponding to the query caption.}
\label{fig:supp-retrieval-6}
\end{figure*}

\begin{figure*}[p]
\centering
\includegraphics[width=\textwidth,height=0.90\textheight,keepaspectratio]{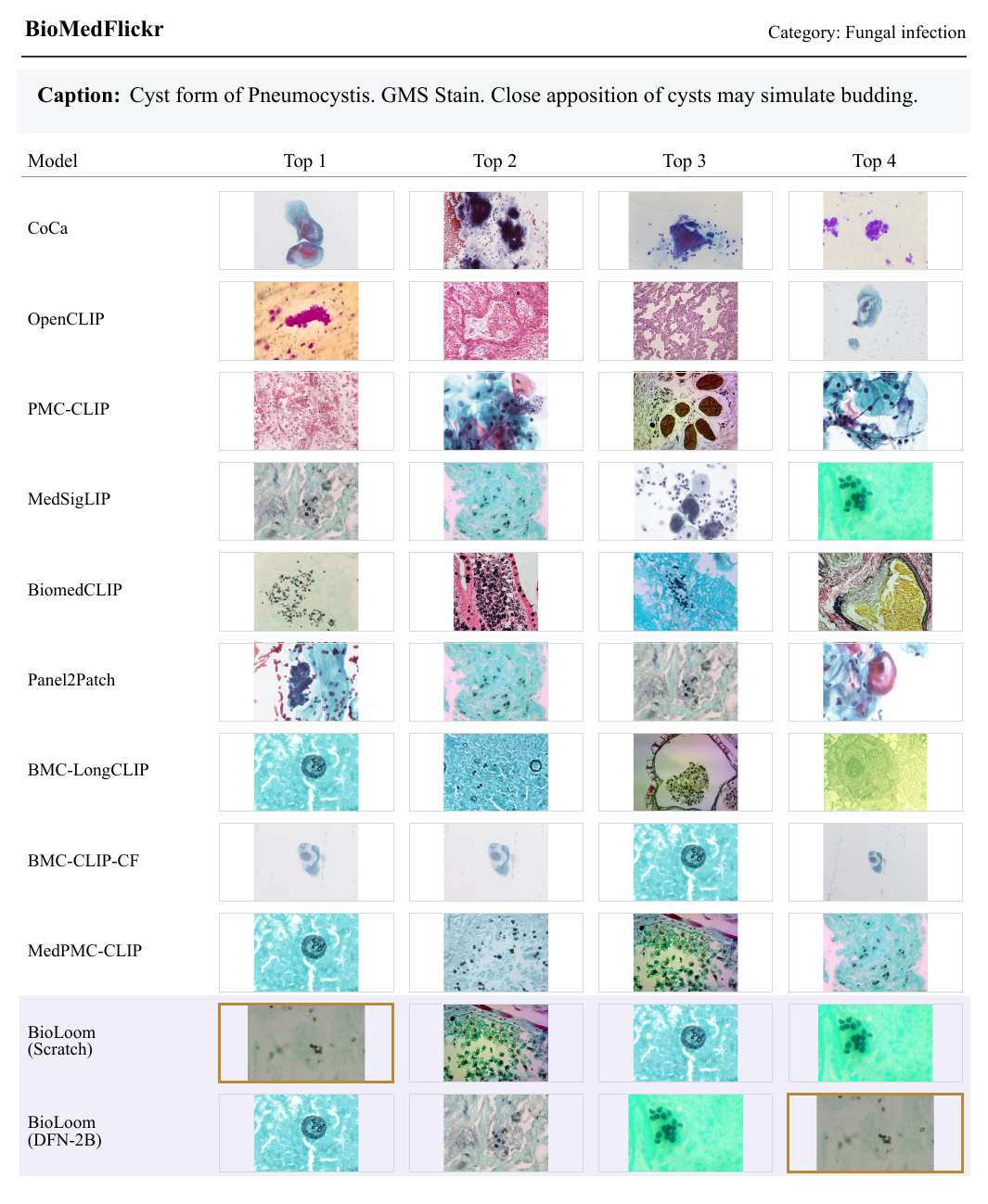}
\caption{\textbf{Original-caption retrieval on BioMedFlickr.} Top-4 results are shown for each model. Gold borders highlight retrieved images corresponding to the query caption.}
\label{fig:supp-retrieval-7}
\end{figure*}

Figs.~\ref{fig:supp-retrieval-6} and~\ref{fig:supp-retrieval-7} show additional examples over the full 7,185-image gallery, with images sharing the query caption treated as relevant. Headers use BioMedFlickr categories.

\section{Linear probing across biomedical datasets}
\label{app:linear-probing}
We report dataset-level accuracy, Macro-F1, and AUC for the 15 frozen image encoders included in the main-paper linear-probing comparison, using all 12 MedMNIST 2D datasets~\citep{yang2023medmnist}. We use the official splits and 224-pixel image release, apply the preprocessing of each encoder, and fit logistic classifiers on normalized features from the official training splits. The regularization parameter $C\in\{0.01,0.1,1,10\}$ is selected on validation data by accuracy for single-label tasks and by \mbox{Macro-F1 for ChestMNIST}.

Tabs.~\ref{tab:linear-probe-acc}, \ref{tab:linear-probe-f1}, and \ref{tab:linear-probe-auc} report the three metrics. ChestMNIST is multilabel: accuracy is per-label agreement, and F1 and AUC are averaged over labels. For single-label tasks, F1 is averaged over classes; AUC uses the positive class for binary tasks and macro one-vs-rest AUC for multiclass tasks. Overall means give each dataset equal weight. The selected TissueMNIST classifiers for EVA-02, SigLIP2, CommonPool, and ImageNet ResNet-50 reached the 500-iteration optimization limit.

\begin{figure*}[!t]
  \centering
  \includegraphics[width=\textwidth,trim=0 14bp 0 0,clip]{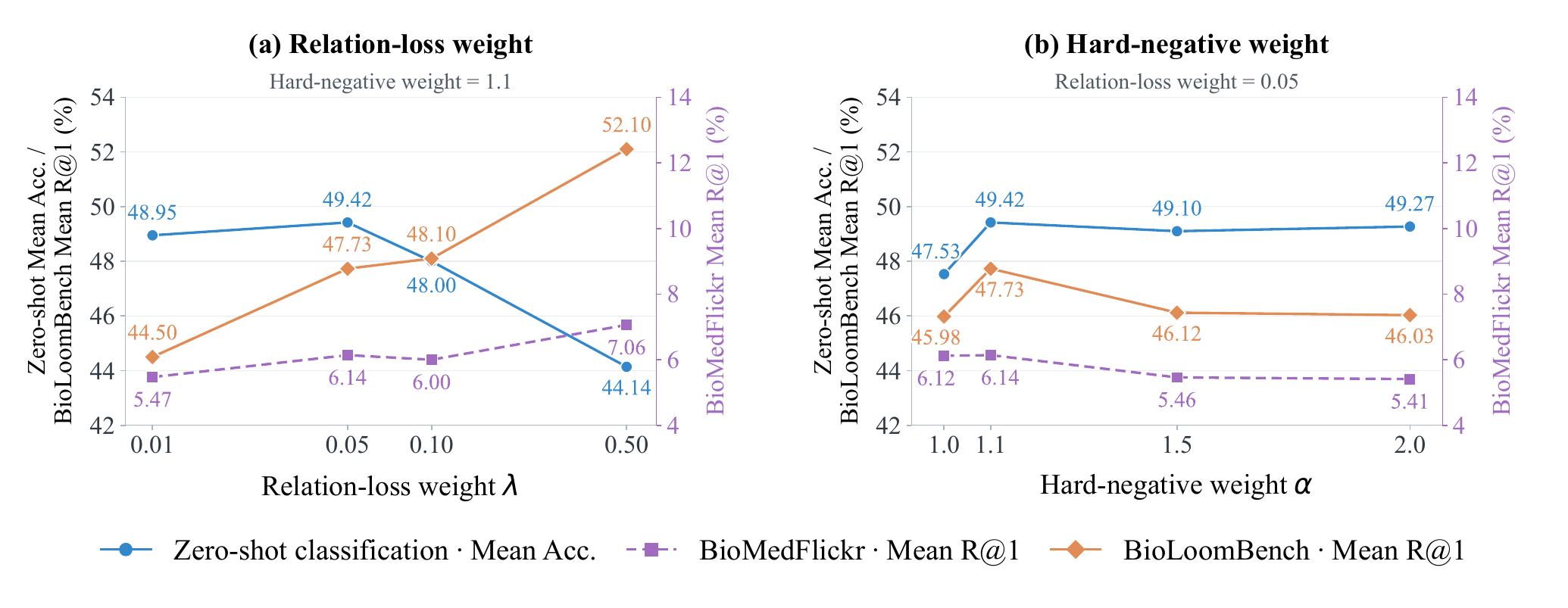}
  \captionsetup{skip=4pt}
  \caption{\textbf{Sensitivity to relation-loss and hard-negative weights using a 25M image-text pair subset of Loom52M.} The left panel varies the relation-loss weight with the hard-negative weight fixed at 1.1; the right panel varies the hard-negative weight with the relation-loss weight fixed at 0.05. We report zero-shot mean accuracy and bidirectional mean R@1 (\%); BioLoomBench uses concise captions.}
  \label{fig:hyperparameter-sensitivity}
\end{figure*}

\section{MMBU question-option evaluation}
\label{app:mmbu}
\subsection{Evaluation protocol}
To broaden our evaluation of biomedical image-text models, we evaluate 11 image-text models on the released MMBU test split~\citep{dcunha2026mmbu}. We use 2,677 examples from three single-answer classification families: whole-image classification, classification with a displayed bounding box, and classification with a displayed segmentation mask. For each answer option, we append \texttt{Answer: [option].} to the original question to form a candidate text. Each model encodes the image and candidate texts using its native tokenizer; texts exceeding its context length are truncated. The model selects the option with the highest cosine similarity. This is a closed-choice evaluation derived from MMBU, rather than its official generative VQA protocol.

We use the official ResNet-50 checkpoint for PMC-CLIP~\cite{lin2023pmcclipcontrastivelanguageimagepretraining}, as in the main tables and \mbox{linear-probing evaluation}.

For the box- and mask-indicated tasks, the released image already displays the reference annotation. We pass the complete image to each encoder without cropping the region or processing the annotation separately. These results measure classification with the target region indicated; they do not measure localization or segmentation. We exclude counting and the 219 coordinate-prediction detection examples. We report accuracy for each task family. Because each example has one correct option and each model selects one option, accuracy is numerically equal to micro-F1.

\subsection{Results}
Tab.~\ref{tab:mmbu-option-caption} reports zero-shot classification accuracy by task family. BioLoom (Scratch) obtains the highest accuracy in all three families: 42.5\% on whole-image classification, 36.7\% on bounding-box-indicated classification, and 33.6\% on mask-indicated classification. DFN-2B initialization performs worse than Scratch in all three families, showing that its gains on the main benchmarks do not extend to every evaluation setting.

\section{Fine-grained and cross-level retrieval}
\label{app:panel-retrieval}
We extend BioLoomBench with panel and cross-level retrieval. Gemini annotates panel bounding boxes, descriptions, and corresponding original-caption excerpts. After automatic checks, visual spot checks, and filtering uncertain crops and mappings, the test set contains 21,975 panels from 7,715 whole figures across 15 domains.

We define four retrieval tasks on this shared test set. Panel-to-description and description-to-panel retrieval use the 21,975 test panels and their descriptions as queries and galleries in opposite directions. Description-to-whole-figure retrieval uses all 21,975 panel descriptions as queries against the original 15,000 whole figures; this gallery contains no panel crops. Panel-to-original-caption retrieval uses 19,622 panel queries against the original 15,000 whole-figure captions. This task retains panels whose original-caption excerpts pass the matching checks, since a whole-figure caption may not describe every panel. The excerpts determine query eligibility; the retrieval targets are the full original captions.

We evaluate all 11 main-table models with the same queries and galleries for each task, using the native text context length of each model. Relevance follows panel-caption and parent-figure links, including duplicate captions or images. Tab.~\ref{tab:panel-retrieval} reports Recall@1 separately for each task. BioLoom (DFN-2B) achieves the highest score in all four settings. Compared with MedPMC-CLIP, it improves description-to-whole-figure and panel-to-original-caption Recall@1 by 9.35 and 21.40 percentage points, respectively. These results suggest better alignment between panel content and whole-figure context.

\section{Hyperparameter Sensitivity}
\label{app:hyperparameter-sensitivity}
We conduct all hyperparameter sensitivity experiments on a 25M image-text pair subset of Loom52M. Fig.~\ref{fig:hyperparameter-sensitivity} compares the relation-loss and hard-negative weights. We test relation-loss weights $\{0.01, 0.05, 0.10, 0.50\}$ at a hard-negative weight of 1.1, and hard-negative weights $\{1.0, 1.1, 1.5, 2.0\}$ at a relation-loss weight of 0.05.

\textbf{Relation-loss weight.} Increasing the weight from 0.05 to 0.50 raises mean R@1 from 47.73\% to 52.10\% on BioLoomBench and from 6.14\% to 7.06\% on BioMedFlickr, while zero-shot accuracy falls from 49.42\% to 44.14\%. A larger weight places more emphasis on learning relations among images and text within the same figure. These endpoint changes suggest a retrieval--classification trade-off, although the BioMedFlickr gains are not monotonic.

\textbf{Hard-negative weight.} Increasing the weight from 1.0 to 1.1 improves accuracy from 47.53\% to 49.42\% and BioLoomBench mean R@1 from 45.98\% to 47.73\%, while BioMedFlickr changes little (6.12\% to 6.14\%). From 1.1 to 2.0, zero-shot accuracy changes little, ranging from 49.10\% to 49.42\%. In contrast, BioLoomBench mean R@1 falls from 47.73\% to 46.03\%, and BioMedFlickr mean R@1 falls from 6.14\% to 5.41\%. Stronger weighting of same-content-type negatives therefore does not provide a consistent classification gain in these runs and is accompanied by lower retrieval scores. These results suggest balancing relation supervision and hard-negative weighting across tasks rather than simply increasing the weights.

\Needspace{14\baselineskip}
\section{Computational Cost}
\label{app:computational-cost}
At Batch API rates, annotating the 13.1M whole figures retained in Loom52M has an estimated cost of \$16,700, or about \$1,270 per million figures. Measured online throughput with 96 concurrent requests corresponds to approximately 139 hours for this subset.

At the measured throughput on eight H200 GPUs, ViT-L/14 pretraining for 10 and 25 epochs is projected to take about 42 and 105 hours, respectively.

\section{Discussion}
\label{app:discussion}

\textbf{Learning from article structure.} Scientific articles provide context and relationships beyond individual image-text pairs. BioLoom uses original captions and related paragraphs to enrich panel descriptions, and figure and article links to guide contrastive learning. These two uses are complementary: the descriptions explain panel content, while the training objective distinguishes exact matches, related pairs, and cross-article negatives. Shared article context does not imply a positive match, so we mask within-article cross-figure pairs rather than treating them as positives or negatives. Among cross-article negatives, we place greater emphasis on pairs of the same content type.

\textbf{Dataset characteristics.} Loom52M combines whole figures paired with original captions and panels paired with data descriptions and generated captions. Panels account for more than half of its 52.4M image-text pairs, providing more localized supervision while preserving links to their parent figures. We retain biomedical images and chart panels within mixed figures. This supports joint learning across these content types and retrieval of both biomedical images and scientific charts. The data-pipeline ablations examine how text supervision, image views, and chart inclusion affect downstream performance.

\textbf{Limitations and future work.} Scaling the vision encoder from ViT-L to ViT-H does not improve performance under our current training setup. This suggests that increasing model capacity alone may be insufficient; future work will examine whether more training data and adjusted optimization settings allow ViT-H to benefit from its larger capacity. BioLoom also uses a dual-encoder architecture and does not generate textual responses. We will explore integrating its vision encoder with a language decoder for generative biomedical vision-language tasks.

\clearpage

\begingroup
\def\jointauccap{}
\begin{table*}[!p]
\captionsetup{skip=3pt}

\ifdefined\jointauccap\begingroup\else\begin{table*}[t]\fi
  \caption{\textbf{Macro-F1 across ten biomedical domains.} We report strict class Macro-F1 (\%) using the same prompts and aggregation as the main-paper accuracy results. Biomedical model rows are ordered by the corresponding Overall score from low to high. Abbreviations follow the main-paper accuracy table. $^{\ast}$ denotes random vision-encoder initialization for image-text pretraining. Among biomedical models, bold and underlined values indicate the best and second-best scores in each column, respectively.}
  \label{tab:zero-shot-f1}
  \centering
  \resulttablefont
  \ifdefined\jointauccap\renewcommand{\arraystretch}{1.0}\else\renewcommand{\arraystretch}{1.12}\fi
  \setlength{\tabcolsep}{1.4pt}
  \begin{tabularx}{\textwidth}{>{\raggedright\arraybackslash}p{0.20\textwidth}*{11}{Y}}
    \toprule
    \textbf{Model} & \multicolumn{1}{c}{\textbf{Cell}} & \multicolumn{1}{c}{\textbf{Struct.}} & \multicolumn{1}{c}{\textbf{Micro.}} & \multicolumn{1}{c}{\textbf{Cytology}} & \multicolumn{1}{c}{\textbf{Histo.}} & \multicolumn{1}{c}{\textbf{Radiol.}} & \multicolumn{1}{c}{\textbf{Ophth.}} & \multicolumn{1}{c}{\textbf{Derm.}} & \multicolumn{1}{c}{\textbf{Surgery}} & \multicolumn{1}{c}{\textbf{Microb.}} & \multicolumn{1}{c}{\textbf{Overall}} \\
    \midrule
    CoCa$^{\ast}$~\cite{yu2022cocacontrastivecaptionersimagetext} & \tabnum{14.77} & \tabnum{22.40} & \tabnum{35.10} & \tabnum{15.04} & \tabnum{29.59} & \tabnum{17.58} & \tabnum{25.29} & \tabnum{13.04} & \tabnum{14.39} & \tabnum{21.12} & \tabnum{20.83} \\
    OpenCLIP$^{\ast}$~\cite{cherti2023reproduciblescalinglawscontrastive} & \tabnum{17.25} & \tabnum{34.54} & \tabnum{31.19} & \tabnum{6.37} & \tabnum{32.25} & \tabnum{29.36} & \tabnum{25.11} & \tabnum{11.92} & \tabnum{16.69} & \tabnum{32.12} & \tabnum{23.68} \\
    \midrule
    Panel2Patch~\cite{yuan2026panel2patch} & \tabnum{19.94} & \tabnum{38.18} & \tabnum{62.93} & \tabnum{5.85} & \tabnum{44.05} & \tabnum{38.31} & \tabnum{24.76} & \tabnum{13.64} & \tabnum{25.20} & \tabnum{21.27} & \tabnum{29.41} \\
    \mbox{BMC-CLIP-CF}~\cite{lozano2025biomedica} & \tabnum{14.58} & \tabnum{44.27} & \tabnum{57.25} & \tabnum{6.23} & \tabnum{42.43} & \tabnum{38.46} & \tabnum{24.72} & \tabnum{15.23} & \tabnum{26.26} & \tabnum{25.84} & \tabnum{29.53} \\
    \mbox{PMC-CLIP}$^{\ast}$~\cite{lin2023pmcclipcontrastivelanguageimagepretraining} & \tabnum{20.64} & \tabnum{36.74} & \tabnum{59.29} & \tabnum{7.33} & \tabnum{44.07} & \tabnum{37.91} & \tabnum{28.07} & \tabnum{13.03} & \tabnum{21.68} & \tabnum{27.49} & \tabnum{29.62} \\
    MedSigLIP~\cite{sellergren2025medgemmatechnicalreport} & \tabnum{15.36} & \tabnum{30.03} & \tabnum{57.37} & \tabnum{19.74} & \tabnum{31.45} & \tabnum{39.63} & \tabnum{36.83} & \tabnum{16.53} & \tabnum{14.13} & \tabnum{36.09} & \tabnum{29.71} \\
    BiomedCLIP~\cite{zhang2023biomedclipmultimodalbiomedicalfoundation} & \tabnum{17.62} & \tabnum{34.31} & \tabnum{61.83} & \tabnum{6.10} & \tabnum{42.35} & \tabnum{41.01} & \tabnum{31.85} & \tabnum{14.97} & \tabnum{29.13} & \tabnum{28.30} & \tabnum{30.75} \\
    \mbox{BMC-LongCLIP}~\cite{sun2025tokenswastedleveraginglong} & \tabnum{17.39} & \tabnum{40.00} & \tabnum{63.86} & \tabnum{16.65} & \tabnum{39.59} & \tabnum{45.48} & \tabnum{30.91} & \tabnum{19.83} & \tabnum{28.32} & \tabnum{25.00} & \tabnum{32.70} \\
    \mbox{MedPMC-CLIP}~\cite{kim2026medpmc} & \tabnum{21.74} & \tabnum{\underline{52.67}} & \tabnum{44.09} & \tabnum{24.96} & \tabnum{\underline{50.49}} & \tabnum{43.53} & \tabnum{40.80} & \tabnum{23.74} & \tabnum{23.49} & \tabnum{33.31} & \tabnum{35.88} \\
    \midrule
    \rowcolor{bioloomlight}
    \methodname{} (Scratch)$^{\ast}$ & \tabnum{\textbf{25.49}} & \tabnum{43.90} & \tabnum{\textbf{66.09}} & \tabnum{\underline{27.30}} & \tabnum{48.91} & \tabnum{\textbf{48.76}} & \tabnum{\underline{45.05}} & \tabnum{\underline{27.11}} & \tabnum{\underline{38.23}} & \tabnum{\underline{38.13}} & \tabnum{\underline{40.90}} \\
    \rowcolor{bioloomlight}
    \methodname{} (DFN-2B) & \tabnum{\underline{22.40}} & \tabnum{\textbf{53.68}} & \tabnum{\underline{64.92}} & \tabnum{\textbf{32.45}} & \tabnum{\textbf{52.77}} & \tabnum{\underline{47.97}} & \tabnum{\textbf{50.43}} & \tabnum{\textbf{31.41}} & \tabnum{\textbf{41.20}} & \tabnum{\textbf{39.90}} & \tabnum{\textbf{43.71}} \\
    \bottomrule
  \end{tabularx}

\ifdefined\jointauccap\endgroup\else\end{table*}\fi

\par\vspace{4pt}

\ifdefined\jointauccap\begingroup\else\begin{table*}[t]\fi
  \caption{\textbf{One-vs-rest AUC across ten biomedical domains.} We report mean class AUC (\%) using the same prompts and aggregation as the main-paper accuracy results. Biomedical model rows are ordered by the corresponding Overall score from low to high. Abbreviations follow the main-paper accuracy table. $^{\ast}$ denotes random vision-encoder initialization for image-text pretraining. Among biomedical models, bold and underlined values indicate the best and second-best scores in each column, respectively.}
  \label{tab:zero-shot-auc}
  \centering
  \resulttablefont
  \ifdefined\jointauccap\renewcommand{\arraystretch}{1.0}\else\renewcommand{\arraystretch}{1.12}\fi
  \setlength{\tabcolsep}{1.4pt}
  \begin{tabularx}{\textwidth}{>{\raggedright\arraybackslash}p{0.20\textwidth}*{11}{Y}}
    \toprule
    \textbf{Model} & \multicolumn{1}{c}{\textbf{Cell}} & \multicolumn{1}{c}{\textbf{Struct.}} & \multicolumn{1}{c}{\textbf{Micro.}} & \multicolumn{1}{c}{\textbf{Cytology}} & \multicolumn{1}{c}{\textbf{Histo.}} & \multicolumn{1}{c}{\textbf{Radiol.}} & \multicolumn{1}{c}{\textbf{Ophth.}} & \multicolumn{1}{c}{\textbf{Derm.}} & \multicolumn{1}{c}{\textbf{Surgery}} & \multicolumn{1}{c}{\textbf{Microb.}} & \multicolumn{1}{c}{\textbf{Overall}} \\
    \midrule
    CoCa$^{\ast}$~\cite{yu2022cocacontrastivecaptionersimagetext} & \tabnum{57.40} & \tabnum{71.24} & \tabnum{82.57} & \tabnum{64.76} & \tabnum{69.28} & \tabnum{48.67} & \tabnum{55.37} & \tabnum{62.17} & \tabnum{57.75} & \tabnum{57.82} & \tabnum{62.70} \\
    OpenCLIP$^{\ast}$~\cite{cherti2023reproduciblescalinglawscontrastive} & \tabnum{62.83} & \tabnum{77.18} & \tabnum{87.32} & \tabnum{58.69} & \tabnum{74.56} & \tabnum{53.15} & \tabnum{52.15} & \tabnum{63.24} & \tabnum{60.73} & \tabnum{66.73} & \tabnum{65.66} \\
    \midrule
    \mbox{PMC-CLIP}$^{\ast}$~\cite{lin2023pmcclipcontrastivelanguageimagepretraining} & \tabnum{70.36} & \tabnum{80.15} & \tabnum{86.59} & \tabnum{56.60} & \tabnum{82.30} & \tabnum{68.52} & \tabnum{67.28} & \tabnum{63.15} & \tabnum{69.53} & \tabnum{60.74} & \tabnum{70.52} \\
    \mbox{BMC-CLIP-CF}~\cite{lozano2025biomedica} & \tabnum{63.83} & \tabnum{82.15} & \tabnum{88.15} & \tabnum{59.18} & \tabnum{79.35} & \tabnum{71.18} & \tabnum{67.57} & \tabnum{71.03} & \tabnum{73.93} & \tabnum{65.77} & \tabnum{72.22} \\
    BiomedCLIP~\cite{zhang2023biomedclipmultimodalbiomedicalfoundation} & \tabnum{62.51} & \tabnum{79.14} & \tabnum{89.73} & \tabnum{55.66} & \tabnum{80.13} & \tabnum{74.36} & \tabnum{70.65} & \tabnum{65.27} & \tabnum{77.11} & \tabnum{68.95} & \tabnum{72.35} \\
    Panel2Patch~\cite{yuan2026panel2patch} & \tabnum{71.61} & \tabnum{79.33} & \tabnum{91.33} & \tabnum{58.40} & \tabnum{79.66} & \tabnum{69.94} & \tabnum{68.32} & \tabnum{71.35} & \tabnum{74.22} & \tabnum{67.50} & \tabnum{73.16} \\
    MedSigLIP~\cite{sellergren2025medgemmatechnicalreport} & \tabnum{61.90} & \tabnum{75.98} & \tabnum{85.64} & \tabnum{75.27} & \tabnum{71.97} & \tabnum{76.00} & \tabnum{75.75} & \tabnum{72.88} & \tabnum{68.08} & \tabnum{\underline{80.26}} & \tabnum{74.37} \\
    \mbox{BMC-LongCLIP}~\cite{sun2025tokenswastedleveraginglong} & \tabnum{66.42} & \tabnum{82.51} & \tabnum{90.72} & \tabnum{69.79} & \tabnum{81.97} & \tabnum{77.36} & \tabnum{72.82} & \tabnum{74.52} & \tabnum{81.65} & \tabnum{74.21} & \tabnum{77.20} \\
    \mbox{MedPMC-CLIP}~\cite{kim2026medpmc} & \tabnum{72.33} & \tabnum{\textbf{86.36}} & \tabnum{\underline{92.46}} & \tabnum{76.47} & \tabnum{\underline{85.34}} & \tabnum{77.74} & \tabnum{77.55} & \tabnum{76.51} & \tabnum{79.53} & \tabnum{79.57} & \tabnum{80.39} \\
    \midrule
    \rowcolor{bioloomlight}
    \methodname{} (Scratch)$^{\ast}$ & \tabnum{\textbf{73.59}} & \tabnum{82.69} & \tabnum{\textbf{93.83}} & \tabnum{\underline{80.75}} & \tabnum{85.02} & \tabnum{\textbf{78.76}} & \tabnum{\textbf{80.39}} & \tabnum{\underline{78.66}} & \tabnum{\underline{82.65}} & \tabnum{\textbf{83.79}} & \tabnum{\underline{82.01}} \\
    \rowcolor{bioloomlight}
    \methodname{} (DFN-2B) & \tabnum{\underline{73.01}} & \tabnum{\underline{84.74}} & \tabnum{91.94} & \tabnum{\textbf{83.44}} & \tabnum{\textbf{86.71}} & \tabnum{\underline{78.09}} & \tabnum{\underline{80.00}} & \tabnum{\textbf{81.24}} & \tabnum{\textbf{85.58}} & \tabnum{79.16} & \tabnum{\textbf{82.39}} \\
    \bottomrule
  \end{tabularx}

\ifdefined\jointauccap\endgroup\else\end{table*}\fi

\par\vspace{4pt}

\ifdefined\jointauccap\begingroup\else\begin{table*}[t]\fi
  \caption{\textbf{Model comparison on BioLoomBench retrieval with original captions and a common 77-token maximum.} Recall@$K$ (\%) is reported in both directions. A model whose native context is shorter than 77 tokens uses its native limit. Biomedical model rows are ordered by bidirectional mean Recall@1 from low to high. Our rows have a light-purple background. I2T and T2I denote image-to-text and text-to-image retrieval, respectively. For \methodname{}, Scratch and DFN-2B denote random and DFN-2B initialization of the vision encoder, respectively; both use the same pretrained text encoder. $^{\ast}$ denotes random vision-encoder initialization for image-text pretraining. Among biomedical models, bold and underlined values indicate the best and second-best scores in each column, respectively.}
  \label{tab:bioloombench-cap77}
  \centering
  \resulttablefont
  \ifdefined\jointauccap\renewcommand{\arraystretch}{1.0}\else\renewcommand{\arraystretch}{1.12}\fi
  \setlength{\tabcolsep}{4pt}
  \begin{tabularx}{\textwidth}{>{\raggedright\arraybackslash}p{0.22\textwidth}*{6}{Y}}
    \toprule
    & \multicolumn{3}{c}{\textbf{Image $\rightarrow$ Text}} & \multicolumn{3}{c}{\textbf{Text $\rightarrow$ Image}} \\
    \cmidrule(lr){2-4}\cmidrule(lr){5-7}
    \textbf{Model} & R@1 & R@10 & R@100 & R@1 & R@10 & R@100 \\
    \midrule
    CoCa$^{\ast}$~\cite{yu2022cocacontrastivecaptionersimagetext} & \tabnum{18.37} & \tabnum{36.88} & \tabnum{64.68} & \tabnum{18.53} & \tabnum{36.72} & \tabnum{62.48} \\
    OpenCLIP$^{\ast}$~\cite{cherti2023reproduciblescalinglawscontrastive} & \tabnum{25.34} & \tabnum{47.37} & \tabnum{73.49} & \tabnum{24.68} & \tabnum{45.66} & \tabnum{71.75} \\
    \midrule
    \mbox{PMC-CLIP}$^{\ast}$~\cite{lin2023pmcclipcontrastivelanguageimagepretraining} & \tabnum{4.47} & \tabnum{13.69} & \tabnum{30.34} & \tabnum{6.05} & \tabnum{18.04} & \tabnum{37.51} \\
    MedSigLIP~\cite{sellergren2025medgemmatechnicalreport} & \tabnum{19.88} & \tabnum{44.66} & \tabnum{72.75} & \tabnum{9.43} & \tabnum{26.52} & \tabnum{53.27} \\
    \mbox{BMC-LongCLIP}~\cite{sun2025tokenswastedleveraginglong} & \tabnum{43.39} & \tabnum{72.66} & \tabnum{90.73} & \tabnum{41.75} & \tabnum{70.69} & \tabnum{89.51} \\
    Panel2Patch~\cite{yuan2026panel2patch} & \tabnum{47.75} & \tabnum{74.83} & \tabnum{91.61} & \tabnum{49.57} & \tabnum{77.01} & \tabnum{92.84} \\
    BiomedCLIP~\cite{zhang2023biomedclipmultimodalbiomedicalfoundation} & \tabnum{50.26} & \tabnum{79.15} & \tabnum{93.98} & \tabnum{48.35} & \tabnum{77.69} & \tabnum{93.53} \\
    \mbox{BMC-CLIP-CF}~\cite{lozano2025biomedica} & \tabnum{52.09} & \tabnum{78.96} & \tabnum{92.94} & \tabnum{51.89} & \tabnum{78.97} & \tabnum{93.25} \\
    \mbox{MedPMC-CLIP}~\cite{kim2026medpmc} & \tabnum{55.42} & \tabnum{80.70} & \tabnum{93.03} & \tabnum{\underline{56.14}} & \tabnum{\underline{81.63}} & \tabnum{93.35} \\
    \midrule
    \rowcolor{bioloomlight}
    \methodname{} (Scratch)$^{\ast}$& \tabnum{\underline{56.03}} & \tabnum{\underline{83.45}} & \tabnum{\underline{95.55}} & \tabnum{53.20} & \tabnum{81.06} & \tabnum{\underline{94.43}} \\
    \rowcolor{bioloomlight}
    \methodname{} (DFN-2B)& \tabnum{\textbf{63.02}} & \tabnum{\textbf{88.49}} & \tabnum{\textbf{96.73}} & \tabnum{\textbf{61.75}} & \tabnum{\textbf{87.45}} & \tabnum{\textbf{96.61}} \\
    \bottomrule
  \end{tabularx}

\ifdefined\jointauccap\endgroup\else\end{table*}\fi

\end{table*}
\endgroup

\begin{table*}[t]
  \caption{\textbf{Model comparison on BioLoomBench retrieval by domain.} Each entry is bidirectional mean Recall@1 (\%), averaging image-to-text (I2T) and text-to-image (T2I) retrieval. The three blocks use concise captions, original captions at native context length, and original captions with a common 77-token maximum. Short headers follow the 15-domain order listed in Sec.~\ref{app:external-retrieval-benchmark}. Figures from each listed domain are queries; the gallery contains all 15,000 figures. Biomedical model rows are ordered by mean Recall@1 across domains within each caption block. Our rows are shaded light purple. $^{\ast}$ denotes random vision-encoder initialization for image-text pretraining. Within each caption block, bold and underlined values mark the best and second-best biomedical-model scores per column, respectively.}
  \label{tab:bioloombench-domain-r1}
  \centering
  \resulttablefont
  \renewcommand{\arraystretch}{1.12}
  \setlength{\tabcolsep}{1pt}
  \newcommand{\domainheader}{\textbf{Model} & \textbf{Bio} & \textbf{Cell} & \textbf{Cyto} & \textbf{Derm} & \textbf{Dev} & \textbf{Gen} & \textbf{Histo} & \textbf{Micro} & \textbf{Mol} & \textbf{Neuro} & \textbf{Ophth} & \textbf{Phys} & \textbf{Plant} & \textbf{Radio} & \textbf{Surg} \\}
  \begin{tabularx}{\textwidth}{>{\raggedright\arraybackslash}p{0.18\textwidth}*{15}{Y}}
    \toprule
    \multicolumn{16}{l}{\textit{Concise caption}} \\
    \domainheader
    \midrule
    CoCa$^{\ast}$~\cite{yu2022cocacontrastivecaptionersimagetext} & \tabnum{38.20} & \tabnum{19.90} & \tabnum{14.15} & \tabnum{18.60} & \tabnum{22.15} & \tabnum{32.50} & \tabnum{10.75} & \tabnum{25.00} & \tabnum{19.30} & \tabnum{24.70} & \tabnum{17.75} & \tabnum{32.90} & \tabnum{36.75} & \tabnum{13.10} & \tabnum{23.80} \\
    OpenCLIP$^{\ast}$~\cite{cherti2023reproduciblescalinglawscontrastive} & \tabnum{40.40} & \tabnum{23.30} & \tabnum{18.20} & \tabnum{18.25} & \tabnum{27.70} & \tabnum{35.85} & \tabnum{14.35} & \tabnum{29.95} & \tabnum{21.55} & \tabnum{29.85} & \tabnum{18.95} & \tabnum{35.75} & \tabnum{40.25} & \tabnum{18.10} & \tabnum{27.60} \\
    \midrule
    \mbox{PMC-CLIP}$^{\ast}$~\cite{lin2023pmcclipcontrastivelanguageimagepretraining} & \tabnum{1.40} & \tabnum{1.05} & \tabnum{3.90} & \tabnum{4.80} & \tabnum{6.95} & \tabnum{0.35} & \tabnum{4.75} & \tabnum{1.15} & \tabnum{0.25} & \tabnum{4.50} & \tabnum{5.85} & \tabnum{3.85} & \tabnum{1.80} & \tabnum{25.70} & \tabnum{5.60} \\
    MedSigLIP~\cite{sellergren2025medgemmatechnicalreport} & \tabnum{32.05} & \tabnum{11.35} & \tabnum{9.25} & \tabnum{20.70} & \tabnum{24.10} & \tabnum{23.90} & \tabnum{11.50} & \tabnum{18.55} & \tabnum{15.30} & \tabnum{22.60} & \tabnum{14.15} & \tabnum{32.25} & \tabnum{36.05} & \tabnum{30.15} & \tabnum{26.40} \\
    \mbox{MedPMC-CLIP}~\cite{kim2026medpmc} & \tabnum{41.95} & \tabnum{33.00} & \tabnum{35.45} & \tabnum{42.35} & \tabnum{44.10} & \tabnum{27.30} & \tabnum{36.10} & \tabnum{34.55} & \tabnum{24.00} & \tabnum{46.55} & \tabnum{38.55} & \tabnum{46.00} & \tabnum{31.45} & \tabnum{55.50} & \tabnum{48.95} \\
    Panel2Patch~\cite{yuan2026panel2patch} & \tabnum{52.55} & \tabnum{35.40} & \tabnum{29.10} & \tabnum{35.80} & \tabnum{40.30} & \tabnum{43.30} & \tabnum{29.55} & \tabnum{41.35} & \tabnum{31.65} & \tabnum{43.85} & \tabnum{33.35} & \tabnum{49.05} & \tabnum{48.35} & \tabnum{39.70} & \tabnum{42.10} \\
    \mbox{BMC-LongCLIP}~\cite{sun2025tokenswastedleveraginglong} & \tabnum{46.60} & \tabnum{32.55} & \tabnum{34.10} & \tabnum{39.50} & \tabnum{40.90} & \tabnum{34.70} & \tabnum{38.35} & \tabnum{37.30} & \tabnum{28.75} & \tabnum{43.30} & \tabnum{38.20} & \tabnum{43.70} & \tabnum{35.05} & \tabnum{58.55} & \tabnum{47.35} \\
    \mbox{BMC-CLIP-CF}~\cite{lozano2025biomedica} & \tabnum{55.45} & \tabnum{38.95} & \tabnum{32.60} & \tabnum{37.30} & \tabnum{43.40} & \tabnum{46.05} & \tabnum{33.35} & \tabnum{43.10} & \tabnum{35.50} & \tabnum{47.30} & \tabnum{36.35} & \tabnum{53.05} & \tabnum{51.50} & \tabnum{44.30} & \tabnum{45.65} \\
    BiomedCLIP~\cite{zhang2023biomedclipmultimodalbiomedicalfoundation} & \tabnum{53.85} & \tabnum{41.75} & \tabnum{37.30} & \tabnum{37.75} & \tabnum{42.25} & \tabnum{54.45} & \tabnum{40.20} & \tabnum{50.15} & \tabnum{39.55} & \tabnum{48.80} & \tabnum{40.70} & \tabnum{55.80} & \tabnum{54.10} & \tabnum{53.15} & \tabnum{45.70} \\
    \rowcolor{bioloomlight}
    \methodname{} (Scratch)$^{\ast}$& \tabnum{\underline{65.50}} & \tabnum{\underline{57.45}} & \tabnum{\underline{48.85}} & \tabnum{\underline{50.75}} & \tabnum{\underline{56.95}} & \tabnum{\underline{59.55}} & \tabnum{\underline{50.80}} & \tabnum{\underline{60.10}} & \tabnum{\underline{50.40}} & \tabnum{\underline{61.80}} & \tabnum{\underline{49.80}} & \tabnum{\underline{60.80}} & \tabnum{\underline{59.55}} & \tabnum{\underline{65.95}} & \tabnum{\underline{58.75}} \\
    \rowcolor{bioloomlight}
    \methodname{} (DFN-2B)& \tabnum{\textbf{69.95}} & \tabnum{\textbf{60.45}} & \tabnum{\textbf{55.95}} & \tabnum{\textbf{59.45}} & \tabnum{\textbf{65.95}} & \tabnum{\textbf{65.30}} & \tabnum{\textbf{56.85}} & \tabnum{\textbf{65.50}} & \tabnum{\textbf{54.00}} & \tabnum{\textbf{68.70}} & \tabnum{\textbf{56.05}} & \tabnum{\textbf{67.40}} & \tabnum{\textbf{68.35}} & \tabnum{\textbf{69.45}} & \tabnum{\textbf{64.60}} \\
    \midrule
    \multicolumn{16}{l}{\textit{Original caption, native context length}} \\
    \domainheader
    \midrule
    CoCa$^{\ast}$~\cite{yu2022cocacontrastivecaptionersimagetext} & \tabnum{31.50} & \tabnum{18.85} & \tabnum{12.15} & \tabnum{12.40} & \tabnum{17.00} & \tabnum{22.65} & \tabnum{11.10} & \tabnum{18.80} & \tabnum{18.10} & \tabnum{20.05} & \tabnum{14.15} & \tabnum{27.45} & \tabnum{27.75} & \tabnum{7.55} & \tabnum{17.25} \\
    OpenCLIP$^{\ast}$~\cite{cherti2023reproduciblescalinglawscontrastive} & \tabnum{37.45} & \tabnum{27.50} & \tabnum{18.15} & \tabnum{14.50} & \tabnum{24.85} & \tabnum{28.40} & \tabnum{16.30} & \tabnum{26.30} & \tabnum{25.25} & \tabnum{28.30} & \tabnum{19.40} & \tabnum{35.25} & \tabnum{37.05} & \tabnum{13.25} & \tabnum{23.20} \\
    \midrule
    \mbox{PMC-CLIP}$^{\ast}$~\cite{lin2023pmcclipcontrastivelanguageimagepretraining} & \tabnum{1.40} & \tabnum{0.75} & \tabnum{4.80} & \tabnum{4.50} & \tabnum{7.35} & \tabnum{0.30} & \tabnum{5.85} & \tabnum{1.30} & \tabnum{0.65} & \tabnum{5.15} & \tabnum{6.55} & \tabnum{4.75} & \tabnum{1.55} & \tabnum{26.75} & \tabnum{7.25} \\
    MedSigLIP~\cite{sellergren2025medgemmatechnicalreport} & \tabnum{22.35} & \tabnum{6.85} & \tabnum{6.75} & \tabnum{13.85} & \tabnum{16.95} & \tabnum{14.40} & \tabnum{7.75} & \tabnum{10.75} & \tabnum{7.05} & \tabnum{14.40} & \tabnum{9.70} & \tabnum{22.55} & \tabnum{24.10} & \tabnum{22.20} & \tabnum{20.15} \\
    Panel2Patch~\cite{yuan2026panel2patch} & \tabnum{59.35} & \tabnum{52.15} & \tabnum{41.60} & \tabnum{39.20} & \tabnum{52.05} & \tabnum{44.15} & \tabnum{48.30} & \tabnum{47.25} & \tabnum{45.95} & \tabnum{55.25} & \tabnum{42.00} & \tabnum{58.85} & \tabnum{52.60} & \tabnum{44.00} & \tabnum{47.25} \\
    \mbox{BMC-CLIP-CF}~\cite{lozano2025biomedica} & \tabnum{61.35} & \tabnum{55.70} & \tabnum{45.55} & \tabnum{43.25} & \tabnum{55.80} & \tabnum{46.80} & \tabnum{51.05} & \tabnum{49.65} & \tabnum{48.85} & \tabnum{58.40} & \tabnum{44.40} & \tabnum{63.30} & \tabnum{54.55} & \tabnum{48.70} & \tabnum{52.50} \\
    \mbox{MedPMC-CLIP}~\cite{kim2026medpmc} & \tabnum{57.45} & \tabnum{58.40} & \tabnum{57.25} & \tabnum{52.25} & \tabnum{63.30} & \tabnum{36.00} & \tabnum{65.60} & \tabnum{49.75} & \tabnum{42.20} & \tabnum{64.05} & \tabnum{54.35} & \tabnum{67.15} & \tabnum{45.40} & \tabnum{64.65} & \tabnum{58.90} \\
    \mbox{BMC-LongCLIP}~\cite{sun2025tokenswastedleveraginglong} & \tabnum{68.90} & \tabnum{68.80} & \tabnum{71.05} & \tabnum{58.35} & \tabnum{76.20} & \tabnum{47.25} & \tabnum{79.85} & \tabnum{60.75} & \tabnum{67.90} & \tabnum{72.55} & \tabnum{66.80} & \tabnum{72.10} & \tabnum{57.75} & \tabnum{76.55} & \tabnum{64.10} \\
    BiomedCLIP~\cite{zhang2023biomedclipmultimodalbiomedicalfoundation} & \tabnum{72.40} & \tabnum{73.30} & \tabnum{69.25} & \tabnum{55.05} & \tabnum{73.50} & \tabnum{\underline{61.35}} & \tabnum{77.25} & \tabnum{67.65} & \tabnum{76.75} & \tabnum{73.80} & \tabnum{65.95} & \tabnum{78.40} & \tabnum{67.90} & \tabnum{69.35} & \tabnum{59.95} \\
    \rowcolor{bioloomlight}
    \methodname{} (Scratch)$^{\ast}$& \tabnum{\underline{78.15}} & \tabnum{\underline{79.70}} & \tabnum{\underline{77.20}} & \tabnum{\underline{64.80}} & \tabnum{\underline{82.05}} & \tabnum{59.95} & \tabnum{\underline{83.40}} & \tabnum{\underline{72.15}} & \tabnum{\underline{84.05}} & \tabnum{\underline{79.25}} & \tabnum{\underline{73.05}} & \tabnum{\underline{78.50}} & \tabnum{\underline{68.30}} & \tabnum{\underline{77.90}} & \tabnum{\underline{67.55}} \\
    \rowcolor{bioloomlight}
    \methodname{} (DFN-2B)& \tabnum{\textbf{82.05}} & \tabnum{\textbf{82.15}} & \tabnum{\textbf{79.45}} & \tabnum{\textbf{69.50}} & \tabnum{\textbf{83.70}} & \tabnum{\textbf{67.25}} & \tabnum{\textbf{84.90}} & \tabnum{\textbf{75.70}} & \tabnum{\textbf{85.00}} & \tabnum{\textbf{83.45}} & \tabnum{\textbf{76.35}} & \tabnum{\textbf{85.25}} & \tabnum{\textbf{74.15}} & \tabnum{\textbf{80.50}} & \tabnum{\textbf{72.80}} \\
    \midrule
    \multicolumn{16}{l}{\textit{Original caption, 77-token maximum}} \\
    \domainheader
    \midrule
    CoCa$^{\ast}$~\cite{yu2022cocacontrastivecaptionersimagetext} & \tabnum{31.50} & \tabnum{18.85} & \tabnum{12.15} & \tabnum{12.40} & \tabnum{17.00} & \tabnum{22.65} & \tabnum{11.10} & \tabnum{18.80} & \tabnum{18.10} & \tabnum{20.05} & \tabnum{14.15} & \tabnum{27.45} & \tabnum{27.75} & \tabnum{7.55} & \tabnum{17.25} \\
    OpenCLIP$^{\ast}$~\cite{cherti2023reproduciblescalinglawscontrastive} & \tabnum{37.45} & \tabnum{27.50} & \tabnum{18.15} & \tabnum{14.50} & \tabnum{24.85} & \tabnum{28.40} & \tabnum{16.30} & \tabnum{26.30} & \tabnum{25.25} & \tabnum{28.30} & \tabnum{19.40} & \tabnum{35.25} & \tabnum{37.05} & \tabnum{13.25} & \tabnum{23.20} \\
    \midrule
    \mbox{PMC-CLIP}$^{\ast}$~\cite{lin2023pmcclipcontrastivelanguageimagepretraining} & \tabnum{1.40} & \tabnum{0.75} & \tabnum{4.80} & \tabnum{4.50} & \tabnum{7.35} & \tabnum{0.30} & \tabnum{5.85} & \tabnum{1.30} & \tabnum{0.65} & \tabnum{5.15} & \tabnum{6.55} & \tabnum{4.75} & \tabnum{1.55} & \tabnum{26.75} & \tabnum{7.25} \\
    MedSigLIP~\cite{sellergren2025medgemmatechnicalreport} & \tabnum{22.35} & \tabnum{6.85} & \tabnum{6.75} & \tabnum{13.85} & \tabnum{16.95} & \tabnum{14.40} & \tabnum{7.75} & \tabnum{10.75} & \tabnum{7.05} & \tabnum{14.40} & \tabnum{9.70} & \tabnum{22.55} & \tabnum{24.10} & \tabnum{22.20} & \tabnum{20.15} \\
    \mbox{BMC-LongCLIP}~\cite{sun2025tokenswastedleveraginglong} & \tabnum{46.70} & \tabnum{32.00} & \tabnum{38.70} & \tabnum{45.00} & \tabnum{42.60} & \tabnum{32.50} & \tabnum{48.70} & \tabnum{31.55} & \tabnum{28.25} & \tabnum{44.80} & \tabnum{44.75} & \tabnum{48.45} & \tabnum{35.25} & \tabnum{65.45} & \tabnum{53.90} \\
    Panel2Patch~\cite{yuan2026panel2patch} & \tabnum{59.35} & \tabnum{52.15} & \tabnum{41.60} & \tabnum{39.20} & \tabnum{52.05} & \tabnum{44.15} & \tabnum{48.30} & \tabnum{47.25} & \tabnum{45.95} & \tabnum{55.25} & \tabnum{42.00} & \tabnum{58.85} & \tabnum{52.60} & \tabnum{44.00} & \tabnum{47.25} \\
    BiomedCLIP~\cite{zhang2023biomedclipmultimodalbiomedicalfoundation} & \tabnum{53.35} & \tabnum{46.20} & \tabnum{44.55} & \tabnum{43.10} & \tabnum{48.25} & \tabnum{\underline{48.70}} & \tabnum{51.15} & \tabnum{44.85} & \tabnum{41.30} & \tabnum{49.85} & \tabnum{46.60} & \tabnum{61.95} & \tabnum{50.80} & \tabnum{59.05} & \tabnum{49.85} \\
    \mbox{BMC-CLIP-CF}~\cite{lozano2025biomedica} & \tabnum{\underline{61.35}} & \tabnum{55.70} & \tabnum{45.55} & \tabnum{43.25} & \tabnum{55.80} & \tabnum{46.80} & \tabnum{51.05} & \tabnum{49.65} & \tabnum{48.85} & \tabnum{58.40} & \tabnum{44.40} & \tabnum{63.30} & \tabnum{\underline{54.55}} & \tabnum{48.70} & \tabnum{52.50} \\
    \mbox{MedPMC-CLIP}~\cite{kim2026medpmc} & \tabnum{57.45} & \tabnum{\underline{58.40}} & \tabnum{\underline{57.25}} & \tabnum{\underline{52.25}} & \tabnum{\underline{63.30}} & \tabnum{36.00} & \tabnum{\textbf{65.60}} & \tabnum{\underline{49.75}} & \tabnum{42.20} & \tabnum{\underline{64.05}} & \tabnum{\underline{54.35}} & \tabnum{\underline{67.15}} & \tabnum{45.40} & \tabnum{64.65} & \tabnum{\underline{58.90}} \\
    \rowcolor{bioloomlight}
    \methodname{} (Scratch)$^{\ast}$& \tabnum{59.30} & \tabnum{52.65} & \tabnum{52.00} & \tabnum{51.50} & \tabnum{57.20} & \tabnum{47.35} & \tabnum{57.25} & \tabnum{48.55} & \tabnum{\underline{49.20}} & \tabnum{56.55} & \tabnum{53.50} & \tabnum{60.75} & \tabnum{49.85} & \tabnum{\underline{67.10}} & \tabnum{56.50} \\
    \rowcolor{bioloomlight}
    \methodname{} (DFN-2B)& \tabnum{\textbf{66.35}} & \tabnum{\textbf{61.25}} & \tabnum{\textbf{60.20}} & \tabnum{\textbf{59.30}} & \tabnum{\textbf{65.45}} & \tabnum{\textbf{55.00}} & \tabnum{\underline{65.05}} & \tabnum{\textbf{57.10}} & \tabnum{\textbf{57.10}} & \tabnum{\textbf{65.00}} & \tabnum{\textbf{58.80}} & \tabnum{\textbf{71.90}} & \tabnum{\textbf{59.25}} & \tabnum{\textbf{70.05}} & \tabnum{\textbf{63.95}} \\
    \bottomrule
  \end{tabularx}

\end{table*}

\begin{table*}[t]
  \caption{\textbf{Model comparison on BioLoomBench retrieval by visual content.} Entries are bidirectional mean Recall@1 (\%), averaging image-to-text (I2T) and text-to-image (T2I) retrieval. The three blocks show concise captions, original captions at native context length, and original captions with a common 77-token maximum. Queries are restricted to each listed content category; the gallery contains all 15,000 figures. Biomedical model rows are ordered by mean Recall@1 across content categories within each caption block. Our rows are shaded light purple. $^{\ast}$ denotes random vision-encoder initialization for image-text pretraining. Among biomedical models, bold and underlined values indicate the best and second-best scores in each column, respectively. Ranks are computed within each caption block.}
  \label{tab:bioloombench-content-r1}
  \centering
  \resulttablefont
  \renewcommand{\arraystretch}{1.12}
  \setlength{\tabcolsep}{5pt}
  \begin{tabularx}{\textwidth}{>{\raggedright\arraybackslash}p{0.22\textwidth}*{4}{Y}}
    \toprule
    \multicolumn{5}{l}{\textit{Concise caption}} \\
    \textbf{Model} & \textbf{Biological image} & \textbf{Image + chart} & \textbf{Diagram} & \textbf{Chart only} \\
    \midrule
    CoCa$^{\ast}$~\cite{yu2022cocacontrastivecaptionersimagetext} & \tabnum{14.86} & \tabnum{16.24} & \tabnum{44.68} & \tabnum{44.53} \\
    OpenCLIP$^{\ast}$~\cite{cherti2023reproduciblescalinglawscontrastive} & \tabnum{17.65} & \tabnum{20.26} & \tabnum{48.16} & \tabnum{47.43} \\
    \midrule
    \mbox{PMC-CLIP}$^{\ast}$~\cite{lin2023pmcclipcontrastivelanguageimagepretraining} & \tabnum{11.47} & \tabnum{1.24} & \tabnum{0.46} & \tabnum{0.57} \\
    MedSigLIP~\cite{sellergren2025medgemmatechnicalreport} & \tabnum{19.79} & \tabnum{12.35} & \tabnum{39.24} & \tabnum{35.60} \\
    \mbox{MedPMC-CLIP}~\cite{kim2026medpmc} & \tabnum{43.73} & \tabnum{32.17} & \tabnum{43.98} & \tabnum{38.93} \\
    \mbox{BMC-LongCLIP}~\cite{sun2025tokenswastedleveraginglong} & \tabnum{43.54} & \tabnum{32.12} & \tabnum{50.78} & \tabnum{37.23} \\
    Panel2Patch~\cite{yuan2026panel2patch} & \tabnum{34.78} & \tabnum{32.28} & \tabnum{58.96} & \tabnum{52.77} \\
    \mbox{BMC-CLIP-CF}~\cite{lozano2025biomedica} & \tabnum{37.85} & \tabnum{35.86} & \tabnum{62.32} & \tabnum{55.10} \\
    BiomedCLIP~\cite{zhang2023biomedclipmultimodalbiomedicalfoundation} & \tabnum{40.77} & \tabnum{42.63} & \tabnum{54.92} & \tabnum{\underline{66.33}} \\
    \rowcolor{bioloomlight}
    \methodname{} (Scratch)$^{\ast}$& \tabnum{\underline{55.35}} & \tabnum{\underline{52.95}} & \tabnum{\underline{68.42}} & \tabnum{60.23} \\
    \rowcolor{bioloomlight}
    \methodname{} (DFN-2B)& \tabnum{\textbf{60.81}} & \tabnum{\textbf{59.38}} & \tabnum{\textbf{72.90}} & \tabnum{\textbf{69.77}} \\
    \midrule
    \multicolumn{5}{l}{\textit{Original caption, native context length}} \\
    \textbf{Model} & \textbf{Biological image} & \textbf{Image + chart} & \textbf{Diagram} & \textbf{Chart only} \\
    \midrule
    CoCa$^{\ast}$~\cite{yu2022cocacontrastivecaptionersimagetext} & \tabnum{9.35} & \tabnum{17.12} & \tabnum{35.22} & \tabnum{28.73} \\
    OpenCLIP$^{\ast}$~\cite{cherti2023reproduciblescalinglawscontrastive} & \tabnum{14.18} & \tabnum{24.97} & \tabnum{43.20} & \tabnum{34.53} \\
    \midrule
    \mbox{PMC-CLIP}$^{\ast}$~\cite{lin2023pmcclipcontrastivelanguageimagepretraining} & \tabnum{12.70} & \tabnum{1.32} & \tabnum{0.58} & \tabnum{0.23} \\
    MedSigLIP~\cite{sellergren2025medgemmatechnicalreport} & \tabnum{15.05} & \tabnum{7.78} & \tabnum{25.60} & \tabnum{20.17} \\
    Panel2Patch~\cite{yuan2026panel2patch} & \tabnum{40.47} & \tabnum{52.65} & \tabnum{59.96} & \tabnum{45.27} \\
    \mbox{MedPMC-CLIP}~\cite{kim2026medpmc} & \tabnum{56.09} & \tabnum{60.99} & \tabnum{53.10} & \tabnum{40.00} \\
    \mbox{BMC-CLIP-CF}~\cite{lozano2025biomedica} & \tabnum{44.36} & \tabnum{56.00} & \tabnum{62.86} & \tabnum{47.13} \\
    \mbox{BMC-LongCLIP}~\cite{sun2025tokenswastedleveraginglong} & \tabnum{67.36} & \tabnum{75.12} & \tabnum{65.94} & \tabnum{40.27} \\
    BiomedCLIP~\cite{zhang2023biomedclipmultimodalbiomedicalfoundation} & \tabnum{62.86} & \tabnum{78.27} & \tabnum{67.76} & \tabnum{\textbf{64.13}} \\
    \rowcolor{bioloomlight}
    \methodname{} (Scratch)$^{\ast}$& \tabnum{\underline{72.63}} & \tabnum{\underline{84.10}} & \tabnum{\underline{77.64}} & \tabnum{46.60} \\
    \rowcolor{bioloomlight}
    \methodname{} (DFN-2B)& \tabnum{\textbf{75.45}} & \tabnum{\textbf{85.85}} & \tabnum{\textbf{81.22}} & \tabnum{\underline{61.30}} \\
    \midrule
    \multicolumn{5}{l}{\textit{Original caption, 77-token maximum}} \\
    \textbf{Model} & \textbf{Biological image} & \textbf{Image + chart} & \textbf{Diagram} & \textbf{Chart only} \\
    \midrule
    CoCa$^{\ast}$~\cite{yu2022cocacontrastivecaptionersimagetext} & \tabnum{9.35} & \tabnum{17.12} & \tabnum{35.22} & \tabnum{28.73} \\
    OpenCLIP$^{\ast}$~\cite{cherti2023reproduciblescalinglawscontrastive} & \tabnum{14.18} & \tabnum{24.97} & \tabnum{43.20} & \tabnum{34.53} \\
    \midrule
    \mbox{PMC-CLIP}$^{\ast}$~\cite{lin2023pmcclipcontrastivelanguageimagepretraining} & \tabnum{12.70} & \tabnum{1.32} & \tabnum{0.58} & \tabnum{0.23} \\
    MedSigLIP~\cite{sellergren2025medgemmatechnicalreport} & \tabnum{15.05} & \tabnum{7.78} & \tabnum{25.60} & \tabnum{20.17} \\
    \mbox{BMC-LongCLIP}~\cite{sun2025tokenswastedleveraginglong} & \tabnum{51.32} & \tabnum{32.08} & \tabnum{51.98} & \tabnum{33.30} \\
    Panel2Patch~\cite{yuan2026panel2patch} & \tabnum{40.47} & \tabnum{52.65} & \tabnum{59.96} & \tabnum{45.27} \\
    BiomedCLIP~\cite{zhang2023biomedclipmultimodalbiomedicalfoundation} & \tabnum{48.25} & \tabnum{45.25} & \tabnum{54.80} & \tabnum{\textbf{58.87}} \\
    \mbox{MedPMC-CLIP}~\cite{kim2026medpmc} & \tabnum{56.09} & \tabnum{\textbf{60.99}} & \tabnum{53.10} & \tabnum{40.00} \\
    \mbox{BMC-CLIP-CF}~\cite{lozano2025biomedica} & \tabnum{44.36} & \tabnum{56.00} & \tabnum{62.86} & \tabnum{47.13} \\
    \rowcolor{bioloomlight}
    \methodname{} (Scratch)$^{\ast}$& \tabnum{\underline{57.48}} & \tabnum{50.74} & \tabnum{\underline{65.32}} & \tabnum{40.50} \\
    \rowcolor{bioloomlight}
    \methodname{} (DFN-2B)& \tabnum{\textbf{62.52}} & \tabnum{\underline{60.83}} & \tabnum{\textbf{71.20}} & \tabnum{\underline{52.90}} \\
    \bottomrule
  \end{tabularx}

\end{table*}

\ifdefined\jointbootstrap\begingroup\else\begin{table*}[t]\fi
  \caption{\textbf{Model comparison on BioMedFlickr under the original exact-pair protocol.} We report our evaluations of all models; only the paired row is relevant for Recall@$K$ (\%). Biomedical model rows are ordered by bidirectional mean Recall@1 from low to high. Our rows have a light-purple background. I2T and T2I denote image-to-text and text-to-image retrieval, respectively. $^{\ast}$ denotes random vision-encoder initialization for image-text pretraining. Among biomedical models, bold and underlined values indicate the best and second-best scores in each column, respectively.}
  \label{tab:retrieval-official-exact}
  \centering
  \resulttablefont
  \renewcommand{\arraystretch}{1.12}
  \setlength{\tabcolsep}{4pt}
  \begin{tabularx}{\textwidth}{>{\raggedright\arraybackslash}p{0.22\textwidth}*{6}{Y}}
    \toprule
    & \multicolumn{3}{c}{\textbf{Image $\rightarrow$ Text}} & \multicolumn{3}{c}{\textbf{Text $\rightarrow$ Image}} \\
    \cmidrule(lr){2-4}\cmidrule(lr){5-7}
    \textbf{Model} & R@1 & R@10 & R@100 & R@1 & R@10 & R@100 \\
    \midrule
    Random & \tabnum{0.01} & \tabnum{0.14} & \tabnum{1.39} & \tabnum{0.01} & \tabnum{0.14} & \tabnum{1.39} \\
    CoCa$^{\ast}$~\cite{yu2022cocacontrastivecaptionersimagetext} & \tabnum{1.41} & \tabnum{5.54} & \tabnum{15.27} & \tabnum{1.70} & \tabnum{5.44} & \tabnum{15.94} \\
    OpenCLIP$^{\ast}$~\cite{cherti2023reproduciblescalinglawscontrastive} & \tabnum{2.21} & \tabnum{8.11} & \tabnum{23.30} & \tabnum{2.07} & \tabnum{7.95} & \tabnum{23.77} \\
    \midrule
    \mbox{PMC-CLIP}$^{\ast}$~\cite{lin2023pmcclipcontrastivelanguageimagepretraining} & \tabnum{1.68} & \tabnum{8.11} & \tabnum{25.50} & \tabnum{1.46} & \tabnum{7.52} & \tabnum{24.04} \\
    MedSigLIP~\cite{sellergren2025medgemmatechnicalreport} & \tabnum{2.76} & \tabnum{10.33} & \tabnum{31.23} & \tabnum{2.31} & \tabnum{9.21} & \tabnum{30.79} \\
    Panel2Patch~\cite{yuan2026panel2patch} & \tabnum{3.30} & \tabnum{12.03} & \tabnum{32.82} & \tabnum{3.55} & \tabnum{13.12} & \tabnum{35.70} \\
    BiomedCLIP~\cite{zhang2023biomedclipmultimodalbiomedicalfoundation} & \tabnum{3.92} & \tabnum{13.43} & \tabnum{35.25} & \tabnum{3.35} & \tabnum{12.48} & \tabnum{35.91} \\
    \mbox{BMC-CLIP-CF}~\cite{lozano2025biomedica} & \tabnum{4.02} & \tabnum{13.50} & \tabnum{35.48} & \tabnum{3.97} & \tabnum{14.98} & \tabnum{37.63} \\
    \mbox{BMC-LongCLIP}~\cite{sun2025tokenswastedleveraginglong} & \tabnum{4.26} & \tabnum{14.89} & \tabnum{37.30} & \tabnum{4.22} & \tabnum{14.64} & \tabnum{38.86} \\
    \mbox{MedPMC-CLIP}~\cite{kim2026medpmc} & \tabnum{5.50} & \tabnum{19.25} & \tabnum{44.89} & \tabnum{5.97} & \tabnum{\underline{20.57}} & \tabnum{\underline{49.10}} \\
    \midrule
    \rowcolor{bioloomlight}
    \methodname{} (Scratch)$^{\ast}$& \tabnum{\underline{7.31}} & \tabnum{\underline{21.29}} & \tabnum{\underline{45.26}} & \tabnum{\underline{6.65}} & \tabnum{\underline{20.57}} & \tabnum{45.85} \\
    \rowcolor{bioloomlight}
    \methodname{} (DFN-2B)& \tabnum{\textbf{8.24}} & \tabnum{\textbf{24.54}} & \tabnum{\textbf{49.24}} & \tabnum{\textbf{8.56}} & \tabnum{\textbf{24.29}} & \tabnum{\textbf{50.70}} \\
    \bottomrule
  \end{tabularx}

\ifdefined\jointbootstrap\endgroup\else\end{table*}\fi

\clearpage

\ifdefined\jointbootstrap\begingroup\else\begin{table*}[p]\fi
\centering
\caption{\textbf{Bootstrap intervals for zero-shot classification.} Each cell reports accuracy (\%) on the first line and its 95\% within-task bootstrap confidence interval below. Task composition and aggregation weights remain fixed.}
\label{tab:bootstrap-classification}
\resulttablefont
\setlength{\tabcolsep}{3pt}
\renewcommand{\arraystretch}{1.15}
\begin{tabularx}{\textwidth}{@{}l*{6}{Y}@{}}
\toprule
Model & Cell & Structure & Microscopy & Cytology & Histopath. & Radiology \\
\midrule
CoCa & \shortstack{\tabnum{22.18}\\{}\tabci{21.30}{22.97}} & \shortstack{\tabnum{33.45}\\{}\tabci{31.55}{35.43}} & \shortstack{\tabnum{56.86}\\{}\tabci{53.76}{59.82}} & \shortstack{\tabnum{41.56}\\{}\tabci{40.40}{42.67}} & \shortstack{\tabnum{39.80}\\{}\tabci{38.52}{41.04}} & \shortstack{\tabnum{36.31}\\{}\tabci{35.00}{37.63}} \\
OpenCLIP & \shortstack{\tabnum{24.49}\\{}\tabci{23.54}{25.44}} & \shortstack{\tabnum{44.44}\\{}\tabci{42.40}{46.27}} & \shortstack{\tabnum{59.04}\\{}\tabci{55.95}{61.99}} & \shortstack{\tabnum{10.25}\\{}\tabci{9.57}{10.95}} & \shortstack{\tabnum{41.03}\\{}\tabci{39.77}{42.24}} & \shortstack{\tabnum{64.86}\\{}\tabci{63.17}{66.50}} \\
\midrule
\mbox{PMC-CLIP} & \shortstack{\tabnum{28.16}\\{}\tabci{27.25}{29.06}} & \shortstack{\tabnum{48.95}\\{}\tabci{47.03}{50.90}} & \shortstack{\tabnum{69.20}\\{}\tabci{67.64}{70.52}} & \shortstack{\tabnum{15.16}\\{}\tabci{14.36}{15.95}} & \shortstack{\tabnum{57.30}\\{}\tabci{56.07}{58.47}} & \shortstack{\tabnum{53.91}\\{}\tabci{52.22}{55.71}} \\
MedSigLIP & \shortstack{\tabnum{22.74}\\{}\tabci{21.79}{23.65}} & \shortstack{\tabnum{44.77}\\{}\tabci{42.87}{46.49}} & \shortstack{\tabnum{65.31}\\{}\tabci{64.05}{66.28}} & \shortstack{\tabnum{37.28}\\{}\tabci{36.19}{38.34}} & \shortstack{\tabnum{38.77}\\{}\tabci{37.70}{39.85}} & \shortstack{\tabnum{49.26}\\{}\tabci{47.56}{51.08}} \\
BiomedCLIP & \shortstack{\tabnum{24.11}\\{}\tabci{23.20}{24.99}} & \shortstack{\tabnum{49.59}\\{}\tabci{47.50}{51.54}} & \shortstack{\tabnum{74.55}\\{}\tabci{73.56}{75.34}} & \shortstack{\tabnum{11.16}\\{}\tabci{10.45}{11.91}} & \shortstack{\tabnum{53.02}\\{}\tabci{51.80}{54.33}} & \shortstack{\tabnum{55.36}\\{}\tabci{53.62}{57.08}} \\
    Panel2Patch & \shortstack{\tabnum{29.40}\\{}\tabci{28.39}{30.40}} & \shortstack{\tabnum{47.12}\\{}\tabci{45.16}{49.07}} & \shortstack{\tabnum{75.66}\\{}\tabci{74.51}{76.54}} & \shortstack{\tabnum{9.50}\\{}\tabci{8.85}{10.18}} & \shortstack{\tabnum{52.30}\\{}\tabci{51.03}{53.70}} & \shortstack{\tabnum{63.31}\\{}\tabci{61.65}{64.92}} \\
\mbox{BMC-CLIP-CF} & \shortstack{\tabnum{24.79}\\{}\tabci{23.99}{25.71}} & \shortstack{\tabnum{56.53}\\{}\tabci{54.69}{58.37}} & \shortstack{\tabnum{69.39}\\{}\tabci{67.25}{71.26}} & \shortstack{\tabnum{8.97}\\{}\tabci{8.38}{9.58}} & \shortstack{\tabnum{51.34}\\{}\tabci{50.06}{52.61}} & \shortstack{\tabnum{60.85}\\{}\tabci{59.22}{62.48}} \\
\mbox{BMC-LongCLIP} & \shortstack{\tabnum{25.62}\\{}\tabci{24.70}{26.55}} & \shortstack{\tabnum{51.49}\\{}\tabci{49.55}{53.44}} & \shortstack{\tabnum{76.23}\\{}\tabci{74.95}{77.21}} & \shortstack{\tabnum{35.61}\\{}\tabci{34.68}{36.58}} & \shortstack{\tabnum{53.94}\\{}\tabci{52.65}{55.20}} & \shortstack{\tabnum{60.73}\\{}\tabci{59.21}{62.15}} \\
\mbox{MedPMC-CLIP} & \shortstack{\tabnum{28.60}\\{}\tabci{27.60}{29.71}} & \shortstack{\tabnum{57.60}\\{}\tabci{55.73}{59.50}} & \shortstack{\tabnum{66.17}\\{}\tabci{63.08}{69.04}} & \shortstack{\tabnum{38.18}\\{}\tabci{37.17}{39.26}} & \shortstack{\tabnum{59.49}\\{}\tabci{58.21}{60.78}} & \shortstack{\tabnum{55.57}\\{}\tabci{53.95}{57.09}} \\
\midrule
\rowcolor{bioloomlight}
BioLoom (Scratch) & \shortstack{\tabnum{34.22}\\{}\tabci{33.29}{35.15}} & \shortstack{\tabnum{56.69}\\{}\tabci{54.90}{58.43}} & \shortstack{\tabnum{79.70}\\{}\tabci{79.17}{80.05}} & \shortstack{\tabnum{37.96}\\{}\tabci{36.86}{39.09}} & \shortstack{\tabnum{61.81}\\{}\tabci{60.61}{63.07}} & \shortstack{\tabnum{64.57}\\{}\tabci{62.95}{66.16}} \\
\rowcolor{bioloomlight}
BioLoom (DFN-2B) & \shortstack{\tabnum{31.05}\\{}\tabci{30.13}{32.00}} & \shortstack{\tabnum{59.73}\\{}\tabci{58.15}{61.42}} & \shortstack{\tabnum{76.32}\\{}\tabci{76.16}{76.49}} & \shortstack{\tabnum{45.61}\\{}\tabci{44.61}{46.58}} & \shortstack{\tabnum{62.55}\\{}\tabci{61.24}{63.78}} & \shortstack{\tabnum{61.40}\\{}\tabci{60.07}{62.71}} \\
\midrule
Model & Ophthalm. & Dermatology & Surgery & Microbial & Overall & \\
\midrule
CoCa & \shortstack{\tabnum{50.53}\\{}\tabci{48.75}{52.41}} & \shortstack{\tabnum{17.37}\\{}\tabci{16.41}{18.38}} & \shortstack{\tabnum{23.05}\\{}\tabci{22.63}{23.47}} & \shortstack{\tabnum{31.48}\\{}\tabci{29.75}{33.34}} & \shortstack{\tabnum{35.26}\\{}\tabci{34.76}{35.76}} & \\
OpenCLIP & \shortstack{\tabnum{42.80}\\{}\tabci{41.07}{44.60}} & \shortstack{\tabnum{15.67}\\{}\tabci{14.70}{16.59}} & \shortstack{\tabnum{31.26}\\{}\tabci{30.89}{31.62}} & \shortstack{\tabnum{44.87}\\{}\tabci{43.05}{46.72}} & \shortstack{\tabnum{37.87}\\{}\tabci{37.36}{38.37}} & \\
\midrule
\mbox{PMC-CLIP} & \shortstack{\tabnum{38.74}\\{}\tabci{36.95}{40.43}} & \shortstack{\tabnum{16.34}\\{}\tabci{15.32}{17.30}} & \shortstack{\tabnum{29.09}\\{}\tabci{28.37}{29.81}} & \shortstack{\tabnum{38.16}\\{}\tabci{36.06}{40.21}} & \shortstack{\tabnum{39.50}\\{}\tabci{39.06}{39.98}} & \\
MedSigLIP & \shortstack{\tabnum{49.24}\\{}\tabci{47.45}{51.17}} & \shortstack{\tabnum{20.62}\\{}\tabci{19.66}{21.63}} & \shortstack{\tabnum{19.38}\\{}\tabci{18.81}{20.00}} & \shortstack{\tabnum{58.58}\\{}\tabci{56.62}{60.91}} & \shortstack{\tabnum{40.59}\\{}\tabci{40.14}{41.04}} & \\
BiomedCLIP & \shortstack{\tabnum{38.27}\\{}\tabci{36.38}{39.98}} & \shortstack{\tabnum{30.48}\\{}\tabci{29.29}{31.67}} & \shortstack{\tabnum{41.83}\\{}\tabci{41.08}{42.61}} & \shortstack{\tabnum{43.63}\\{}\tabci{41.83}{45.53}} & \shortstack{\tabnum{42.20}\\{}\tabci{41.75}{42.64}} & \\
    Panel2Patch & \shortstack{\tabnum{47.56}\\{}\tabci{45.62}{49.33}} & \shortstack{\tabnum{43.76}\\{}\tabci{42.70}{44.80}} & \shortstack{\tabnum{37.34}\\{}\tabci{36.57}{38.03}} & \shortstack{\tabnum{27.77}\\{}\tabci{26.17}{29.51}} & \shortstack{\tabnum{43.37}\\{}\tabci{42.97}{43.78}} & \\
\mbox{BMC-CLIP-CF} & \shortstack{\tabnum{46.38}\\{}\tabci{44.52}{48.17}} & \shortstack{\tabnum{43.76}\\{}\tabci{42.64}{44.85}} & \shortstack{\tabnum{41.35}\\{}\tabci{40.56}{42.10}} & \shortstack{\tabnum{32.25}\\{}\tabci{30.68}{33.92}} & \shortstack{\tabnum{43.56}\\{}\tabci{43.10}{44.01}} & \\
\mbox{BMC-LongCLIP} & \shortstack{\tabnum{35.53}\\{}\tabci{33.84}{37.43}} & \shortstack{\tabnum{40.87}\\{}\tabci{39.60}{42.14}} & \shortstack{\tabnum{37.38}\\{}\tabci{36.53}{38.28}} & \shortstack{\tabnum{28.62}\\{}\tabci{26.83}{30.43}} & \shortstack{\tabnum{44.60}\\{}\tabci{44.15}{45.02}} & \\
\mbox{MedPMC-CLIP} & \shortstack{\tabnum{51.40}\\{}\tabci{49.51}{53.07}} & \shortstack{\tabnum{45.49}\\{}\tabci{44.30}{46.62}} & \shortstack{\tabnum{35.11}\\{}\tabci{34.30}{35.93}} & \shortstack{\tabnum{50.11}\\{}\tabci{48.42}{51.89}} & \shortstack{\tabnum{48.77}\\{}\tabci{48.27}{49.29}} & \\
\midrule
\rowcolor{bioloomlight}
BioLoom (Scratch) & \shortstack{\tabnum{56.65}\\{}\tabci{54.70}{58.41}} & \shortstack{\tabnum{38.75}\\{}\tabci{37.53}{39.91}} & \shortstack{\tabnum{52.65}\\{}\tabci{51.85}{53.52}} & \shortstack{\tabnum{45.90}\\{}\tabci{44.16}{47.61}} & \shortstack{\tabnum{52.89}\\{}\tabci{52.47}{53.30}} & \\
\rowcolor{bioloomlight}
BioLoom (DFN-2B) & \shortstack{\tabnum{61.64}\\{}\tabci{59.72}{63.54}} & \shortstack{\tabnum{48.21}\\{}\tabci{46.99}{49.45}} & \shortstack{\tabnum{57.62}\\{}\tabci{56.77}{58.52}} & \shortstack{\tabnum{64.09}\\{}\tabci{62.11}{65.98}} & \shortstack{\tabnum{56.82}\\{}\tabci{56.40}{57.22}} & \\
\bottomrule
\end{tabularx}
\ifdefined\jointbootstrap\endgroup\else\end{table*}\fi

\begin{table*}[p]
\centering
\caption{\textbf{BioLoomBench concise-caption retrieval with 95\% query-bootstrap confidence intervals.} Each cell gives Recall@$K$ (\%) and its interval. The gallery and relevance rules match the main evaluation.}
\label{tab:bootstrap-concise}
\resulttablefont
\setlength{\tabcolsep}{3pt}
\renewcommand{\arraystretch}{1.15}
\begin{tabularx}{\textwidth}{@{}l*{6}{Y}@{}}
\toprule
 & \multicolumn{3}{c}{Image-to-text} & \multicolumn{3}{c}{Text-to-image} \\
\cmidrule(lr){2-4}\cmidrule(lr){5-7}
Model & R@1 & R@10 & R@100 & R@1 & R@10 & R@100 \\
\midrule
CoCa & \shortstack{\tabnum{24.25}\\{}\tabci{23.57}{24.94}} & \shortstack{\tabnum{46.37}\\{}\tabci{45.57}{47.21}} & \shortstack{\tabnum{74.44}\\{}\tabci{73.78}{75.14}} & \shortstack{\tabnum{22.35}\\{}\tabci{21.70}{23.02}} & \shortstack{\tabnum{42.85}\\{}\tabci{42.00}{43.62}} & \shortstack{\tabnum{71.15}\\{}\tabci{70.43}{71.92}} \\
OpenCLIP & \shortstack{\tabnum{27.65}\\{}\tabci{26.97}{28.35}} & \shortstack{\tabnum{50.47}\\{}\tabci{49.65}{51.27}} & \shortstack{\tabnum{76.93}\\{}\tabci{76.23}{77.57}} & \shortstack{\tabnum{25.69}\\{}\tabci{24.97}{26.38}} & \shortstack{\tabnum{47.82}\\{}\tabci{47.07}{48.60}} & \shortstack{\tabnum{75.24}\\{}\tabci{74.55}{75.90}} \\
\midrule
\mbox{PMC-CLIP} & \shortstack{\tabnum{4.75}\\{}\tabci{4.43}{5.11}} & \shortstack{\tabnum{14.76}\\{}\tabci{14.20}{15.31}} & \shortstack{\tabnum{33.11}\\{}\tabci{32.37}{33.84}} & \shortstack{\tabnum{4.84}\\{}\tabci{4.52}{5.18}} & \shortstack{\tabnum{15.87}\\{}\tabci{15.29}{16.51}} & \shortstack{\tabnum{35.96}\\{}\tabci{35.21}{36.73}} \\
MedSigLIP & \shortstack{\tabnum{27.39}\\{}\tabci{26.70}{28.10}} & \shortstack{\tabnum{56.07}\\{}\tabci{55.31}{56.86}} & \shortstack{\tabnum{82.69}\\{}\tabci{82.12}{83.27}} & \shortstack{\tabnum{16.38}\\{}\tabci{15.81}{16.92}} & \shortstack{\tabnum{38.63}\\{}\tabci{37.86}{39.37}} & \shortstack{\tabnum{67.55}\\{}\tabci{66.84}{68.29}} \\
\mbox{MedPMC-CLIP} & \shortstack{\tabnum{41.27}\\{}\tabci{40.49}{42.05}} & \shortstack{\tabnum{70.51}\\{}\tabci{69.69}{71.23}} & \shortstack{\tabnum{89.33}\\{}\tabci{88.87}{89.85}} & \shortstack{\tabnum{36.84}\\{}\tabci{36.06}{37.58}} & \shortstack{\tabnum{63.63}\\{}\tabci{62.89}{64.37}} & \shortstack{\tabnum{84.71}\\{}\tabci{84.14}{85.31}} \\
    Panel2Patch & \shortstack{\tabnum{42.43}\\{}\tabci{41.61}{43.17}} & \shortstack{\tabnum{71.55}\\{}\tabci{70.85}{72.27}} & \shortstack{\tabnum{92.02}\\{}\tabci{91.62}{92.44}} & \shortstack{\tabnum{36.96}\\{}\tabci{36.19}{37.71}} & \shortstack{\tabnum{64.62}\\{}\tabci{63.86}{65.42}} & \shortstack{\tabnum{87.45}\\{}\tabci{86.92}{87.99}} \\
\mbox{BMC-LongCLIP} & \shortstack{\tabnum{44.39}\\{}\tabci{43.53}{45.20}} & \shortstack{\tabnum{78.15}\\{}\tabci{77.49}{78.81}} & \shortstack{\tabnum{95.83}\\{}\tabci{95.49}{96.14}} & \shortstack{\tabnum{35.47}\\{}\tabci{34.71}{36.25}} & \shortstack{\tabnum{68.14}\\{}\tabci{67.41}{68.90}} & \shortstack{\tabnum{90.77}\\{}\tabci{90.31}{91.25}} \\
\mbox{BMC-CLIP-CF} & \shortstack{\tabnum{47.11}\\{}\tabci{46.33}{47.85}} & \shortstack{\tabnum{76.90}\\{}\tabci{76.21}{77.53}} & \shortstack{\tabnum{94.85}\\{}\tabci{94.47}{95.19}} & \shortstack{\tabnum{38.74}\\{}\tabci{37.91}{39.46}} & \shortstack{\tabnum{66.15}\\{}\tabci{65.39}{66.83}} & \shortstack{\tabnum{88.41}\\{}\tabci{87.92}{88.90}} \\
BiomedCLIP & \shortstack{\tabnum{51.65}\\{}\tabci{50.91}{52.47}} & \shortstack{\tabnum{84.38}\\{}\tabci{83.79}{84.98}} & \shortstack{\tabnum{97.67}\\{}\tabci{97.43}{97.91}} & \shortstack{\tabnum{41.09}\\{}\tabci{40.27}{41.84}} & \shortstack{\tabnum{75.63}\\{}\tabci{74.96}{76.27}} & \shortstack{\tabnum{94.71}\\{}\tabci{94.37}{95.09}} \\
\midrule
\rowcolor{bioloomlight}
BioLoom (Scratch) & \shortstack{\tabnum{61.93}\\{}\tabci{61.20}{62.68}} & \shortstack{\tabnum{90.61}\\{}\tabci{90.14}{91.08}} & \shortstack{\tabnum{98.79}\\{}\tabci{98.61}{98.96}} & \shortstack{\tabnum{52.33}\\{}\tabci{51.53}{53.17}} & \shortstack{\tabnum{83.86}\\{}\tabci{83.29}{84.46}} & \shortstack{\tabnum{97.44}\\{}\tabci{97.18}{97.71}} \\
\rowcolor{bioloomlight}
BioLoom (DFN-2B) & \shortstack{\tabnum{66.81}\\{}\tabci{66.03}{67.54}} & \shortstack{\tabnum{92.81}\\{}\tabci{92.41}{93.22}} & \shortstack{\tabnum{99.28}\\{}\tabci{99.15}{99.41}} & \shortstack{\tabnum{59.59}\\{}\tabci{58.85}{60.35}} & \shortstack{\tabnum{88.81}\\{}\tabci{88.37}{89.31}} & \shortstack{\tabnum{98.77}\\{}\tabci{98.59}{98.94}} \\
\bottomrule
\end{tabularx}
\end{table*}

\begin{table*}[p]
\centering
\caption{\textbf{BioLoomBench original-caption retrieval with 95\% query-bootstrap confidence intervals.} Each cell gives Recall@$K$ (\%) and its interval. The gallery and relevance rules match the main evaluation.}
\label{tab:bootstrap-source}
\resulttablefont
\setlength{\tabcolsep}{3pt}
\renewcommand{\arraystretch}{1.15}
\begin{tabularx}{\textwidth}{@{}l*{6}{Y}@{}}
\toprule
 & \multicolumn{3}{c}{Image-to-text} & \multicolumn{3}{c}{Text-to-image} \\
\cmidrule(lr){2-4}\cmidrule(lr){5-7}
Model & R@1 & R@10 & R@100 & R@1 & R@10 & R@100 \\
\midrule
CoCa & \shortstack{\tabnum{18.37}\\{}\tabci{17.77}{18.98}} & \shortstack{\tabnum{36.88}\\{}\tabci{36.13}{37.59}} & \shortstack{\tabnum{64.68}\\{}\tabci{63.95}{65.46}} & \shortstack{\tabnum{18.53}\\{}\tabci{17.94}{19.13}} & \shortstack{\tabnum{36.72}\\{}\tabci{35.93}{37.45}} & \shortstack{\tabnum{62.48}\\{}\tabci{61.72}{63.25}} \\
OpenCLIP & \shortstack{\tabnum{25.34}\\{}\tabci{24.65}{26.01}} & \shortstack{\tabnum{47.37}\\{}\tabci{46.49}{48.17}} & \shortstack{\tabnum{73.49}\\{}\tabci{72.78}{74.23}} & \shortstack{\tabnum{24.68}\\{}\tabci{23.99}{25.33}} & \shortstack{\tabnum{45.66}\\{}\tabci{44.85}{46.41}} & \shortstack{\tabnum{71.75}\\{}\tabci{71.03}{72.49}} \\
\midrule
\mbox{PMC-CLIP} & \shortstack{\tabnum{4.47}\\{}\tabci{4.14}{4.81}} & \shortstack{\tabnum{13.69}\\{}\tabci{13.16}{14.26}} & \shortstack{\tabnum{30.34}\\{}\tabci{29.59}{31.05}} & \shortstack{\tabnum{6.05}\\{}\tabci{5.69}{6.43}} & \shortstack{\tabnum{18.04}\\{}\tabci{17.44}{18.65}} & \shortstack{\tabnum{37.51}\\{}\tabci{36.79}{38.28}} \\
MedSigLIP & \shortstack{\tabnum{19.88}\\{}\tabci{19.23}{20.53}} & \shortstack{\tabnum{44.66}\\{}\tabci{43.87}{45.47}} & \shortstack{\tabnum{72.75}\\{}\tabci{72.05}{73.46}} & \shortstack{\tabnum{9.43}\\{}\tabci{8.97}{9.91}} & \shortstack{\tabnum{26.52}\\{}\tabci{25.79}{27.27}} & \shortstack{\tabnum{53.27}\\{}\tabci{52.46}{54.01}} \\
    Panel2Patch & \shortstack{\tabnum{47.75}\\{}\tabci{46.95}{48.52}} & \shortstack{\tabnum{74.83}\\{}\tabci{74.14}{75.49}} & \shortstack{\tabnum{91.61}\\{}\tabci{91.20}{92.07}} & \shortstack{\tabnum{49.57}\\{}\tabci{48.75}{50.33}} & \shortstack{\tabnum{77.01}\\{}\tabci{76.30}{77.65}} & \shortstack{\tabnum{92.84}\\{}\tabci{92.43}{93.25}} \\
\mbox{BMC-CLIP-CF} & \shortstack{\tabnum{52.09}\\{}\tabci{51.32}{52.88}} & \shortstack{\tabnum{78.96}\\{}\tabci{78.27}{79.60}} & \shortstack{\tabnum{92.94}\\{}\tabci{92.53}{93.36}} & \shortstack{\tabnum{51.89}\\{}\tabci{51.09}{52.71}} & \shortstack{\tabnum{78.97}\\{}\tabci{78.33}{79.62}} & \shortstack{\tabnum{93.25}\\{}\tabci{92.87}{93.65}} \\
\mbox{MedPMC-CLIP} & \shortstack{\tabnum{55.42}\\{}\tabci{54.62}{56.21}} & \shortstack{\tabnum{80.70}\\{}\tabci{80.07}{81.31}} & \shortstack{\tabnum{93.03}\\{}\tabci{92.61}{93.41}} & \shortstack{\tabnum{56.14}\\{}\tabci{55.36}{56.95}} & \shortstack{\tabnum{81.63}\\{}\tabci{81.01}{82.25}} & \shortstack{\tabnum{93.35}\\{}\tabci{92.92}{93.73}} \\
\mbox{BMC-LongCLIP} & \shortstack{\tabnum{67.35}\\{}\tabci{66.57}{68.08}} & \shortstack{\tabnum{87.45}\\{}\tabci{86.91}{87.96}} & \shortstack{\tabnum{95.83}\\{}\tabci{95.52}{96.15}} & \shortstack{\tabnum{67.17}\\{}\tabci{66.37}{67.91}} & \shortstack{\tabnum{87.65}\\{}\tabci{87.15}{88.19}} & \shortstack{\tabnum{95.90}\\{}\tabci{95.58}{96.21}} \\
BiomedCLIP & \shortstack{\tabnum{69.80}\\{}\tabci{69.06}{70.56}} & \shortstack{\tabnum{89.67}\\{}\tabci{89.16}{90.18}} & \shortstack{\tabnum{97.11}\\{}\tabci{96.84}{97.39}} & \shortstack{\tabnum{69.11}\\{}\tabci{68.43}{69.81}} & \shortstack{\tabnum{89.48}\\{}\tabci{88.99}{89.95}} & \shortstack{\tabnum{96.97}\\{}\tabci{96.71}{97.24}} \\
\midrule
\rowcolor{bioloomlight}
BioLoom (Scratch) & \shortstack{\tabnum{75.91}\\{}\tabci{75.25}{76.57}} & \shortstack{\tabnum{92.94}\\{}\tabci{92.55}{93.33}} & \shortstack{\tabnum{98.09}\\{}\tabci{97.87}{98.31}} & \shortstack{\tabnum{74.22}\\{}\tabci{73.54}{74.97}} & \shortstack{\tabnum{91.52}\\{}\tabci{91.11}{92.01}} & \shortstack{\tabnum{97.43}\\{}\tabci{97.19}{97.67}} \\
\rowcolor{bioloomlight}
BioLoom (DFN-2B) & \shortstack{\tabnum{79.29}\\{}\tabci{78.63}{79.89}} & \shortstack{\tabnum{94.57}\\{}\tabci{94.24}{94.95}} & \shortstack{\tabnum{98.39}\\{}\tabci{98.17}{98.58}} & \shortstack{\tabnum{78.34}\\{}\tabci{77.68}{79.01}} & \shortstack{\tabnum{94.10}\\{}\tabci{93.73}{94.47}} & \shortstack{\tabnum{98.41}\\{}\tabci{98.21}{98.61}} \\
\bottomrule
\end{tabularx}
\end{table*}

\begin{table*}[p]
\centering
\caption{\textbf{BioMedFlickr retrieval with 95\% query-bootstrap confidence intervals.} Each cell gives Recall@$K$ (\%) and its interval. The gallery and relevance rules match the main evaluation.}
\label{tab:bootstrap-flickr}
\resulttablefont
\setlength{\tabcolsep}{3pt}
\renewcommand{\arraystretch}{1.15}
\begin{tabularx}{\textwidth}{@{}l*{6}{Y}@{}}
\toprule
 & \multicolumn{3}{c}{Image-to-text} & \multicolumn{3}{c}{Text-to-image} \\
\cmidrule(lr){2-4}\cmidrule(lr){5-7}
Model & R@1 & R@10 & R@100 & R@1 & R@10 & R@100 \\
\midrule
CoCa & \shortstack{\tabnum{1.64}\\{}\tabci{1.38}{1.93}} & \shortstack{\tabnum{5.58}\\{}\tabci{5.02}{6.11}} & \shortstack{\tabnum{15.30}\\{}\tabci{14.42}{16.10}} & \shortstack{\tabnum{2.06}\\{}\tabci{1.77}{2.38}} & \shortstack{\tabnum{7.95}\\{}\tabci{7.35}{8.60}} & \shortstack{\tabnum{23.98}\\{}\tabci{23.08}{24.96}} \\
OpenCLIP & \shortstack{\tabnum{2.41}\\{}\tabci{2.07}{2.76}} & \shortstack{\tabnum{8.48}\\{}\tabci{7.86}{9.12}} & \shortstack{\tabnum{23.42}\\{}\tabci{22.46}{24.40}} & \shortstack{\tabnum{3.13}\\{}\tabci{2.71}{3.54}} & \shortstack{\tabnum{11.36}\\{}\tabci{10.66}{12.04}} & \shortstack{\tabnum{34.25}\\{}\tabci{33.21}{35.39}} \\
\midrule
\mbox{PMC-CLIP} & \shortstack{\tabnum{2.03}\\{}\tabci{1.71}{2.37}} & \shortstack{\tabnum{8.38}\\{}\tabci{7.75}{9.03}} & \shortstack{\tabnum{25.61}\\{}\tabci{24.63}{26.62}} & \shortstack{\tabnum{2.67}\\{}\tabci{2.31}{3.02}} & \shortstack{\tabnum{11.75}\\{}\tabci{11.05}{12.50}} & \shortstack{\tabnum{34.38}\\{}\tabci{33.26}{35.48}} \\
MedSigLIP & \shortstack{\tabnum{3.56}\\{}\tabci{3.15}{3.98}} & \shortstack{\tabnum{10.83}\\{}\tabci{10.13}{11.55}} & \shortstack{\tabnum{31.45}\\{}\tabci{30.47}{32.51}} & \shortstack{\tabnum{4.62}\\{}\tabci{4.16}{5.08}} & \shortstack{\tabnum{15.18}\\{}\tabci{14.29}{15.99}} & \shortstack{\tabnum{38.78}\\{}\tabci{37.68}{39.87}} \\
BiomedCLIP & \shortstack{\tabnum{4.76}\\{}\tabci{4.27}{5.28}} & \shortstack{\tabnum{14.04}\\{}\tabci{13.26}{14.92}} & \shortstack{\tabnum{35.46}\\{}\tabci{34.43}{36.48}} & \shortstack{\tabnum{4.68}\\{}\tabci{4.22}{5.14}} & \shortstack{\tabnum{19.05}\\{}\tabci{18.18}{19.93}} & \shortstack{\tabnum{50.45}\\{}\tabci{49.23}{51.59}} \\
    Panel2Patch & \shortstack{\tabnum{3.60}\\{}\tabci{3.19}{4.06}} & \shortstack{\tabnum{12.40}\\{}\tabci{11.64}{13.12}} & \shortstack{\tabnum{33.00}\\{}\tabci{31.89}{34.04}} & \shortstack{\tabnum{7.00}\\{}\tabci{6.46}{7.60}} & \shortstack{\tabnum{19.82}\\{}\tabci{18.93}{20.72}} & \shortstack{\tabnum{46.93}\\{}\tabci{45.82}{48.14}} \\
\mbox{BMC-LongCLIP} & \shortstack{\tabnum{5.00}\\{}\tabci{4.50}{5.48}} & \shortstack{\tabnum{15.57}\\{}\tabci{14.75}{16.37}} & \shortstack{\tabnum{37.52}\\{}\tabci{36.40}{38.62}} & \shortstack{\tabnum{6.67}\\{}\tabci{6.08}{7.18}} & \shortstack{\tabnum{20.97}\\{}\tabci{20.01}{21.98}} & \shortstack{\tabnum{52.32}\\{}\tabci{51.11}{53.36}} \\
\mbox{BMC-CLIP-CF} & \shortstack{\tabnum{4.86}\\{}\tabci{4.38}{5.34}} & \shortstack{\tabnum{14.18}\\{}\tabci{13.39}{14.99}} & \shortstack{\tabnum{35.81}\\{}\tabci{34.68}{36.92}} & \shortstack{\tabnum{6.99}\\{}\tabci{6.42}{7.57}} & \shortstack{\tabnum{22.95}\\{}\tabci{22.00}{23.90}} & \shortstack{\tabnum{50.47}\\{}\tabci{49.35}{51.64}} \\
\mbox{MedPMC-CLIP} & \shortstack{\tabnum{5.94}\\{}\tabci{5.39}{6.51}} & \shortstack{\tabnum{19.94}\\{}\tabci{19.04}{20.92}} & \shortstack{\tabnum{45.26}\\{}\tabci{44.13}{46.49}} & \shortstack{\tabnum{9.16}\\{}\tabci{8.50}{9.78}} & \shortstack{\tabnum{31.61}\\{}\tabci{30.51}{32.78}} & \shortstack{\tabnum{61.38}\\{}\tabci{60.28}{62.46}} \\
\midrule
\rowcolor{bioloomlight}
BioLoom (Scratch) & \shortstack{\tabnum{8.00}\\{}\tabci{7.36}{8.60}} & \shortstack{\tabnum{21.85}\\{}\tabci{20.93}{22.83}} & \shortstack{\tabnum{45.66}\\{}\tabci{44.55}{46.92}} & \shortstack{\tabnum{8.84}\\{}\tabci{8.18}{9.52}} & \shortstack{\tabnum{27.63}\\{}\tabci{26.56}{28.70}} & \shortstack{\tabnum{58.75}\\{}\tabci{57.56}{59.92}} \\
\rowcolor{bioloomlight}
BioLoom (DFN-2B) & \shortstack{\tabnum{9.73}\\{}\tabci{8.99}{10.38}} & \shortstack{\tabnum{25.58}\\{}\tabci{24.50}{26.54}} & \shortstack{\tabnum{49.46}\\{}\tabci{48.34}{50.54}} & \shortstack{\tabnum{11.45}\\{}\tabci{10.72}{12.14}} & \shortstack{\tabnum{31.90}\\{}\tabci{30.84}{33.03}} & \shortstack{\tabnum{63.51}\\{}\tabci{62.37}{64.55}} \\
\bottomrule
\end{tabularx}
\end{table*}

\clearpage
\clearpage

\begingroup
\def\jointlinearprobes{}
\begin{table*}[!p]

\ifdefined\jointlinearprobes\begingroup\else\begin{table*}[t]\fi
\caption{\textbf{Accuracy of linear probes on MedMNIST.} Scores are percentages; column names omit the MNIST suffix. The mean gives equal weight to all 12 datasets. Bold and underlined values indicate the best and second-best scores in each column, respectively.}
\label{tab:linear-probe-acc}
\centering
\resulttablefont
\renewcommand{\arraystretch}{1.0}
\setlength{\tabcolsep}{1.3pt}
\begin{tabularx}{\textwidth}{@{}>{\raggedright\arraybackslash}p{0.22\textwidth}*{13}{Y}@{}}
\toprule
Model & Path & Chest & Derma & OCT & Pneu. & Blood & Tissue & OrganA & OrganC & OrganS & Retina & Breast & Mean \\
\midrule
CoCa ViT-L/14 & \tabnum{90.36} & \tabnum{94.73} & \tabnum{81.75} & \tabnum{68.80} & \tabnum{83.97} & \tabnum{93.63} & \tabnum{58.87} & \tabnum{87.42} & \tabnum{81.32} & \tabnum{72.84} & \tabnum{60.50} & \tabnum{81.41} & \tabnum{79.63} \\
EVA-CLIP ViT-L/14~\cite{sun2023evaclip} & \tabnum{88.09} & \tabnum{94.74} & \tabnum{80.55} & \tabnum{70.70} & \tabnum{80.45} & \tabnum{94.48} & \tabnum{61.08} & \tabnum{89.13} & \tabnum{83.24} & \tabnum{74.78} & \tabnum{61.50} & \tabnum{82.69} & \tabnum{80.12} \\
RetCCL ResNet-50~\cite{wang2023retccl} & \tabnum{\underline{94.53}} & \tabnum{94.74} & \tabnum{79.15} & \tabnum{58.70} & \tabnum{86.86} & \tabnum{\underline{97.92}} & \tabnum{59.33} & \tabnum{87.27} & \tabnum{84.07} & \tabnum{75.60} & \tabnum{63.25} & \tabnum{83.33} & \tabnum{80.40} \\
SigLIP2 ViT-L/16~\cite{tschannen2025siglip2} & \tabnum{91.85} & \tabnum{94.73} & \tabnum{79.30} & \tabnum{72.10} & \tabnum{88.30} & \tabnum{94.68} & \tabnum{59.63} & \tabnum{90.11} & \tabnum{82.95} & \tabnum{74.06} & \tabnum{61.00} & \tabnum{80.13} & \tabnum{80.74} \\
OpenCLIP (CommonPool) & \tabnum{88.55} & \tabnum{94.73} & \tabnum{82.24} & \tabnum{69.90} & \tabnum{83.01} & \tabnum{95.97} & \tabnum{62.16} & \tabnum{90.43} & \tabnum{84.35} & \tabnum{74.45} & \tabnum{60.75} & \tabnum{86.54} & \tabnum{81.09} \\
ImageNet-1k ResNet-50 & \tabnum{90.46} & \tabnum{94.73} & \tabnum{77.96} & \tabnum{71.20} & \tabnum{86.70} & \tabnum{95.50} & \tabnum{59.50} & \tabnum{91.89} & \tabnum{86.89} & \tabnum{76.12} & \tabnum{60.00} & \tabnum{\underline{87.18}} & \tabnum{81.51} \\
Phikon-v2 ViT-L~\cite{filiot2024phikonv2} & \tabnum{93.80} & \tabnum{94.75} & \tabnum{82.74} & \tabnum{71.10} & \tabnum{87.02} & \tabnum{97.84} & \tabnum{62.27} & \tabnum{87.48} & \tabnum{83.89} & \tabnum{76.27} & \tabnum{59.75} & \tabnum{83.33} & \tabnum{81.69} \\
SigLIP ViT-L/16 & \tabnum{94.32} & \tabnum{94.73} & \tabnum{83.14} & \tabnum{75.30} & \tabnum{87.82} & \tabnum{96.46} & \tabnum{60.85} & \tabnum{90.30} & \tabnum{83.33} & \tabnum{74.25} & \tabnum{61.25} & \tabnum{80.13} & \tabnum{81.82} \\
MetaCLIP ViT-L/14~\cite{xu2024metaclip} & \tabnum{91.87} & \tabnum{94.74} & \tabnum{82.94} & \tabnum{77.80} & \tabnum{87.34} & \tabnum{95.70} & \tabnum{62.29} & \tabnum{88.49} & \tabnum{82.42} & \tabnum{73.60} & \tabnum{62.00} & \tabnum{83.97} & \tabnum{81.93} \\
EVA-02 ViT-L/14~\cite{fang2024eva02} & \tabnum{94.09} & \tabnum{94.75} & \tabnum{80.45} & \tabnum{72.60} & \tabnum{86.06} & \tabnum{96.64} & \tabnum{60.08} & \tabnum{91.26} & \tabnum{88.18} & \tabnum{76.78} & \tabnum{59.50} & \tabnum{84.62} & \tabnum{82.08} \\
CLIP (DFN-2B) ViT-L/14 & \tabnum{92.12} & \tabnum{94.73} & \tabnum{\textbf{84.64}} & \tabnum{72.90} & \tabnum{85.42} & \tabnum{96.29} & \tabnum{61.99} & \tabnum{91.01} & \tabnum{85.55} & \tabnum{74.34} & \tabnum{60.75} & \tabnum{\textbf{89.10}} & \tabnum{82.40} \\
\mbox{PMC-CLIP} ResNet-50 & \tabnum{92.98} & \tabnum{\underline{94.77}} & \tabnum{81.30} & \tabnum{69.50} & \tabnum{\underline{91.35}} & \tabnum{96.84} & \tabnum{\underline{63.51}} & \tabnum{\textbf{92.57}} & \tabnum{\textbf{90.32}} & \tabnum{\underline{77.66}} & \tabnum{62.25} & \tabnum{83.97} & \tabnum{83.09} \\
DINOv2 ViT-L/14~\cite{oquab2024dinov2} & \tabnum{93.62} & \tabnum{94.74} & \tabnum{82.44} & \tabnum{79.50} & \tabnum{87.82} & \tabnum{96.61} & \tabnum{63.33} & \tabnum{90.17} & \tabnum{85.56} & \tabnum{76.79} & \tabnum{\underline{66.50}} & \tabnum{82.69} & \tabnum{83.31} \\
ImageNet-21k ViT-L/16 & \tabnum{93.97} & \tabnum{94.76} & \tabnum{\underline{83.54}} & \tabnum{\underline{79.90}} & \tabnum{87.50} & \tabnum{\textbf{98.22}} & \tabnum{\textbf{64.12}} & \tabnum{\underline{92.30}} & \tabnum{\underline{88.36}} & \tabnum{\textbf{78.14}} & \tabnum{65.00} & \tabnum{\underline{87.18}} & \tabnum{\underline{84.42}} \\
\midrule
\rowcolor{bioloomlight}
BioLoom (Scratch) & \tabnum{\textbf{95.50}} & \tabnum{\textbf{94.83}} & \tabnum{81.50} & \tabnum{\textbf{90.60}} & \tabnum{\textbf{93.75}} & \tabnum{97.11} & \tabnum{59.30} & \tabnum{88.23} & \tabnum{83.63} & \tabnum{75.38} & \tabnum{\textbf{67.00}} & \tabnum{86.54} & \tabnum{\textbf{84.45}} \\
\bottomrule
\end{tabularx}
\ifdefined\jointlinearprobes\endgroup\else\end{table*}\fi

\par\vspace{4pt}

\ifdefined\jointlinearprobes\begingroup\else\begin{table*}[t]\fi
\caption{\textbf{Macro-F1 of linear probes on MedMNIST.} Scores are percentages; column names omit the MNIST suffix. The mean gives equal weight to all 12 datasets. Bold and underlined values indicate the best and second-best scores in each column, respectively.}
\label{tab:linear-probe-f1}
\centering
\resulttablefont
\renewcommand{\arraystretch}{1.0}
\setlength{\tabcolsep}{1.3pt}
\begin{tabularx}{\textwidth}{@{}>{\raggedright\arraybackslash}p{0.22\textwidth}*{13}{Y}@{}}
\toprule
Model & Path & Chest & Derma & OCT & Pneu. & Blood & Tissue & OrganA & OrganC & OrganS & Retina & Breast & Mean \\
\midrule
CoCa ViT-L/14 & \tabnum{87.66} & \tabnum{0.39} & \tabnum{59.74} & \tabnum{64.53} & \tabnum{81.51} & \tabnum{92.82} & \tabnum{45.98} & \tabnum{85.66} & \tabnum{78.48} & \tabnum{66.71} & \tabnum{42.48} & \tabnum{71.16} & \tabnum{64.76} \\
EVA-CLIP ViT-L/14~\cite{sun2023evaclip} & \tabnum{86.44} & \tabnum{0.60} & \tabnum{58.11} & \tabnum{66.40} & \tabnum{76.26} & \tabnum{94.11} & \tabnum{48.94} & \tabnum{88.13} & \tabnum{79.88} & \tabnum{68.97} & \tabnum{42.56} & \tabnum{73.15} & \tabnum{65.30} \\
RetCCL ResNet-50~\cite{wang2023retccl} & \tabnum{\underline{92.86}} & \tabnum{2.20} & \tabnum{53.69} & \tabnum{53.53} & \tabnum{84.83} & \tabnum{\underline{98.02}} & \tabnum{46.19} & \tabnum{85.61} & \tabnum{81.42} & \tabnum{70.07} & \tabnum{43.48} & \tabnum{75.55} & \tabnum{65.62} \\
SigLIP2 ViT-L/16~\cite{tschannen2025siglip2} & \tabnum{89.16} & \tabnum{1.45} & \tabnum{57.00} & \tabnum{67.75} & \tabnum{86.62} & \tabnum{94.08} & \tabnum{47.48} & \tabnum{89.54} & \tabnum{80.39} & \tabnum{68.95} & \tabnum{43.28} & \tabnum{68.41} & \tabnum{66.18} \\
ImageNet-1k ResNet-50 & \tabnum{86.90} & \tabnum{3.59} & \tabnum{54.55} & \tabnum{67.57} & \tabnum{84.67} & \tabnum{95.52} & \tabnum{47.46} & \tabnum{91.05} & \tabnum{85.11} & \tabnum{71.46} & \tabnum{39.58} & \tabnum{81.20} & \tabnum{67.39} \\
SigLIP ViT-L/16 & \tabnum{91.48} & \tabnum{1.77} & \tabnum{67.81} & \tabnum{72.25} & \tabnum{86.06} & \tabnum{95.89} & \tabnum{48.05} & \tabnum{88.71} & \tabnum{80.05} & \tabnum{69.05} & \tabnum{41.82} & \tabnum{67.60} & \tabnum{67.55} \\
OpenCLIP (CommonPool) & \tabnum{86.62} & \tabnum{0.64} & \tabnum{62.41} & \tabnum{68.89} & \tabnum{80.01} & \tabnum{95.63} & \tabnum{50.74} & \tabnum{89.60} & \tabnum{81.68} & \tabnum{68.68} & \tabnum{46.36} & \tabnum{80.04} & \tabnum{67.61} \\
MetaCLIP ViT-L/14~\cite{xu2024metaclip} & \tabnum{90.10} & \tabnum{0.94} & \tabnum{65.43} & \tabnum{75.19} & \tabnum{85.41} & \tabnum{95.32} & \tabnum{50.38} & \tabnum{86.87} & \tabnum{79.43} & \tabnum{67.66} & \tabnum{42.20} & \tabnum{73.87} & \tabnum{67.73} \\
EVA-02 ViT-L/14~\cite{fang2024eva02} & \tabnum{91.63} & \tabnum{1.33} & \tabnum{58.91} & \tabnum{71.00} & \tabnum{83.75} & \tabnum{96.42} & \tabnum{48.23} & \tabnum{90.28} & \tabnum{86.19} & \tabnum{71.38} & \tabnum{44.59} & \tabnum{77.43} & \tabnum{68.43} \\
Phikon-v2 ViT-L~\cite{filiot2024phikonv2} & \tabnum{91.50} & \tabnum{3.15} & \tabnum{63.24} & \tabnum{69.70} & \tabnum{84.96} & \tabnum{97.96} & \tabnum{50.63} & \tabnum{86.45} & \tabnum{81.09} & \tabnum{70.56} & \tabnum{47.71} & \tabnum{76.06} & \tabnum{68.58} \\
CLIP (DFN-2B) ViT-L/14 & \tabnum{89.85} & \tabnum{1.26} & \tabnum{66.64} & \tabnum{71.14} & \tabnum{83.10} & \tabnum{95.97} & \tabnum{50.90} & \tabnum{89.68} & \tabnum{82.88} & \tabnum{68.69} & \tabnum{43.85} & \tabnum{\textbf{84.50}} & \tabnum{69.04} \\
DINOv2 ViT-L/14~\cite{oquab2024dinov2} & \tabnum{91.11} & \tabnum{1.08} & \tabnum{63.58} & \tabnum{78.96} & \tabnum{86.02} & \tabnum{96.64} & \tabnum{\underline{52.36}} & \tabnum{89.59} & \tabnum{83.80} & \tabnum{72.29} & \tabnum{\textbf{52.66}} & \tabnum{73.77} & \tabnum{70.15} \\
\mbox{PMC-CLIP} ResNet-50 & \tabnum{90.27} & \tabnum{\underline{5.06}} & \tabnum{62.40} & \tabnum{66.34} & \tabnum{\underline{90.29}} & \tabnum{96.81} & \tabnum{51.90} & \tabnum{\textbf{92.07}} & \tabnum{\textbf{88.63}} & \tabnum{\underline{72.62}} & \tabnum{49.82} & \tabnum{76.74} & \tabnum{70.25} \\
ImageNet-21k ViT-L/16 & \tabnum{91.35} & \tabnum{3.54} & \tabnum{\textbf{68.91}} & \tabnum{\underline{79.50}} & \tabnum{85.53} & \tabnum{\textbf{98.35}} & \tabnum{\textbf{54.05}} & \tabnum{\underline{91.58}} & \tabnum{\underline{86.86}} & \tabnum{\textbf{73.96}} & \tabnum{\underline{52.01}} & \tabnum{\underline{82.60}} & \tabnum{\textbf{72.36}} \\
\midrule
\rowcolor{bioloomlight}
BioLoom (Scratch) & \tabnum{\textbf{93.49}} & \tabnum{\textbf{8.52}} & \tabnum{\underline{68.24}} & \tabnum{\textbf{90.45}} & \tabnum{\textbf{93.09}} & \tabnum{96.99} & \tabnum{46.86} & \tabnum{86.71} & \tabnum{80.36} & \tabnum{70.47} & \tabnum{50.00} & \tabnum{81.22} & \tabnum{\underline{72.20}} \\
\bottomrule
\end{tabularx}
\ifdefined\jointlinearprobes\endgroup\else\end{table*}\fi

\par\vspace{4pt}

\ifdefined\jointlinearprobes\begingroup\else\begin{table*}[t]\fi
\caption{\textbf{AUC of linear probes on MedMNIST.} Scores are percentages; column names omit the MNIST suffix. The mean gives equal weight to all 12 datasets. Bold and underlined values indicate the best and second-best scores in each column, respectively.}
\label{tab:linear-probe-auc}
\centering
\resulttablefont
\renewcommand{\arraystretch}{1.0}
\setlength{\tabcolsep}{1.3pt}
\begin{tabularx}{\textwidth}{@{}>{\raggedright\arraybackslash}p{0.22\textwidth}*{13}{Y}@{}}
\toprule
Model & Path & Chest & Derma & OCT & Pneu. & Blood & Tissue & OrganA & OrganC & OrganS & Retina & Breast & Mean \\
\midrule
CoCa ViT-L/14 & \tabnum{99.23} & \tabnum{66.00} & \tabnum{95.18} & \tabnum{96.48} & \tabnum{96.46} & \tabnum{99.66} & \tabnum{88.34} & \tabnum{99.12} & \tabnum{98.24} & \tabnum{96.62} & \tabnum{83.63} & \tabnum{86.93} & \tabnum{92.16} \\
EVA-CLIP ViT-L/14~\cite{sun2023evaclip} & \tabnum{99.23} & \tabnum{65.33} & \tabnum{94.76} & \tabnum{97.03} & \tabnum{95.41} & \tabnum{99.72} & \tabnum{89.53} & \tabnum{99.30} & \tabnum{98.59} & \tabnum{96.94} & \tabnum{83.58} & \tabnum{86.95} & \tabnum{92.20} \\
EVA-02 ViT-L/14~\cite{fang2024eva02} & \tabnum{99.53} & \tabnum{67.80} & \tabnum{95.03} & \tabnum{97.98} & \tabnum{97.15} & \tabnum{99.86} & \tabnum{89.05} & \tabnum{99.53} & \tabnum{99.13} & \tabnum{97.30} & \tabnum{82.07} & \tabnum{86.53} & \tabnum{92.58} \\
RetCCL ResNet-50~\cite{wang2023retccl} & \tabnum{99.57} & \tabnum{70.92} & \tabnum{93.86} & \tabnum{95.00} & \tabnum{96.96} & \tabnum{99.90} & \tabnum{88.59} & \tabnum{98.95} & \tabnum{98.55} & \tabnum{97.02} & \tabnum{83.44} & \tabnum{88.18} & \tabnum{92.58} \\
OpenCLIP (CommonPool) & \tabnum{99.51} & \tabnum{67.77} & \tabnum{96.06} & \tabnum{95.61} & \tabnum{95.86} & \tabnum{99.81} & \tabnum{89.94} & \tabnum{99.43} & \tabnum{98.76} & \tabnum{96.93} & \tabnum{84.55} & \tabnum{89.16} & \tabnum{92.78} \\
SigLIP2 ViT-L/16~\cite{tschannen2025siglip2} & \tabnum{99.29} & \tabnum{72.36} & \tabnum{95.11} & \tabnum{98.06} & \tabnum{97.73} & \tabnum{99.73} & \tabnum{88.72} & \tabnum{99.36} & \tabnum{98.48} & \tabnum{96.74} & \tabnum{83.91} & \tabnum{86.42} & \tabnum{92.99} \\
DINOv2 ViT-L/14~\cite{oquab2024dinov2} & \tabnum{99.36} & \tabnum{62.88} & \tabnum{96.21} & \tabnum{\underline{99.14}} & \tabnum{97.52} & \tabnum{99.86} & \tabnum{\underline{90.74}} & \tabnum{99.35} & \tabnum{98.82} & \tabnum{97.29} & \tabnum{\textbf{88.77}} & \tabnum{88.45} & \tabnum{93.20} \\
MetaCLIP ViT-L/14~\cite{xu2024metaclip} & \tabnum{\underline{99.58}} & \tabnum{69.21} & \tabnum{95.98} & \tabnum{97.77} & \tabnum{97.59} & \tabnum{99.83} & \tabnum{89.90} & \tabnum{99.15} & \tabnum{98.56} & \tabnum{96.85} & \tabnum{84.99} & \tabnum{89.43} & \tabnum{93.24} \\
SigLIP ViT-L/16 & \tabnum{99.44} & \tabnum{73.67} & \tabnum{96.30} & \tabnum{98.47} & \tabnum{97.84} & \tabnum{99.86} & \tabnum{89.22} & \tabnum{99.41} & \tabnum{98.57} & \tabnum{96.88} & \tabnum{84.20} & \tabnum{85.17} & \tabnum{93.25} \\
CLIP (DFN-2B) ViT-L/14 & \tabnum{\underline{99.58}} & \tabnum{69.57} & \tabnum{\textbf{96.87}} & \tabnum{98.41} & \tabnum{97.33} & \tabnum{99.87} & \tabnum{89.97} & \tabnum{99.47} & \tabnum{98.90} & \tabnum{96.96} & \tabnum{84.25} & \tabnum{89.66} & \tabnum{93.40} \\
Phikon-v2 ViT-L~\cite{filiot2024phikonv2} & \tabnum{99.49} & \tabnum{73.45} & \tabnum{96.10} & \tabnum{96.37} & \tabnum{97.93} & \tabnum{\underline{99.91}} & \tabnum{90.16} & \tabnum{99.01} & \tabnum{98.75} & \tabnum{97.23} & \tabnum{82.77} & \tabnum{92.50} & \tabnum{93.64} \\
ImageNet-1k ResNet-50 & \tabnum{99.15} & \tabnum{74.01} & \tabnum{93.51} & \tabnum{98.21} & \tabnum{97.18} & \tabnum{99.81} & \tabnum{88.69} & \tabnum{\underline{99.59}} & \tabnum{99.05} & \tabnum{97.31} & \tabnum{83.99} & \tabnum{\underline{93.25}} & \tabnum{93.65} \\
\mbox{PMC-CLIP} ResNet-50 & \tabnum{99.52} & \tabnum{\underline{76.51}} & \tabnum{95.76} & \tabnum{97.30} & \tabnum{\underline{98.97}} & \tabnum{99.88} & \tabnum{90.62} & \tabnum{99.52} & \tabnum{\textbf{99.37}} & \tabnum{\textbf{97.68}} & \tabnum{84.77} & \tabnum{92.40} & \tabnum{\underline{94.36}} \\
ImageNet-21k ViT-L/16 & \tabnum{99.52} & \tabnum{74.59} & \tabnum{\underline{96.80}} & \tabnum{98.93} & \tabnum{98.01} & \tabnum{\textbf{99.95}} & \tabnum{\textbf{91.12}} & \tabnum{\textbf{99.60}} & \tabnum{\underline{99.21}} & \tabnum{\underline{97.66}} & \tabnum{86.39} & \tabnum{\textbf{94.28}} & \tabnum{\textbf{94.67}} \\
\midrule
\rowcolor{bioloomlight}
BioLoom (Scratch) & \tabnum{\textbf{99.60}} & \tabnum{\textbf{78.05}} & \tabnum{95.79} & \tabnum{\textbf{99.61}} & \tabnum{\textbf{99.27}} & \tabnum{99.86} & \tabnum{88.27} & \tabnum{99.11} & \tabnum{98.64} & \tabnum{97.00} & \tabnum{\underline{87.25}} & \tabnum{87.68} & \tabnum{94.18} \\
\bottomrule
\end{tabularx}
\ifdefined\jointlinearprobes\endgroup\else\end{table*}\fi

\end{table*}
\endgroup

\begin{table*}[t]
\centering
\caption{\textbf{Zero-shot classification results on MMBU.} Accuracy (\%) is shown. BBox denotes a bounding box; BBox and mask indicate the region annotations displayed in the released classification images. $^{\ast}$ denotes random vision-encoder initialization for image-text pretraining. Among biomedical models, bold and underlined values indicate the best and second-best displayed scores in each column, respectively.}
\label{tab:mmbu-option-caption}
\resulttablefont
\setlength{\tabcolsep}{3.5pt}
\renewcommand{\arraystretch}{1.12}
\begin{tabularx}{\textwidth}{>{\raggedright\arraybackslash}p{0.2\textwidth}*{3}{>{\centering\arraybackslash}X}}
\toprule
\textbf{Model} & \shortstack[c]{\textbf{Whole-image}\\\textbf{classification} ($n=1{,}813$)} & \shortstack[c]{\textbf{BBox-indicated}\\\textbf{classification} ($n=343$)} & \shortstack[c]{\textbf{Mask-indicated}\\\textbf{classification} ($n=521$)}\\
\midrule
CoCa$^{\ast}$~\cite{yu2022cocacontrastivecaptionersimagetext} & \tabnum{27.03} & \tabnum{16.33} & \tabnum{18.43}\\
OpenCLIP$^{\ast}$~\cite{cherti2023reproduciblescalinglawscontrastive} & \tabnum{27.25} & \tabnum{21.57} & \tabnum{22.84}\\
    \midrule
MedSigLIP~\cite{sellergren2025medgemmatechnicalreport} & \tabnum{32.32} & \tabnum{22.16} & \tabnum{19.77}\\
    Panel2Patch~\cite{yuan2026panel2patch} & \tabnum{34.64} & \tabnum{22.16} & \tabnum{22.84} \\
BMC-CLIP-CF~\cite{lozano2025biomedica} & \tabnum{31.94} & \tabnum{25.36} & \tabnum{22.46}\\
PMC-CLIP (ResNet-50)$^{\ast}$~\cite{lin2023pmcclipcontrastivelanguageimagepretraining} & \tabnum{34.20} & \tabnum{25.36} & \tabnum{24.18}\\
BiomedCLIP~\cite{zhang2023biomedclipmultimodalbiomedicalfoundation} & \tabnum{32.93} & \tabnum{29.74} & \tabnum{22.65}\\
MedPMC-CLIP~\cite{kim2026medpmc} & \tabnum{\underline{41.53}} & \tabnum{29.74} & \tabnum{27.64}\\
BMC-LongCLIP~\cite{sun2025tokenswastedleveraginglong} & \tabnum{39.00} & \tabnum{30.61} & \tabnum{\underline{29.37}}\\
\midrule
\rowcolor{bioloomlight}
\methodname{} (Scratch)$^{\ast}$ & \tabnum{\textbf{42.47}} & \tabnum{\textbf{36.73}} & \tabnum{\textbf{33.59}}\\
\rowcolor{bioloomlight}
\methodname{} (DFN-2B) & \tabnum{34.31} & \tabnum{\underline{32.94}} & \tabnum{28.02}\\
\bottomrule
\end{tabularx}

\end{table*}
\begin{table*}[t]
\centering
\caption{\textbf{Model comparison on fine-grained and cross-level retrieval on BioLoomBench.} Recall@1 (\%) is reported for each task. Arrows indicate query $\rightarrow$ retrieval target. Panel denotes a cropped subfigure, and whole figure denotes an uncropped figure. Description denotes generated panel text; original caption denotes the whole-figure caption from the article. $^{\ast}$ denotes random vision-encoder initialization for image-text pretraining. Among biomedical models, bold marks the best score and underlining the second-best in each column.}
\label{tab:panel-retrieval}
\resulttablefont
\setlength{\tabcolsep}{2pt}
\renewcommand{\arraystretch}{1.12}
\begin{tabularx}{\textwidth}{l>{\hsize=.90\hsize\linewidth=\hsize\centering\arraybackslash}X>{\hsize=.90\hsize\linewidth=\hsize\centering\arraybackslash}X>{\hsize=1.15\hsize\linewidth=\hsize\centering\arraybackslash}X>{\hsize=1.05\hsize\linewidth=\hsize\centering\arraybackslash}X}
\toprule
\textbf{Model} & \mbox{\textbf{Panel $\rightarrow$ description}} & \mbox{\textbf{Description $\rightarrow$ panel}} & \mbox{\textbf{Description $\rightarrow$ whole figure}} & \mbox{\textbf{Panel $\rightarrow$ original caption}}\\
\midrule
CoCa$^{\ast}$~\cite{yu2022cocacontrastivecaptionersimagetext} & \tabnum{22.57} & \tabnum{21.02} & \tabnum{7.58} & \tabnum{5.09}\\
OpenCLIP$^{\ast}$~\cite{cherti2023reproduciblescalinglawscontrastive} & \tabnum{25.76} & \tabnum{23.23} & \tabnum{10.73} & \tabnum{8.67}\\
\midrule
PMC-CLIP$^{\ast}$~\cite{lin2023pmcclipcontrastivelanguageimagepretraining} & \tabnum{6.50} & \tabnum{5.61} & \tabnum{1.08} & \tabnum{5.53}\\
MedSigLIP~\cite{sellergren2025medgemmatechnicalreport} & \tabnum{16.91} & \tabnum{8.38} & \tabnum{5.84} & \tabnum{8.04}\\
BMC-LongCLIP~\cite{sun2025tokenswastedleveraginglong} & \tabnum{30.31} & \tabnum{28.40} & \tabnum{7.89} & \tabnum{12.72}\\
BiomedCLIP~\cite{zhang2023biomedclipmultimodalbiomedicalfoundation} & \tabnum{32.54} & \tabnum{30.34} & \tabnum{7.92} & \tabnum{10.94}\\
Panel2Patch~\cite{yuan2026panel2patch} & \tabnum{38.06} & \tabnum{35.34} & \tabnum{15.67} & \tabnum{18.01}\\
BMC-CLIP-CF~\cite{lozano2025biomedica} & \tabnum{39.40} & \tabnum{35.86} & \tabnum{14.49} & \tabnum{14.54}\\
MedPMC-CLIP~\cite{kim2026medpmc} & \tabnum{41.27} & \tabnum{38.32} & \tabnum{17.86} & \tabnum{22.36}\\
\midrule
\rowcolor{bioloomlight}
BioLoom (Scratch)$^{\ast}$ & \tabnum{\underline{48.18}} & \tabnum{\underline{38.35}} & \tabnum{\underline{19.21}} & \tabnum{\underline{25.69}}\\
\rowcolor{bioloomlight}
BioLoom (DFN-2B) & \tabnum{\textbf{51.78}} & \tabnum{\textbf{43.22}} & \tabnum{\textbf{27.21}} & \tabnum{\textbf{43.76}}\\
\bottomrule
\end{tabularx}
\end{table*}
\end{document}